\documentclass[10pt]{article} 
\usepackage{xcolor} 
\definecolor{mygray}{RGB}{80,80,80}
\definecolor{lightpurple}{RGB}{153,102,204}

\definecolor{YesColor}{RGB}{220,240,220}      
\definecolor{LimitedColor}{RGB}{255,245,220}  
\definecolor{NoColor}{RGB}{245,220,220}       
\definecolor{BothColor}{RGB}{220,235,255}     

\definecolor{DeepBlue}{HTML}{1f77b4}
\definecolor{BurntOrange}{HTML}{ff7f0e}
\definecolor{ForestGreen}{HTML}{2ca02c}
\definecolor{PlumPurple}{HTML}{9467bd}

\definecolor{RoyalBlue}{HTML}{0047AB}
\definecolor{CrimsonRed}{HTML}{DC143C}
\definecolor{DarkGreen}{HTML}{006400}
\definecolor{Goldenrod}{HTML}{DAA520}

\usepackage{tikz}
\usepackage{pgf-pie}
\usetikzlibrary{positioning, shapes.geometric, arrows.meta}
\usepackage{wrapfig}
\usetikzlibrary{fit}
\usepackage{subcaption} 
\tikzset{
  block/.style={
  rectangle,
  draw,
  fill=gray!10,
  thick,
  minimum width=3cm,
  minimum height=1cm,
  rounded corners,
  text width=3cm,    
  align=center       
  },
  child/.style={
  rectangle, 
  draw,
  thick,
  fill=gray!10,
  minimum width=3cm, 
  minimum height=1cm,
  rounded corners,
  text width=2.6cm, 
  align=center
  },
  arrow/.style={
  -{Stealth}, 
  thick,
  draw=gray,
  },
}

\usepackage[preprint]{tmlr}

\usepackage{enumitem}

\usepackage{amsmath,amsfonts,bm}

\def\eqref#1{equation~\ref{#1}}

\def\1{\bm{1}}

\DeclareMathAlphabet{\mathsfit}{\encodingdefault}{\sfdefault}{m}{sl}
\SetMathAlphabet{\mathsfit}{bold}{\encodingdefault}{\sfdefault}{bx}{n}

\usepackage{hyperref}
\usepackage{url}
\usepackage{booktabs}
\usepackage{adjustbox}
\usepackage{multirow}
\usepackage{mathtools}

\usepackage[normalem]{ulem}

\title{Distributed Hybrid Parallelism for Large Language Models: Comparative Study and System Design Guide}

\author{\name Hossam Amer$^*$ 
        \email hossam.amer1@huawei.com \\
        \addr Toronto Ascend Team, Huawei Canada
        \AND
        \name Rezaul Karim$^*$ 
        \email rezaul.karim3@huawei.com \\
        \addr Toronto Ascend Team, Huawei Canada
        \AND
        \name Ali Pourranjbar 
        \email ali.pourranjbar@h-partners.com\\
        \addr Toronto Ascend Team, Huawei Canada
        \AND
        \name Weiwei Zhang 
        \email weiwei.zhang2@huawei.com\\
        \addr Toronto Ascend Team, Huawei Canada
        \AND
        \name Walid Ahmed 
        \email walid.ahmed1@huawei.com\\
        \addr Toronto Ascend Team, Huawei Canada
        \AND
        \name Boxing Chen 
        \email boxing.chen@huawei.com\\
        \addr Toronto Ascend Team, Huawei Canada
        }

\def\month{MM}  
\def\year{YYYY} 
\def\openreview{\url{https://openreview.net/forum?id=XXXX}} 

\usepackage{hyperref}
\hypersetup{pdfborder=0 0 0}

\begin{document}

\maketitle
\renewcommand{\thefootnote}{\fnsymbol{footnote}}
\footnotetext[1]{Equal contribution.}

\begin{abstract}

With the rapid growth of large language models (LLMs), a wide range of methods have been developed to distribute computation and memory across hardware devices for efficient training and inference. While existing surveys provide descriptive overviews of these techniques, systematic analysis of their benefits and trade-offs—and how such insights can inform principled methodology for designing optimal distributed systems—remain limited. This paper offers a comprehensive review of collective operations and distributed parallel strategies, complemented by mathematical formulations to deepen theoretical understanding. We further examine hybrid parallelization designs, emphasizing communication–computation overlap across different stages of model deployment, including both training and inference. Recent advances in automated search for optimal hybrid parallelization strategies using cost models are also discussed. Moreover, we present case studies with mainstream architecture categories to reveal empirical insights to guide researchers and practitioners in parallelism strategy selection. Finally, we highlight open challenges and limitations of current LLM training paradigms and outline promising directions for the next generation of large-scale model development.


\end{abstract}

\pagebreak
\tableofcontents
\newpage

\section{Introduction}

The empirical findings of neural scaling laws demonstrate continued performance gains when scaling model size, training data, and computation budget, underscoring the importance of efficient distributed training strategies~\cite{kaplan2020scaling}. In recent years, models with hundreds of billions to trillions of parameters have achieved unprecedented results across diverse benchmarks, marking a significant step toward artificial general intelligence~\cite{gamal2023federated,adler2024nemotron,amer2024device,grattafiori2024llama,guo2025deepseek, tawfilis2025distributed}. These successes, coupled with the scaling law evidence, have fueled growing demand for efficient training and inference on large distributed systems comprising GPUs, TPUs, or NPUs.

To meet these demands, modern systems combine parallelization techniques such as Data, Pipeline, Tensor, and Context Parallelism, often augmented with memory optimization techniques like activation recomputation~\cite{chen2016training} and distributed optimizers~\cite{rajbhandari2020zero}. Communication overlap methods further reduce bottlenecks in distributed environment. Notably, the best mix of strategies varies between training and inference due to differing workload requirements~\cite{rasley2020deepspeed,grattafiori2024llama}. Since each parallelization method introduces distinct communication and efficiency trade-offs, there is a pressing need for a unified understanding of how to integrate them effectively.

Despite progress, distributed strategies for large-scale model training have often been developed independently, leaving a fragmented understanding of the overall landscape. Existing surveys provide valuable overviews~\cite{duan2024efficient,zhou2024training}, but a deeper analysis of their trade offs in balancing scalability, memory efficiency, and communication cost is needed to guide system design. Even a modest efficiency gains in  a distributed strategy design for a large scale model can yield significant benefits as training of foundation models consumes vast energy resources~\cite{strubell2020energy}. Hardware-aware optimization and datacenter efficiency can further mitigate the carbon impact~\cite{patterson2021carbon}, aligning distributed AI with the principles of “Green AI,” where sustainability is prioritized alongside accuracy~\cite{schwartz2020green}. Therefore, a holistic perspective on distributed strategies is crucial for both advancing large-scale AI and ensuring ecological sustainability.

\begin{figure}[ht]
  \centering
\resizebox{\textwidth}{!}{

\begin{tikzpicture}[node distance=1cm and 1cm]

\node[block, minimum height=7.4cm] (start) { 
{ 
\textcolor{RoyalBlue}{Parallelism \\ for \\Large \\ Language \\  Models}
}};

\node[block, right=4cm of start.north, anchor=north] (box1) { \textcolor{blue!70!black}{ Background and  Fundamentals} \ref{sec:distributed_fundamentals}};
\node[block, below=of box1] (box2) {
\textcolor{blue!30!red}{Hybrid Parallelization Systems} \ref{sec:hybrid_systems}};
\node[block, below=of box2] (box3) {
\textcolor{green!50!blue}{System Design Approaches} \ref{sec:system_design}};
\node[block, below=of box3] (box4) {
\textcolor{blue!70!red}{Systematic Analysis and Insights}
};

\draw[arrow] (start.east|-box1) -- (box1.west);
\draw[arrow] (start.east|-box2) -- (box2.west);
\draw[arrow] (start.east|-box3) -- (box3.west);
\draw[arrow] (start.east|-box4) -- (box4.west);

\node[child, right=1cm of box1] (b1c1) {
\textcolor{blue!70!black}{Collective Operations}  \ref{sec:collective_ops}

};
\node[child, right=of b1c1] (b1c2) {
\textcolor{blue!70!black}{Parallelization} \ref{sec:data_parallel} ~\ref{sec:parallelization} \ref{sec:activation_parallel}
};
\node[child, right=of b1c2] (b1c3) {
\textcolor{blue!70!black}{Memory Reduction} \ref{sec:memory_optimization}
};
\draw[arrow] (box1.east) -- (b1c1.west);
\draw[arrow] (b1c1.east) -- (b1c2.west);
\draw[arrow] (b1c2.east) -- (b1c3.west);

\node[child, right=1cm of box2] (b2c1) {
\textcolor{blue!30!red}{3D/4D Multi-parallel} \ref{sec:multiparallelism}};
\node[child, right=of b2c1] (b2c2) {
\textcolor{blue!30!red}{Frameworks} \ref{sec:frameworks} };
\node[child, right=of b2c2] (b2c3) { \textcolor{blue!70!black}{Communication Overlap} \ref{sec:comm_overlap} };
\draw[arrow] (box2.east) -- (b2c1.west);
\draw[arrow] (b2c1.east) -- (b2c2.west);
\draw[arrow] (b1c3.south) -- (b2c3.north);

\node[child, right=1cm of box3] (b3c1) { 
\textcolor{green!50!blue}{Design Implication} \ref{sec:system_design_impl} };
\node[child, right=of b3c1] (b3c2) {
\textcolor{green!50!blue}{Design Consideration} \ref{sec:system_design_cons}};
\node[child, right=of b3c2] (b3c3) {
\textcolor{green!50!blue}{Auto Parallelism} \ref{sec:auto_parallel}};
\draw[arrow] (box3.east) -- (b3c1.west);
\draw[arrow] (b3c1.east) -- (b3c2.west);
\draw[arrow] (b3c2.east) -- (b3c3.west);

\node[child, right=1cm of box4] (b4c1) {
\textcolor{blue!70!red}{Theoretical Analysis} \ref{sec:theory_parallel_strategies}};
\node[child, right=of b4c1] (b4c2) {
\textcolor{blue!70!red}{Experimental Analysis} \ref{sec:cae_studies} };
\node[block, right=of b4c2] (b4c3) {
\textcolor{blue!70!red}{Future Directions} \\ \ref{sec:future_directions}
};
\draw[arrow] (box4.east) -- (b4c1.west);
\draw[arrow] (b4c1.east) -- (b4c2.west);
\draw[arrow] (b4c2.east) -- (b4c3.west);

\end{tikzpicture}
}
\caption{Overview of the key aspects of scalable efficient distributed systems for AI workloads that we cover in this study.}
\label{fig:outlook}
\end{figure}
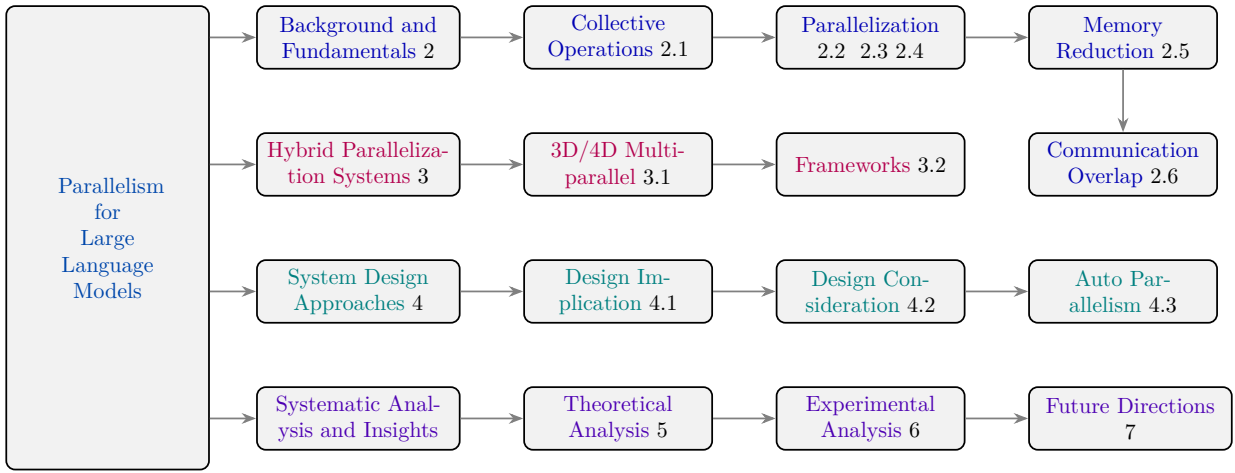

In this paper, we address the above needs and fill the gap in literature with a comprehensive review, design guidelines, systematic analysis and insights. We omit discussion on model compression techniques~\cite{zhu2024survey,girija2025optimizing} including pruning~\cite{cheng2024survey}, quantization~\cite{yang2025survey}, low-rank adaptation~\cite{yang2024low} in favor of broader discussion on system level optimization. The key concepts covered in this study are highlighted in Figure~\ref{fig:outlook}. We begin with a comprehensive background on parallel distributed strategies and collective operations. Then, we turn to hybrid parallelization and their associated deep learning frameworks. We also add a section to discuss the parallelization system design considerations including evaluation metrics with discussions on training and inference considerations. Based on this, we formulate the hybrid parallelization problem for system design, review existing frameworks, examine automatic parallelization methods. We support our analysis with theoretical and empirical validations.

This study goes beyond a traditional survey by using a comparative study, system design guidelines, and practical use cases as highlighted in Table~\ref{tab:survey-comparison}. We offer this paper to both researchers and practitioners as a guide for distributed parallel strategies to support model development. While this paper does not aim to be fully exhaustive, we hope it provides a useful foundation upon which the community can build and extend the analyses. A concise outline of our key contributions include:

\begin{itemize}
\item {Present a cohesive and coherent literature review covering parallelization strategies, memory reduction, communication overlap, hybrid system design, and automatic parallelization aiming for a holistic understanding of efficient training and inference in distributed setting.}  

\item {Present a comparative study of tradeoffs and benefits of workload distribution strategies accompanied by examples to provide practical insight and use cases of these strategies.}  


\item {Highlight our key findings from this study that would be insightful for researcher and practitioners in this field. Key findings highlight the need for distinct parallelization strategies tailored to the unique workload requirements of training and inference. These strategies are essential for hybrid approaches that combine complementary methods to achieve optimal performance in both stages. Additionally, efficient system design requires jointly considering memory, compute throughput, bandwidth, and collective communication patterns. }

\item {Consolidate our insights on emerging trends, emphasizing automated discovery of optimal parallelization strategies across large-scale cluster topologies. This involves leveraging cost models and aligning with service-level requirements to guide efficient strategy selection.}

\item {Provide both theoretical and empirical analysis to parallel strategies to support researchers and practitioners in parallel strategy selection.}

\item {Summarize important future key points that can be extended from this paper.}


\end{itemize}

\begin{table}[ht]
\centering
\caption{Comparison of related surveys on distributed parallel strategies.}
\label{tab:survey-comparison}
\resizebox{\textwidth}{!}{%
\renewcommand{\arraystretch}{1.2}

\definecolor{YesColor}{RGB}{220,240,220}      
\definecolor{LimitedColor}{RGB}{255,245,220}  
\definecolor{NoColor}{RGB}{245,220,220}       
\definecolor{BothColor}{RGB}{220,235,255}     

\begin{tabular}{|p{0.35\textwidth}|p{0.12\textwidth}|p{0.09\textwidth}|p{0.12\textwidth}|p{0.12\textwidth}|p{0.07\textwidth}|}
\hline
\textbf{Category} &
\textbf{\small{Zeng et al. (2023) \cite{zeng2023distributed}}} &
\textbf{\small{Li et al. (2024) \cite{li2024efficient}}} &
\textbf{\small{Duan et al. (2024) \cite{duan2024efficient}}} &
\textbf{\small{Zeng et al. (2025) \cite{zeng2025distributed}}} &
\textbf{This survey} \\
\hline

Covers distributed parallel strategies broadly
& \colorbox{LimitedColor}{Limited}
& \colorbox{LimitedColor}{Limited}
& \colorbox{YesColor}{Yes}
& \colorbox{YesColor}{Yes}
& \colorbox{YesColor}{Yes} \\ \hline

Covers memory reduction strategies broadly
& \colorbox{YesColor}{Yes}
& \colorbox{YesColor}{Yes}
& \colorbox{YesColor}{Yes}
& \colorbox{YesColor}{Yes}
& \colorbox{YesColor}{Yes} \\ \hline

Covers communication overlap strategies broadly
& \colorbox{NoColor}{No}
& \colorbox{NoColor}{No}
& \colorbox{YesColor}{Yes}
& \colorbox{YesColor}{Yes}
& \colorbox{YesColor}{Yes} \\ \hline

Discusses distributed parallel frameworks
& \colorbox{LimitedColor}{Limited}
& \colorbox{YesColor}{Yes}
& \colorbox{YesColor}{Yes}
& \colorbox{NoColor}{No}
& \colorbox{YesColor}{Yes} \\ \hline

Discusses auto parallelism
& \colorbox{NoColor}{No}
& \colorbox{NoColor}{No}
& \colorbox{YesColor}{Yes}
& \colorbox{LimitedColor}{Limited}
& \colorbox{YesColor}{Yes} \\ \hline

Discusses the parallel strategy selection success metrics
& \colorbox{NoColor}{No}
& \colorbox{YesColor}{Yes}
& \colorbox{NoColor}{No}
& \colorbox{LimitedColor}{Limited}
& \colorbox{YesColor}{Yes} \\ \hline

Covers both training and inference
& \colorbox{NoColor}{Training}
& \colorbox{YesColor}{Yes}
& \colorbox{NoColor}{Training}
& \colorbox{NoColor}{Training}
& \colorbox{YesColor}{Yes} \\ \hline

Provides mathematical and analytical understanding to parallel strategies
& \colorbox{NoColor}{No}
& \colorbox{NoColor}{No}
& \colorbox{NoColor}{No}
& \colorbox{YesColor}{Yes}
& \colorbox{YesColor}{Yes} \\ \hline

Provides empirical analysis of parallel strategy selection using case studies
& \colorbox{NoColor}{No}
& \colorbox{NoColor}{No}
& \colorbox{NoColor}{No}
& \colorbox{NoColor}{No}
& \colorbox{YesColor}{Yes} \\ \hline

Provides parallelization strategy selection guidelines
& \colorbox{NoColor}{No}
& \colorbox{NoColor}{No}
& \colorbox{LimitedColor}{Limited}
& \colorbox{YesColor}{Yes}
& \colorbox{YesColor}{Yes} \\ \hline

Proposes future research directions
& \colorbox{LimitedColor}{Limited}
& \colorbox{LimitedColor}{Limited}
& \colorbox{LimitedColor}{Limited}
& \colorbox{YesColor}{Yes}
& \colorbox{YesColor}{Yes} \\ \hline

\end{tabular}
}
\end{table}

\section{Distributed Strategies Background}
\label{sec:distributed_fundamentals}

Efficient distributed deep learning strategies rely on utilizing complementary benefits from a broad set of parallelization, memory optimization, and communication–computation overlap techniques. These sets of parallelization techniques can be data parallel, model parallel, or activation parallel algorithms that have evolved over time. \textit{Data parallelism} replicates the model across accelerators (e.g., GPU, TPU, NPU) and distributes input data across them, while \textit{model parallelism} distributes the model parameters across accelerators. Another form of parallelism, \textit{activation parallelism}, distributes the intermediate activations across accelerators, usually along the sequence dimension. Since a typical parallelization of a large model involves using a combination of these, an in-depth understanding of individual parallelization strategies as well as comparative benefits and trade-offs is critically important for both researchers and practitioners.

Complementary methods for efficient scaling of deep learning workloads address the memory and communication bottlenecks stemming from large sequences, large model parameters, and synchronization among distributed workers. Memory optimization techniques play an important role in fitting activations for large input sequences and parameters of large models within hardware memory constraints. Collective communication is required to synchronize computation on data subsets or computations from subsets of model parameters. Hence, computation–communication overlap plays an important role in reducing latency by minimizing wait time for communication to be completed.

To develop a comprehensive foundation for efficient distributed deep learning, this section first introduces the core concepts of collective communication operations. It then examines prominent strategies for parallelization, memory optimization, and communication–computation overlap, and concludes with a comparative analysis of these approaches.

\subsection{Collective Operations}
\label{sec:collective_ops}
Collective communication primitives are fundamental to distributed training, enabling synchronization and data exchange across multiple devices or ranks. Below we summarize the most widely used operations.

\textbf{Reduce.}
The \emph{Reduce} operation aggregates data from multiple ranks and delivers the result to a designated root rank. Typical reduction functions include but not limited to summation, product, minimum and maximum. 


\textbf{Gather.}
The \emph{Gather} operation collects data from all participating ranks and delivers it to a single root rank. Unlike Reduce, the data is simply gathered rather than aggregated.

\begin{wrapfigure}{r}{0.71\textwidth} 
        \centering
        \resizebox{\linewidth}{!}{%
        \begin{tikzpicture}[>=Stealth, font=\small]
        \definecolor{sliceA}{RGB}{255,153,153} 
        \definecolor{sliceB}{RGB}{153,204,255} 
        \definecolor{sliceC}{RGB}{153,255,153} 
        \definecolor{sliceD}{RGB}{255,229,153} 
        
        \node at (8.5,7.5) {\textbf{All-Gather}};
        
        \node at (0.5,7) {Rank 0};
        \node at (2.5,7) {Rank 1};
        \node at (4.5,7) {Rank 2};
        \node at (6.5,7) {Rank 3};
        
        \draw[gray!70] (0,5.5) rectangle (1,6.5);
        \fill[sliceA] (0,5.5) rectangle (1,6.5);
        \node[black] at (0.5,6) {a};
        
        \draw[gray!70] (2,4.5) rectangle (3,5.5);
        \fill[sliceB] (2,4.5) rectangle (3,5.5);
        \node[black] at (2.5,5) {b};
        
        \draw[gray!70] (4,3.5) rectangle (5,4.5);
        \fill[sliceC] (4,3.5) rectangle (5,4.5);
        \node[black] at (4.5,4) {c};
        
        \draw[gray!70] (6,2.5) rectangle (7,3.5);
        \fill[sliceD] (6,2.5) rectangle (7,3.5);
        \node[black] at (6.5,3) {d};
        
        \draw[->, ultra thick] (7.5,4.5) -- (9.5,4.5);
        
        \node at (10.5,7) {Rank 0};
        \node at (12.5,7) {Rank 1};
        \node at (14.5,7) {Rank 2};
        \node at (16.5,7) {Rank 3};
        
        \foreach \x in {10,12,14,16} {
          \draw[gray!70] (\x,2.5) rectangle (\x+1,6.5);
        }
        \fill[sliceA] (10,5.5) rectangle (11,6.5); \node[black] at (10.5,6) {a};
        \fill[sliceB] (10,4.5) rectangle (11,5.5); \node[black] at (10.5,5) {b};
        \fill[sliceC] (10,3.5) rectangle (11,4.5); \node[black] at (10.5,4) {c};
        \fill[sliceD] (10,2.5) rectangle (11,3.5); \node[black] at (10.5,3) {d};
        
        \fill[sliceA] (12,5.5) rectangle (13,6.5); \node[black] at (12.5,6) {a};
        \fill[sliceB] (12,4.5) rectangle (13,5.5); \node[black] at (12.5,5) {b};
        \fill[sliceC] (12,3.5) rectangle (13,4.5); \node[black] at (12.5,4) {c};
        \fill[sliceD] (12,2.5) rectangle (13,3.5); \node[black] at (12.5,3) {d};
        
        \fill[sliceA] (14,5.5) rectangle (15,6.5); \node[black] at (14.5,6) {a};
        \fill[sliceB] (14,4.5) rectangle (15,5.5); \node[black] at (14.5,5) {b};
        \fill[sliceC] (14,3.5) rectangle (15,4.5); \node[black] at (14.5,4) {c};
        \fill[sliceD] (14,2.5) rectangle (15,3.5); \node[black] at (14.5,3) {d};
        
        \fill[sliceA] (16,5.5) rectangle (17,6.5); \node[black] at (16.5,6) {a};
        \fill[sliceB] (16,4.5) rectangle (17,5.5); \node[black] at (16.5,5) {b};
        \fill[sliceC] (16,3.5) rectangle (17,4.5); \node[black] at (16.5,4) {c};
        \fill[sliceD] (16,2.5) rectangle (17,3.5); \node[black] at (16.5,3) {d};
        
        \end{tikzpicture}
        }
        \caption{All-Gather operation: before (left) each rank holds only its reduced slice; after (right) each rank holds all slices.}
\label{fig:all-gather-wrap}
\end{wrapfigure}
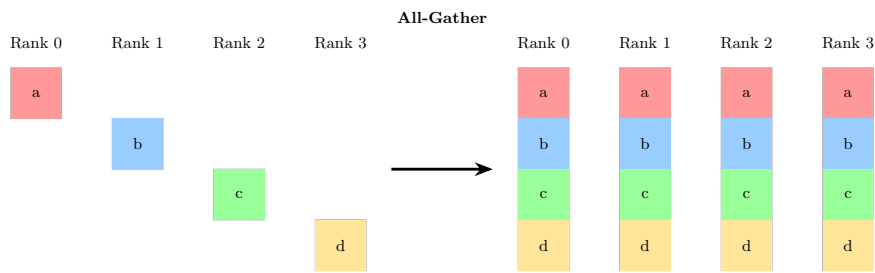

\textbf{AllGather.}
The \emph{AllGather} operation extends Gather by collecting data from every rank. When a group of $N$ devices (ranks) participate in this collective communication with each containing data of a $\texttt{tensor\_size}$, after this operation each of the ranks in the group will have data of size $N\times \texttt{tensor\_size}$.  
This differs from Gather, where only the root rank receives the result. A visual depiction of the \emph{AllGather} operation for four ranks is shown in Figure~\ref{fig:all-gather-wrap}. In this example, each of the four ranks initially contains a distinct data element: rank~$0$ holds $a$, rank~$1$ holds $b$, rank~$2$ holds $c$, and rank~$3$ holds $d$. After the \emph{All-Gather} operation, all ranks receive the complete set of data elements aggregated across all ranks. As a result, each rank contains the full sequence $(a, b, c, d)$.

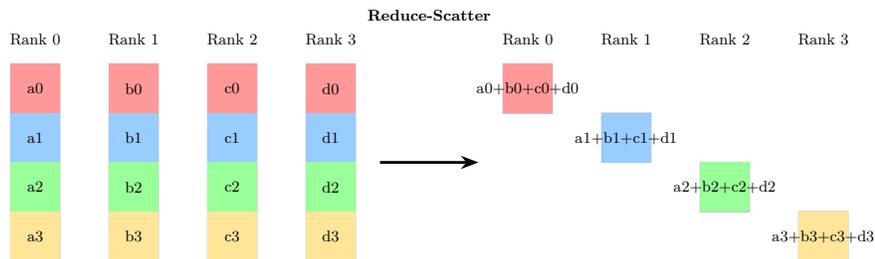
\begin{wrapfigure}{r}{0.71\textwidth} 
\centering
        \resizebox{\linewidth}{!}{%
        \begin{tikzpicture}[>=Stealth, font=\small]
        
        \definecolor{sliceA}{RGB}{255,153,153} 
        \definecolor{sliceB}{RGB}{153,204,255} 
        \definecolor{sliceC}{RGB}{153,255,153} 
        \definecolor{sliceD}{RGB}{255,229,153} 
        
        \node at (8.5,7.5) {\textbf{Reduce-Scatter}};
        
        \node at (0.5,7) {Rank 0};
        \node at (2.5,7) {Rank 1};
        \node at (4.5,7) {Rank 2};
        \node at (6.5,7) {Rank 3};
        
        \foreach \x in {0,2,4,6} {
          \draw[gray!70] (\x,2.5) rectangle (\x+1,6.5);
        }
        
        \fill[sliceA] (0,5.5) rectangle (1,6.5); \node[black] at (0.5,6) {a0};
        \fill[sliceB] (0,4.5) rectangle (1,5.5); \node[black] at (0.5,5) {a1};
        \fill[sliceC] (0,3.5) rectangle (1,4.5); \node[black] at (0.5,4) {a2};
        \fill[sliceD] (0,2.5) rectangle (1,3.5); \node[black] at (0.5,3) {a3};
        
        \fill[sliceA] (2,5.5) rectangle (3,6.5); \node[black] at (2.5,6) {b0};
        \fill[sliceB] (2,4.5) rectangle (3,5.5); \node[black] at (2.5,5) {b1};
        \fill[sliceC] (2,3.5) rectangle (3,4.5); \node[black] at (2.5,4) {b2};
        \fill[sliceD] (2,2.5) rectangle (3,3.5); \node[black] at (2.5,3) {b3};
        
        \fill[sliceA] (4,5.5) rectangle (5,6.5); \node[black] at (4.5,6) {c0};
        \fill[sliceB] (4,4.5) rectangle (5,5.5); \node[black] at (4.5,5) {c1};
        \fill[sliceC] (4,3.5) rectangle (5,4.5); \node[black] at (4.5,4) {c2};
        \fill[sliceD] (4,2.5) rectangle (5,3.5); \node[black] at (4.5,3) {c3};
        
        \fill[sliceA] (6,5.5) rectangle (7,6.5); \node[black] at (6.5,6) {d0};
        \fill[sliceB] (6,4.5) rectangle (7,5.5); \node[black] at (6.5,5) {d1};
        \fill[sliceC] (6,3.5) rectangle (7,4.5); \node[black] at (6.5,4) {d2};
        \fill[sliceD] (6,2.5) rectangle (7,3.5); \node[black] at (6.5,3) {d3};
        
        \draw[->, ultra thick] (7.5,4.5) -- (9.5,4.5);
        
        \node at (10.5,7) {Rank 0};
        \node at (12.5,7) {Rank 1};
        \node at (14.5,7) {Rank 2};
        \node at (16.5,7) {Rank 3};
        
        \draw[gray!70] (10,5.5) rectangle (11,6.5);
        \fill[sliceA] (10,5.5) rectangle (11,6.5);
        \node[black] at (10.5,6) {a0+b0+c0+d0};
        
        \draw[gray!70] (12,4.5) rectangle (13,5.5);
        \fill[sliceB] (12,4.5) rectangle (13,5.5);
        \node[black] at (12.5,5) {a1+b1+c1+d1};
        
        \draw[gray!70] (14,3.5) rectangle (15,4.5);
        \fill[sliceC] (14,3.5) rectangle (15,4.5);
        \node[black] at (14.5,4) {a2+b2+c2+d2};
        
        \draw[gray!70] (16,2.5) rectangle (17,3.5);
        \fill[sliceD] (16,2.5) rectangle (17,3.5);
        \node[black] at (16.5,3) {a3+b3+c3+d3};
        
        \end{tikzpicture}
        }
        \caption{Reduce-Scatter operation: before (left) each rank holds all slices; after (right) each rank holds the reduced slice.}
\label{fig:reduce-scatter-wrap}
\end{wrapfigure}

\textbf{ReduceScatter.}
The \emph{ReduceScatter} operation first performs a reduction across ranks (as in Reduce) and then evenly scatters the reduced result across all ranks. Each rank thus receives only the portion of the output corresponding to its index, reducing communication overhead compared to AllReduce. An example of the reduce-scatter is visualized in Figure~\ref{fig:reduce-scatter-wrap}. In this example, the four ranks initially contain the data sequences 
$(a_0, a_1, a_2, a_3)$, $(b_0, b_1, b_2, b_3)$, $(c_0, c_1, c_2, c_3)$, and $(d_0, d_1, d_2, d_3)$, respectively. After the \emph{ReduceScatter} operation, the data is reduced via sum aggregation, and each rank receives a slice of the aggregated sequence corresponding to its index. As a result, rank~$0$ contains the first slice of the reduced (aggregated) result, which is $(a_0 + b_0 + c_0 + d_0)$. The other ranks receive similar slices.

\textbf{AllReduce.}
The \emph{AllReduce} operation performs a reduction (e.g., sum, min, max) across all ranks and distributes the final result back to every rank. Conceptually, AllReduce is equivalent to performing a Reduce followed by a Broadcast. However, this naive strategy creates a bottleneck at the root server. In practice, optimized implementations use the \emph{Ring-AllReduce} algorithm, in which data is first partitioned and reduced in a distributed manner (ReduceScatter), followed by redistribution of the reduced blocks (AllGather). High-performance systems, such as IBM’s BlueConnect, adopt this approach~\cite{cho2019blueconnect}. We consider a similar example of \emph{AllReduce} with four ranks containing initial data sequences 
$(a_0, a_1, a_2, a_3)$, $(b_0, b_1, b_2, b_3)$, $(c_0, c_1, c_2, c_3)$, and $(d_0, d_1, d_2, d_3)$, respectively . 
In such case, the data in the ranks $0, 1, 2, 3$ after the \emph{ReduceScatter} operation becomes $(a_0 + b_0 + c_0 + d_0)$, $(a_1 + b_1 + c_1 + d_1)$, $(a_2+ b_2 + c_2 + d_2)$, and $(a_3 + b_3 + c_3 + d_3)$. A following \emph{All-Gather} operation on these reduced slices will result in all rank containing all the slices, which is the complete reduced sequence:
$(a_0 + b_0 + c_0 + d_0$, $a_1 + b_1 + c_1 + d_1$, $a_2 + b_2 + c_2 + d_2$, $a_3 + b_3 + c_3 + d_3)$. This result corresponds to the output of \emph{All-Reduce}. We omit including a visual example for \emph{All-Reduce} for brevity, since its behavior is quite understandable from the visual examples of \emph{ReduceScatter} and \emph{All-Gather}.

\textbf{All-to-All.}
The \emph{All-to-All} operation (also known as total exchange or personalized communication) enables each rank to send a distinct message to every other rank, including itself. At the end of the operation, each rank holds one piece of data from every participant. A common use case is in Mixture-of-Experts (MoE) training, where tokens must be dynamically routed to experts across ranks via one All-to-All exchange, and then results are routed back with another~\cite{shazeer2017outrageously}.

A comparative summary of the volume of data communication per device for the aforementioned collective operations are presented in Table~\ref{tab:collective-cost}. For \textit{Reduce} and \textit{Gather}, each rank sends its tensor to a designated root, resulting in a communication cost of $(N-1)\times\texttt{tensor\_size}$. In contrast, ring-based collectives such as \textit{AllGather} and \textit{ReduceScatter} evenly distribute communication across all devices, thus reducing the per-device cost to $\frac{(N-1)\times\texttt{tensor\_size}}{N}$ for a group of $N$ devices. This balanced communication pattern improves scalability by mitigating bottlenecks and making more efficient use of available bandwidth. More complex collectives are often constructed from these primitives; for example, \textit{AllReduce} is commonly implemented as a \textit{ReduceScatter} followed by an \textit{AllGather}, resulting in a total cost of $2\times\frac{(N-1)\times\texttt{tensor\_size}}{N}$. 


\begin{table}[ht]
\centering
\caption{Data movement per device for common collective operations. $N$ is the number of ranks and \texttt{tensor\_size} is the size of the tensor in each rank before the collective operation.}
\label{tab:collective-cost}
\begin{tabular}{lc}
\toprule
\textbf{Collective Operation} & \textbf{Data Moved Per Device} \\
\midrule
Reduce & $(N-1) \times \text{tensor\_size}$ \\
Gather & $(N-1) \times \text{tensor\_size}$ \\
Ring AllGather & $\frac{(N-1) \times \text{tensor\_size}}{N}$ \\
Ring ReduceScatter & $\frac{(N-1) \times \text{tensor\_size}}{N}$ \\
All-to-All & $\frac{(N-1) \times \text{tensor\_size}}{N}$ \\
AllReduce (ReduceScatter + Ring AllGather) & $2 \times \frac{(N-1) \times \text{tensor\_size}}{N}$ \\
\bottomrule
\end{tabular}
\end{table}

\subsection{Data Parallelism}
\label{sec:data_parallel}

Data parallelism (DP) is the most common parallel training strategy for deep neural networks \cite{goyal2017accurate, li2020pytorch}. This approach splits the entire mini-batch across multiple devices, where each device executes a single replica of the model and communicates with each for synchronization at the end of each training step. This synchronization requires an all-reduce communication among the gradients of all ranks within the DP group. Enabling more devices to train large minibatches with this strategy is easier. Several recommendations from the literature suggest that training speed varies almost linearly with the number of devices \cite{smith2017don18,you2017large,goyal2017accurate,you2018imagenet}. However, arbitrarily increasing data parallelism beyond a critical batch size may yield diminishing returns or even negatively impact the convergence~\cite{shuai2024scaling}.
Furthermore, data parallelism improves throughput by distributing different inputs across devices, but it does not reduce latency since each device must still process the full forward and backward pass of a single input independently and sequentially. Thus, more efforts are needed to ensure the convergence performance \cite{keskar2016large,bian2021maximizing}. In addition, replicating the weights of large transformers is limited by the memory available per device.

\subsection{Model Parallelism}
\label{sec:parallelization}
As transformer models scale towards trillions of parameters, model parallelism is required to distribute
model parameters, activations, and optimizer state across devices for them to fit into device memory
and be trainable in a realistic amount of time. In general, model parallelism reduces the number of parameters stored on each device approximately linearly with the parallelism size. For example, the number of parameters per device is halved when the model parallel size is doubled. Different forms of model parallelism require distinct communication patterns, and the efficiency of these patterns depends heavily on connectivity bandwidth and cluster topology. These factors play a critical role in determining the trade-off considerations for various model parallel techniques and their combinations  \cite{zhuang2023optimizing}. In the following, we discuss different model parallelism techniques commonly used for LLM training and inference.

\textbf{Pipeline Parallelism (PP)}

Pipeline Parallelism (PP) distributes consecutive layers as pipeline stages across multiple GPUs and each GPU processes a different stage of the network sequentially \cite{narayanan2019pipedream}. The pipelining of stages can increase the throughput by pipelining multiple micro batches although it does not help in improving latency of a single batch. However, the sequential dependency of the pipeline stages introduces idle time in the pipeline which is referred as \textit{bubbles}. These \textit{bubbles} causes efficiency issues, and various pipeline scheduling algorithms have emerged to address the bubble issue. Popular instances of pipeline scheduling algorithms are \textit{GPipe} \cite{huang2019gpipe}, \textit{PipeDream} \cite{narayanan2019pipedream}, TeraPipe \cite{li2021terapipe}, \textit{Zero Bubble 1F1B}~\cite{qi2023zero}, and \textit{DualPipe} \cite{wang2025review}. 

In \textit{GPipe}, a mini-training batch is split into smaller microbatches for which a group of deep neural network layers is executed \cite{huang2019gpipe}. This approach caused some pipeline bubbles between stages. \textit{PipeDream} improves upon \textit{GPipe} by eliminating pipeline bubbles through an interleaved scheduling strategy, known as 1 Forward 1 Backward (1F1B)~\cite{narayanan2019pipedream}. In this 1F1B pipeline scheduling, each pipeline stage alternates between executing a forward pass for a new micro-batch and a backward pass for an earlier one. This ensures that all devices remain fully utilized once the pipeline is warm, unlike \textit{GPipe}’s schedule which incurs idle periods at the start and end of a batch. To maintain correctness, \textit{PipeDream} introduces \textit{weight versioning}, storing multiple copies of the parameters so that the backward pass of each micro-batch uses the same weight version as its forward pass. While this design significantly increases training throughput, it does so at the cost of additional memory overhead from maintaining multiple weight copies, as well as scheduling complexity. To address this memory increase, works in \cite{chen2018efficient} and \cite{narayanan2021memory} address this issue through weight prediction, and a novel weight update scheme, respectively. Another following pipeline scheduling algorithm, \textit{TeraPipe}, introduced a novel form of pipelining tailored to single-transformer architectures by performing pipelining across tokens rather than micro-batches \cite{li2021terapipe}. Despite these advances, there still remains the pipeline bubble issue. Conversely, Zero Bubble 1F1B proposed an idea to independently schedule the backward computation of two parts, backward for input and backward for gradient, which allow for a pipeline schedule with minimal bubbles. Further replacing the before-hand global state synchronization with a post validation, this algorithm allows to bypass synchronizations during the optimizer step and can result in zero bubbles in the pipeline \cite{qi2023zero}. Most recently, DeepSeek extended the Zero Bubble 1F1B schedule with bidirectional pipeline scheduling to introduce the DualPipe algorithm, which achieves greater computation–communication overlap and further reduces pipeline \textit{bubbles} \cite{liu2024deepseek}.

\textbf{Tensor Parallelism (TP)}


Tensor Parallelism (TP) enables the training of large models by distributing model parameters across multiple GPUs, thereby reducing the computational workload per device \cite{shoeybi2019megatron}. In contrast to PP, which partitions the model vertically into sequential stages distributed across devices, TP partitions the parameters of projection layers and distributes them across devices. Specifically, matrices involved in matrix multiplications are divided either column-wise or row-wise depending on the operation, followed by a communication step to synchronize partial results. More concretely, TP requires collective operations, such as \texttt{all-reduce}, to synchronize activations during the forward pass and gradients during the backward pass. While TP is effective to accommodate extremely large models by sharding the weight matrices, the associated communication overhead can become a significant bottleneck in bandwidth limited scenarios. Because of bandwidth constraints, a common practice is to use TP among GPUs within a single host node. Furthermore, increasing the degree of TP too high leads to too small matrix multiplications, which can be less efficient and may reduce overall resource utilization. Research efforts have addressed in these limitation leading to various communication overhead reduction techniques \cite{wang2022tesseract,bian2021maximizing,shazeer2018mesh,chen2016training}, further discussed in Section ~\ref{sec:comm_overlap}. 



Beyond this standard one-dimensional tensor parallelism (1D TP), several extensions have been proposed to improve scalability and communication efficiency. Two-dimensional tensor parallelism (2D TP) partitions tensors along both row and column dimensions to reduce communication overhead~\cite{xu2023efficient}. Building on this, two-and-a-half-dimensional tensor parallelism (2.5D TP) strikes a balance between memory usage and communication cost by combining aspects of 1-D and 2-D schemes~\cite{wang2022tesseract}. It is termed as 2.5D as the algorithm contains special cases of both 2D and 3D matrix multiplication. Conversely, three-dimensional tensor parallelism (3D TP) generalizes these approaches by adopting a 3D parallel matrix multiplication approach, enabling efficient training of extremely large models on massive GPU clusters~\cite{bian2021maximizing}. While these higher-dimensional schemes provide improved trade-offs between communication cost, memory efficiency, and scalability, one-dimensional tensor parallelism remains the most widely adopted form of TP in practice due to its simplicity and broad framework support. Therefore, we primarily focus on 1-D TP throughout this paper, and unless otherwise specified, TP would refer to the generic 1-D TP approach.

\textbf{Expert Parallelism in Mixture of Experts (MoE)}

Expert Parallelism (EP) is a parallelization strategy designed specifically for Mixture-of-Experts (MoE) architectures. The evolution of MoE from early works such as GShard~\cite{lepikhin2020gshard} and Switch Transformer~\cite{fedus2022switch} to modern implementations including Mixtral~\cite{jiang2024mixtral} and DeepSeek~\cite{dai2024deepseekmoe} reflects the ongoing trend of leveraging sparse gating and distributed computation to scale large language models efficiently. In MoE architectures, a set of smaller multilayer perceptron (MLP) networks, referred to as experts, replaces a single large MLP. Tokens are dynamically routed to a sparse subset of experts~\cite{shazeer2017outrageously}, which reduces computation per token while enabling massive overall model capacity and scalable training.

In expert parallelism, these small experts are distributed across multiple GPUs  and tokens are dynamically routed to the appropriate experts across GPUs \cite{liu2022gating}.
For example, in DeepSeek-MoE-16B, the model includes 64 routed experts and 2 shared experts~\cite{dai2024deepseekmoe}. When EP=16, each GPU holds 4 routed experts and the 2 shared experts, totaling 6 experts per device. DeepSeek uses device-limited routing to reduce communication overhead—ensuring that the top-k experts selected for a token are located on a limited number of devices. This reduces cross-device traffic without significantly hurting model performance. For balanced utilization, the MoE router must distribute tokens evenly across experts to prevent scenarios where some experts are overused and others are starved. This is often achieved using load balancing losses or bias-based methods. An all-to-all collective communication is performed before the MoE layer to route tokens to the designated GPUs where their corresponding activated experts reside. A second all-to-all communication is then performed at the end of the MoE layer to reorganize the tokens into their appropriate sequential order and redistribute them across GPUs according to the parallelization strategy of the succeeding layer.

\subsection{Activation Parallelization}
\label{sec:activation_parallel}
In LLM training, a large memory space is needed to store the input and output activations of the network layers. NeMo Framework provides effective activation distribution methods, which is critical in training LLM with a large sequence length or large per-GPU micro-batch size.
\cite{korthikanti2023reducing} \cite{brakel2024modelSurvey}

\textbf{Sequence Parallelism (SP)}

Sequence Parallelism (SP) is a parallelism technique introduced to efficiently train large language models on long input sequences. First formally presented in ACL 2023 by \cite{korthikanti2023reducing}, SP addresses the memory bottlenecks encountered in traditional parallelism methods when handling long contexts. Unlike data parallelism, which replicates input batches across devices, sequence parallelism partitions the input along the sequence dimension, allowing activations to be distributed across devices while keeping the full model replicated. This enables more efficient memory utilization without altering the model’s architecture.

In practice, SP divides each input sequence into chunks that are processed in parallel across devices.  Although SP still replicates model parameters and optimizer states across devices, similar to data parallelism, its unique ability to scale with sequence length makes it particularly useful for long-context training \cite{li2021sequence}. To maximize efficiency and scalability, SP is typically used in conjunction with tensor parallelism (TP), pipeline parallelism (PP), and data parallelism (DP).

Modern implementations include Megatron’s Tensor-Parallel Sequence Parallelism (TPSP), DeepSpeed-Ulysses’ Sequence Parallelism (UP, a.k.a. TPUP) \cite{jacobs2023deepspeed}, and Ring-Attention Context Parallelism (CP)—all designed to extend SP for training increasingly large models with longer context windows.







\textbf{Context Parallelism (CP)}


Context parallelism (CP) is a system optimization technique that improves the latency and scalability of LLM inference, particularly for long contexts. Without modifying the underlying dense attention algorithms, CP offers several advantages for long-context LLM inference \cite{yang2024context}. For example, it is reported to take around 60 seconds to serve 128K context length or around 1200 seconds to serve 1M context length for Llama3 405B model with a single H100 GPU host of 8 GPUs making a suitable application scenario for CP~\cite{yang2024context}. 

CP uses sequence parallelization within attention layer. The input sequence is split into multiple chunks along the sequence dimension, with each chunk handled by a different GPU or processor. A ring communication is used to transfer Q or KV chunks between GPUs through point-to-point (P2P) communication. The communication happens in a ring network where each rank communicates with its predecessor and successor \cite{liu2023ring}. 

To tackle the scalability and efficiency issue of sequence parallelism, LoongTrain proposed a 2D-Attention mechanism termed as Double-Ring-Attention~\cite{gu2024loongtrain}. In particular, it combines head-parallel of DeepSpeed UP~\cite{jacobs2023deepspeed} and context-parallel techniques of Ring Attention~\cite{liu2023ring} in a 2D parallelism fashion to improve scalability, p2p communication efficiency, and better compute-communication overlap. 

\textbf{Linear Attention Sequence Parallelism (LASP)}

Modern generative AI architectures often combine traditional softmax attention with linear attention to handle longer context lengths efficiently. To accelerate both training and inference of such models, two versions of Linear Attention Sequence Parallelism (LASP) have been proposed \cite{sun2024linear, sun2025lasp}. LASP-1 introduces a P2P ring-style communication scheme for both forward and backward passes across devices, either within a node or across multiple nodes. This approach optimizes the use of right-product kernel tricks in linear attention by exchanging only a single intermediate state, rather than both key and value states as in alternative methods. LASP-2 further improves communication and computation parallelism for training linear attention transformers on very long sequences. Unlike the first version, LASP-2 requires only a single AllGather operation on intermediate memory states, whose size is independent of sequence length, reducing communication overhead and enabling better overlap between computation and communication. This version has also been extended to support hybrid model architectures that encompass both attention and Mamba blocks.



\subsection{Memory Optimization Techniques}
\label{sec:memory_optimization}
Training and inference of large language models (LLMs) are limited not only by compute throughput but also by GPU memory capacity. As models scale to hundreds of billions of parameters, memory reduction strategies become essential. Three widely used approaches are activation checkpointing, redundancy elimination, and memory offloading, often integrated into modern training frameworks.  

\paragraph{Activation Checkpointing.} 
Checkpointing reduces activation memory by storing only a subset of intermediates during the forward pass and recomputing the rest during backpropagation~\cite{chen2016training}. Optimized scheduling methods such as Checkmate~\cite{jain2020checkmate} and Tempo~\cite{andoorveedu2022tempo} balance recomputation cost with memory savings, enabling significantly larger models within fixed budgets. 

\paragraph{Gradient Release.} 
Gradient release is a layer‑wise execution strategy to achieve constant memory usage~\cite{pudipeddi2020training,optimigithub} by fusing backward computation and optimizer step. This approach compute gradient and take optimizer step layer by layer enabling early release of gradient memory of each layer and hence gradient of full model is not realized at once. This allows for efficient gradient handling and memory savings during large‑scale neural network training.

\paragraph{Redundancy Elimination (ZeRO).} 
Data-parallel training typically replicates optimizer states, gradients, and parameters across devices. ZeRO~\cite{rajbhandari2020zero} partitions these components, with ZeRO-1 sharding optimizer states, ZeRO-2 extending to gradients, and ZeRO-3 extending to parameters. Though later stages increase communication overhead, ZeRO underpins large-scale frameworks such as DeepSpeed and has enabled trillion-parameter training.  

\paragraph{Memory Offloading.} 
Offloading transfers activations or model states from GPU to CPU or NVMe storage, effectively extending memory at the cost of latency. Early work (vDNN~\cite{rhu2016vdnn}, SuperNeurons~\cite{wang2018superneurons}) explored static/dynamic policies, while later systems (Capuchin~\cite{peng2020capuchin}, Mobius~\cite{feng2023mobius}) improved efficiency via adaptive scheduling and compression. ProTrain addresses this by autonomously optimizing memory, computation, and IO for efficient, expert-free training. Recently, ProTrain introduces an automated training system that coordinates memory, computation, and IO, achieving faster performance than existing frameworks without manual tuning~\cite{yang2024protrain}.

\paragraph{System-Level Integration.} 
Frameworks such as DeepSpeed, Colossal-AI, and Megatron-LM combine checkpointing, ZeRO, and offloading in unified systems. Recent work like FPDT~\cite{yao2024training} addresses fragmentation and inefficient allocation to further improve effective utilization.  

\paragraph{Performance Implications.} 
Beyond memory savings, these techniques can reduce communication overhead by lowering tensor or pipeline parallelism requirements, and increase batch size for higher kernel efficiency, thereby directly improving training throughput.

\subsection{Communication Overlap Techniques}
\label{sec:comm_overlap}



Co-optimization and overlap of computation and communication have become critical techniques for improving scalability and system efficiency of distributed deep learning. A major challenge in achieving such overlap is data dependency, which can be categorized into \textit{weak} and \textit{strong} forms. Weak data dependency occurs when the computation of one component can proceed concurrently with the communication associated with a different component. A common example is the all-reduce of gradients in data parallelism, where gradient synchronization for one mini-batch can be overlapped with the forward computation of another mini-batch. In contrast, strong data dependency arises when the next computation step requires strict completion of a communication phase. For instance, in Mixture-of-Experts (MoE) architectures, the all-to-all routing of token embeddings must fully complete before expert computation. The feasibility of overlapping communication with computation depends directly on the type of dependency involved as strong dependencies introduce strict synchronization barriers and substantially limit overlap, often necessitating more sophisticated algorithms. To address these challenges, recent research proposes combined computation–communication algorithms that explicitly account for dependency patterns and employ fused kernels to reduce synchronization overhead. 
This section will begin with the communication volume comparison for different parallelization and then summarize communication optimization techniques.

\textbf{Communication Volume Comparison.}
Different parallelization techniques require distinct collective communication patterns aligned with the chosen parallelization strategies. Table~\ref{tab:parallelism-comm} reports the 
{communication volume and collective operations} per device for the forward pass and the attention layer:  

\begin{itemize}
    \item \textbf{DP:} No communication in the forward pass; gradient synchronization requires 
    an all-reduce in the backward pass.  
    \item \textbf{TP:} Communication volume scales as $\tfrac{TP-1}{TP}$, implemented with 
    ring all-reduce collectives. For attention in Table~\ref{tab:parallelism-comm}, the communication cost is $bs \frac{d}{TP}$ over $(TP - 1)$ links, multiplied by four. The factor of four accounts for two contributions from reduce-scatter and two from all-gather, since each collective operation involves both sending and receiving the tensor.
    \item \textbf{CP:} Less communication volume than TP and typically realized with 
    point-to-point or ring structures. Due to the limited bandwidth of inter-node communication, CP helps TP in inter-node communication.The factor of four in Table~\ref{tab:parallelism-comm} arises from two contributions for the key tensor and two for the value tensor, with each tensor involving both a send and a receive operation.
    \item \textbf{TPSP:} Requires both reduce-scatter and all-gather collectives, with volume 
    growing alongside the sequence partitioning factor. The attention's communication volume for TPSP is the same as TP. 
    \item \textbf{TPUP:} Requires four all-to-all exchanges attention layer, for query, key, value and output. The communication scales as 
    $\tfrac{TPUP-1}{TPUP^2}$, which implies less communication volume per device for increasing number of GPUs. Here, we show the breakdown of communication volume for query, key, value, and output. 
        
    \[
    \text{Comm}_\text{query} = 2 \cdot \frac{s}{TPUP}\cdot \frac{ad_h}{TPUP} \cdot (TPUP-1) 
    \]

     \[
    \text{Comm}_\text{key} = 2 \cdot \frac{s}{TPUP}\cdot \frac{kd_h}{TPUP} \cdot (TPUP-1) 
    \]
    
     \[
    \text{Comm}_\text{value} = 2 \cdot \frac{s}{TPUP}\cdot \frac{kd_h}{TPUP} \cdot (TPUP-1) 
    \]
    
     \[
    \text{Comm}_\text{output} = 2 \cdot \frac{s}{TPUP}\cdot \frac{ad_h}{TPUP} \cdot (TPUP-1) 
    \]
    Hence, the total communication volume for each attention layer is the sum of these four components as shown in Table~\ref{tab:parallelism-comm}. By adding the components, we get the following:
    \[
    \text{Comm}_\text{attention} = 4 \frac{(TPUP-1)}{TPUP^2} \cdot bsd_h (a+k)
    \]

\end{itemize}

Overall, these results highlight distinct trade-offs: DP minimizes communication but maximizes 
weight redundancy, TP and TPSP provide better memory efficiency at the cost of collective 
communication overhead, and TPUP achieves compute and activation savings but incurs the 
highest communication complexity. CP distributed activation memory and computation which is beneficial for ultra-long context. 

\begin{table*}[htbp]
\centering
\caption{Communication volume and collective operations per attention layer per device for different parallelism strategies during forward pass. Here, $b$ is the batch size, $s$ is the sequence length, $d$ is the transformer hidden dimension, $k$ is number of key-value heads, $d_h$ is head dim.}
\label{tab:parallelism-comm}
\begingroup
\renewcommand{\arraystretch}{1.5} 

\begin{tabular}{lp{4cm}p{8cm}}
\toprule
\textbf{Parallelism} & \textbf{Communication Volume} & \textbf{Collective Operation} \\
\midrule
DP & $0$ & None in forward pass \\
TP & $\frac{4(TP-1)}{TP} \cdot bsd$ & Ring AllReduce (ReduceScatter + Ring AllGather) \\
CP & $\frac{4(CP-1)}{CP} \cdot bskd_h$ & Point-to-Point / Ring  ($kd_h < d$ for GQA and 2 for each of send \& recv)\\
TPSP & $\frac{4(TPSP-1)}{TPSP} \cdot bsd$ & Ring AllGather / Ring ReduceScatter \\
TPUP & $4 \frac{(TPUP-1)}{TPUP^2} \cdot bsd_h (a+k)$ & All-to-All (4 per layer) \\
\bottomrule
\end{tabular}
\endgroup
\end{table*}

\paragraph{Data Decomposition.}
To alleviate the overhead stemming from inter-device communication, most commonly used computation-communication overlap strategy employ data decomposition techniques. This type of methods partition the data into smaller slices and then interleave the computation of one slice of the data to the computation communication of another slice using asynchronous communication. In this manner, computation and communication can proceed concurrently to reduce idle time of computing unit and improve overall resource utilization. This type of approaches are widely supported in large-scale AI infrastructure frameworks, including \textit{Nvidia Megatron} and \textit{Huawei MindSpeed}. In Megatron-LM, this approach is used for all-reduce in tensor parallelism(TP)~\cite{shoeybi2019megatron} which is subsequently extended for all-gather and reduce-scatter in Tensor Parallelism with Sequence Parallelism (TPSP) ~\cite{korthikanti2023reducing}. Similarly, this type of merged computation-communication approach have been applied in the training of \textit{Pangu Ultra} on Ascend NPUs~\cite{yin2025pangu}. Nevertheless, despite the advantages of such decomposition techniques, residual communication overhead persists. Specifically, while intermediate chunks benefit from overlapping communication with computation, the final chunk incurs unavoidable latency, often referred to as \textit{tail overhead}. This limitation underscores the need for further optimization in balancing communication and computation in large-scale distributed training.

\paragraph{Communication Algorithm Decomposition.} 
Collective communication decompose based algorithms decompose the collective operation into steps of different communication pattern, such as P2P, while decoupling the interdependency of the steps. This decomposition of algorithm into different communication pattern allows to loosen the strong data dependency in a manner to allow better overlap possibility than data decomposition based approaches. Notable that the decomposed algorithm also use data decompose (e.g., slicing) and schedule the computation and communication jointly for a better overlap than the approaches that decompose only the data.  Example approach is the Decompose algorithm used in TPU platform ~\cite{wang2022overlap}. A graph transformation is used for transforming the fine grained computation to semantically equivalent graph with decoupled asynchronous instruction scheduling that allow higher overlapping. This approach breaks down blocking collective communication into a sequence of single-step, non-blocking collectives. It also decomposes computation operations into finer-grained tasks that can be overlapped, then combines them in a simple sequence to accumulate partial results.

\paragraph{Kernel Fusion.}
Central to the computation-communication overlap is efficient kernel fusion where both the computation and communication of a layer is implemented in a single kernel utilizing some overlap strategy. The fused kernel may implement data decomposition or communication decompose based overlap approach in a single kernel. Early work on kernel fusion include CoCoNet~\cite{jangda2022breaking} which provided a domain specific language(DSL). They developed a compiler-based system that facilitates generating co-optimized custom computation and communication kernels with hardware specific optimized implementations for target specific topology, and data sizes. Followingly, various kernel frameworks and libraries has evolved over time~\cite{li2023automated,chang2024flux}. A recent framework, Concerto, presented a compiler framework that can automatically optimize and schedule communication along with auto-decomposition to create overlap opportunity for critical communication   ~\cite{cheng2025concerto}. Another recent framework is TileLink, which addresses the challenges of operator decomposition and kernel fusion in terms of performance and complexity by decoupling communication and computation and linking them through tile-centric primitives~\cite{zheng2025tilelink}. Multi-gpu kernels are also emerging as a promising direction. A recent framework in this direction is Parallel Kittens that allows to easily write fast computation-communication overlapped multi-GPU kernels supporting data, tensor, sequence, and expert parallelism \cite{sul2025parallelkittens}.


\paragraph{Communication Optimization.}
Another direction uses communication optimization integrated with parallelization strategy rank assignments and work scheduling and automated tuning of communication parameters. One example of such approach is implemented in \textit{MiCS} that introduces a data-parallel training system to minimizes communication overhead by reducing communication scale through scale-aware partitioning and hierarchical communication ~\cite{zhang2022mics}. Another similar example is AlpaComm~\cite{zhuang2023optimizing} which proposed a pipelining schedule named \textit{eager-1F1B} extending the \textit{1F1B}~\cite{narayanan2019pipedream} and developed communication library, AlpaComm, that supports integration to the parallelization library Alpa~\cite{zheng2022alpa}.

\paragraph{Automated tuning.}
Automated tuning based approaches rely on automated search of optimal low-level performance sensitive parameters for communication. An example of such approach is
AutoCCL, which implemented automated tuning method on top of NCCL to search for optimal low-level performance sensitive parameters using a divide-and-conquer algorithm~\cite{xu2025autoccl}.

\section{Hybrid Systems}
\label{sec:hybrid_systems}

Training and inference of large scale language models at trillion-parameter scale requires more than a single form of parallelism, as strategy brings unique advantages and limits when used in isolation. To overcome these bottlenecks, hybrid approaches has become common practice in industry, often referred to as multi-parallelism. This section discusses most common multi-parallelism strategies, focusing on 3D and 4D parallelization, and popular frameworks that enable their practical deployment.

\subsection{Multi-Parallelism}
\label{sec:multiparallelism}

Different parallelization strategies offer distinct advantages by distributing weight memory, activation memory, and computation across groups of devices. Efficient training and inference of large-scale models must address challenges in memory capacity, bandwidth, and compute requirements, while meeting task-specific goals such as high throughput or low latency. These challenges are best addressed through tailored combinations of strategies that balance trade-offs. Comparative analyses show that hybrid parallelism achieves greater speedups than single-device training~\cite{baligodugula2025optimizing}, and recent large-scale models further validate the effectiveness of hybrid approaches~\cite{grattafiori2024llama,yang2024context}.



\begin{figure}[t]
    \centering
    \resizebox{0.84\linewidth}{!}{%
    \begin{tikzpicture}[gpu/.style={draw, fill=gray!10, rounded corners=2pt,
                  minimum width=2.5cm, minimum height=1.6cm,
                  font=\scriptsize, align=center},
    ]

    \def\tpgap{2.5}   
    \def\ppgap{3.0}   
    \def\dpgap{7.5}  

    \pgfmathsetmacro{\xoffset}{(1*\tpgap + 1*\dpgap)/2-0.5}
    \pgfmathsetmacro{\yoffset}{(1*\ppgap)/2+5.5}

    \node[font=\bfseries] at (0,9.5) {3D Parallelism using (DP, PP, TP)};

    \node[anchor=north] at (0,9.2) {
    \begin{tabular}{@{}ll@{\hskip 0.5cm}ll@{\hskip 0.5cm}ll@{}}
        \tikz{\draw[blue, thick] (0,0) -- +(0.5cm,0);} & Data Parallelism (DP) & 
        \tikz{\draw[green!60!black, thick] (0,0) -- +(0.5cm,0);} & Pipeline Parallelism (PP) &
        \tikz{\draw[magenta, thick] (0,0) -- +(0.5cm,0);} & Tensor Parallelism (TP) 
      \end{tabular}
    };
    \centering
    \foreach \DP in {0,1} {
      \foreach \PP in {0,1} {
          \foreach \TP in {0,1} {
            \pgfmathsetmacro{\x}{\TP * \tpgap + \DP * \dpgap - \xoffset}
            \pgfmathsetmacro{\y}{-( \PP * \ppgap) + \yoffset}
            \pgfmathsetmacro{\x}{\x + (0.6*\PP==0 ? 0.3 : 0)}
            \pgfmathtruncatemacro{\G}{\DP*4 + \PP*2 +  \TP}
            \node[gpu] (gpu\G) at (\x,\y) {
              {\textcolor{magenta}{TP: \TP}}\\
              {\textcolor{green!60!black}{PP: \PP}}\\
              {\textcolor{blue}{DP: \DP}}\\
              GPU: \G
            };
          }
      }
    }

    \node[draw=green!60!black, thick, fit=(gpu0)(gpu1),
          outer sep=0.1cm] (dp0pp0box) {};
    \node[green!60!black, anchor=north] at (dp0pp0box.south) {DP0 PP0};

    \node[draw=green!60!black, thick, fit=(gpu2)(gpu3),
          outer sep=0.1cm] (dp0pp1box) {};
    \node[green!60!black, anchor=north] at (dp0pp1box.south) {DP0 PP1};

    \node[draw=green!60!black, thick, fit=(gpu4)(gpu5),
          outer sep=0.1cm] (dp1pp0box) {};
    \node[green!60!black, anchor=north] at (dp1pp0box.south) {DP1 PP0};

    \node[draw=green!60!black, thick, fit=(gpu6)(gpu7),
          outer sep=0.1cm] (dp1pp1box) {};
    \node[green!60!black, anchor=north] at (dp1pp1box.south) {DP1 PP1};

    \node[draw=blue, thick, fit=(dp0pp0box)(dp0pp1box),
          inner sep=0.5cm, label={[blue]below:DP Group 0}] {};
    \node[draw=blue, thick, fit=(dp1pp0box)(dp1pp1box),
          inner sep=0.5cm, label={[blue]below:DP Group 1}] {};

    \end{tikzpicture}
    } 
    \caption{Illustration of 3D parallelism showing DP, PP, and TP groupings for 8 GPUs with (DP,PP,TP) as (2,2,2). The ranks assignments for parallelization groups here are DP: [[0,1,2,3], [4,5,6,7]]; PP: [[0,2], [1,3], [4,6], [5,7]]; TP: [[0,1], [2,3], [4,5], [6,7]].}
    \label{fig:multiparallel-3d}
 \end{figure}
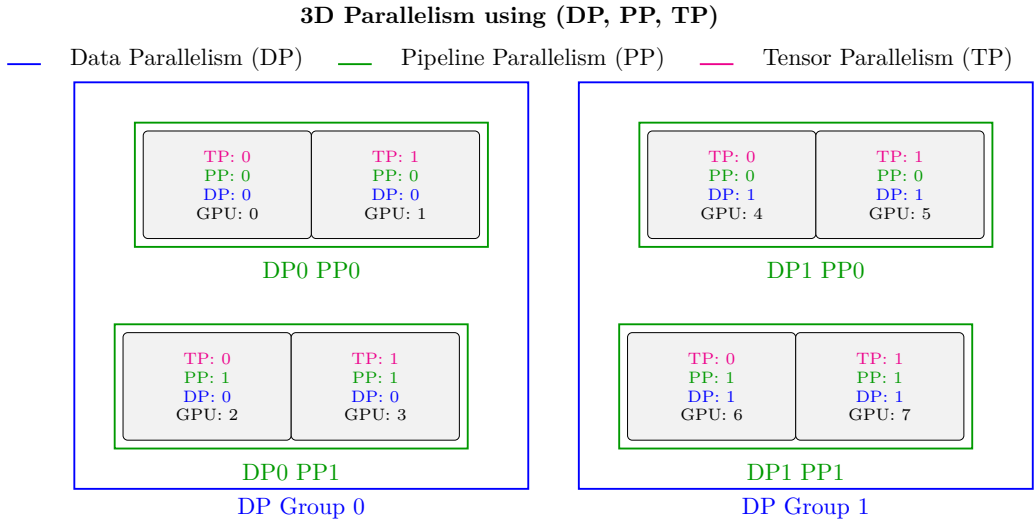

\paragraph{3D Multi-Parallelism.}
3D multi-parallelism combines data parallelism (DP), pipeline parallelism (PP), and tensor parallelism (TP) by assigning GPUs to these respective parallel groups~\cite{rasley2020deepspeed}. For a total of $N$ GPUs, with $N_{DP}$, $N_{PP}$, and $N_{TP}$ GPUs allocated to DP, PP, and TP groups respectively, the relationship is given by $ N_{DP} \times N_{PP} \times N_{TP} = N$. The combination of multi-parallelism is designed to optimize resource allocation, e.g., allocating accelerators, to achieve an optimal trade-off that integrates the complementary benefits of different parallelization strategies. Integrating these three orthogonal strategies with an optimal rank group assignment enables efficient training of models with hundreds of billions to trillions of parameters that would otherwise exceed the capacity of single parallelization techniques. We present an example of 3D multi-parallelism in Figure~\ref{fig:multiparallel-3d}. This example of 3D multi-parallelism considers $N=8$ GPUs with GPU ranks $0,1,..., 7$ and the parallelization strategy considers $(N_{DP},N_{PP},N_{TP})=(2,2,2)$. This $(2,2,2)$ grouping of 8 GPUs is for example purpose only, any combination that meets $ N_{DP} \times N_{PP} \times N_{TP} = N$ is valid where the grouping strategy is an important concern for parallelization design, detailed explanation is presented in Section \ref{sec:system_design}. Hence, the most outer parallelization is DP with the DP group containing two replica of PP,TP groups with GPU ranks $0,...,3$ and ranks $4,...,7$. Within each DP group, there are two identical PP groups.  Each PP group contains a TP group of two GPUs. Corresponding collective communication operation happens within the GPU ranks of each parallelization groups.

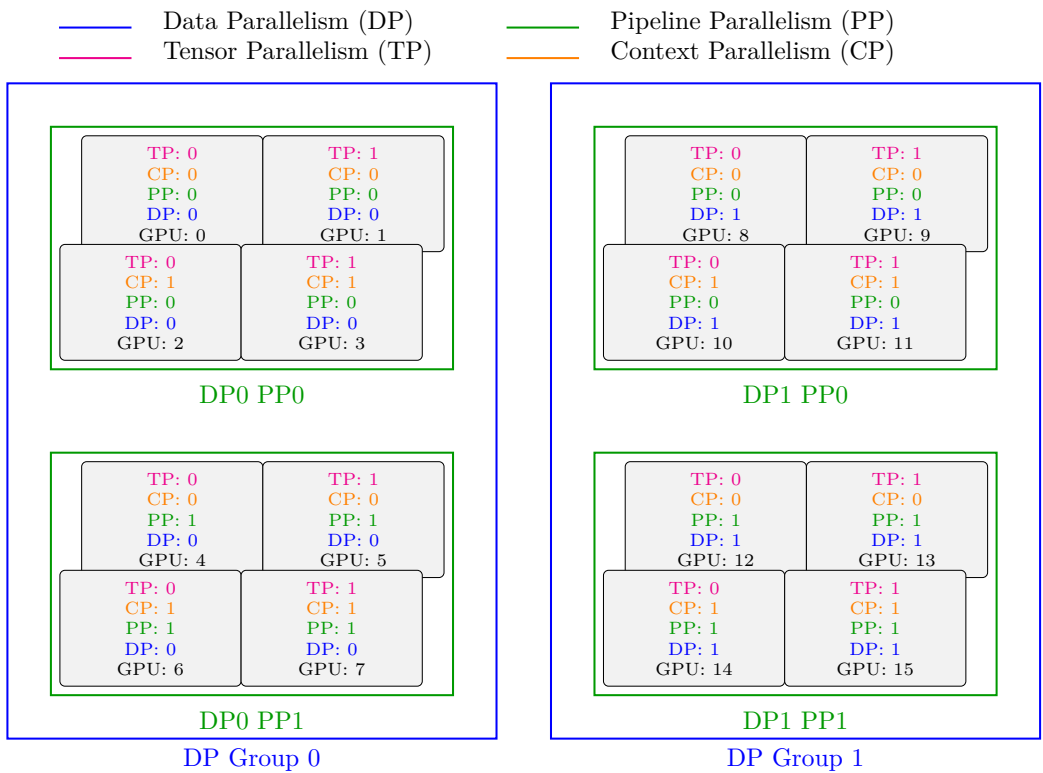
\begin{figure}[t]
    \centering
    \resizebox{0.84\linewidth}{!}{%
    \begin{tikzpicture}[gpu/.style={draw, fill=gray!10, rounded corners=2pt,
                  minimum width=2.5cm, minimum height=1.6cm,
                  font=\scriptsize, align=center},
    ]

    \def\tpgap{2.5}   
    \def\cpgap{1.5}   
    \def\ppgap{4.5}   
    \def\dpgap{7.5}  

    \pgfmathsetmacro{\xoffset}{(1*\tpgap + 1*\dpgap)/2-0.5}
    \pgfmathsetmacro{\yoffset}{(1*\cpgap + 1*\ppgap)/2+3.5}

    \node[font=\bfseries] at (0,10) {4D Parallelism using (DP, PP, TP, CP)};

    \node[anchor=north] at (0,9.2) {
      \begin{tabular}{@{}ll@{\hskip 1cm}ll@{}}
        \tikz{\draw[blue, thick] (0,0) -- +(1cm,0);} & Data Parallelism (DP) & 
        \tikz{\draw[green!60!black, thick] (0,0) -- +(1cm,0);} & Pipeline Parallelism (PP) \\
        \tikz{\draw[magenta, thick] (0,0) -- +(1cm,0);} & Tensor Parallelism (TP) & 
        \tikz{\draw[orange, thick] (0,0) -- +(1cm,0);} & Context Parallelism (CP)
      \end{tabular}
    };

    \foreach \DP in {0,1} {
      \foreach \PP in {0,1} {
        \foreach \CP in {0,1} {
          \foreach \TP in {0,1} {
            \pgfmathsetmacro{\x}{\TP * \tpgap + \DP * \dpgap - \xoffset}
            \pgfmathsetmacro{\y}{-(\CP * \cpgap + \PP * \ppgap) + \yoffset}
            \pgfmathsetmacro{\x}{\x + (0.6*\CP==0 ? 0.3 : 0)}
            \pgfmathtruncatemacro{\G}{\DP*8 + \PP*4 + \CP*2 + \TP}
            \node[gpu] (gpu\G) at (\x,\y) {
              {\textcolor{magenta}{TP: \TP}}\\
              {\textcolor{orange}{CP: \CP}}\\
              {\textcolor{green!60!black}{PP: \PP}}\\
              {\textcolor{blue}{DP: \DP}}\\
              GPU: \G
            };
          }
        }
      }
    }

    \node[draw=green!60!black, thick, fit=(gpu0)(gpu1)(gpu2)(gpu3),
          outer sep=0.1cm] (dp0pp0box) {};
    \node[green!60!black, anchor=north] at (dp0pp0box.south) {DP0 PP0};

    \node[draw=green!60!black, thick, fit=(gpu4)(gpu5)(gpu6)(gpu7),
          outer sep=0.1cm] (dp0pp1box) {};
    \node[green!60!black, anchor=north] at (dp0pp1box.south) {DP0 PP1};

    \node[draw=green!60!black, thick, fit=(gpu8)(gpu9)(gpu10)(gpu11),
          outer sep=0.1cm] (dp1pp0box) {};
    \node[green!60!black, anchor=north] at (dp1pp0box.south) {DP1 PP0};

    \node[draw=green!60!black, thick, fit=(gpu12)(gpu13)(gpu14)(gpu15),
          outer sep=0.1cm] (dp1pp1box) {};
    \node[green!60!black, anchor=north] at (dp1pp1box.south) {DP1 PP1};

    \node[draw=blue, thick, fit=(dp0pp0box)(dp0pp1box),
          inner sep=0.5cm, label={[blue]below:DP Group 0}] {};
    \node[draw=blue, thick, fit=(dp1pp0box)(dp1pp1box),
          inner sep=0.5cm, label={[blue]below:DP Group 1}] {};

    \end{tikzpicture}
    } 
    \caption{Illustration of 4D parallelism showing DP, PP, TP, and CP groupings for a cluster of 16 GPUs with (DP,PP,TP,CP) as (2,2,2,2). The ranks assignments for parallelization groups here are DP: [[0,1,2,3,4,5,6,7], [8,9,10,11,12,13,14,15]]; PP: [[0,1,2,3], [4,5,6,7],[8,9,10,11], [12,13,14,15]]; CP: [[0,2], [1,3],...]; TP: [[0,1], [2,3],...].}
    \label{fig:multiparallel-4d}
 \end{figure}

\paragraph{4D Multi-parallelism}
4D multi-parallelism extends 3D multi-parallelism by adding context parallelism (CP) as another orthogonal parallelization strategy~\cite{grattafiori2024llama}. For a total of $N$ GPUs, with $N_{DP}$, $N_{PP}$, $N_{TP}$, and $N_{CP}$ GPUs allocated to DP, PP, TP, and CP groups respectively, the relationship is given by $ N_{DP} \times N_{PP} \times N_{TP} \times N_{CP} = N$.  This extension to 4D parallelism includes CP to support for ultra-long sequences with extremely large models. Since the parallelization ranks assigned to different GPUs are more sophisticated compared to 3D parallelism, we also present an example of 4D multi-parallelism in Figure~\ref{fig:multiparallel-4d}. This example of 4D multi-parallelism considers $N=16$ GPUs with GPU ranks $0,1,..., 15$. The parallelization strategy considers $(N_{DP},N_{PP},N_{TP},N_{CP})=(2,2,2,2)$. Similar to the example of 3D parallelism, here also we present this $(2, 2,2,2)$ grouping of 16 GPUs in favor of simplicity and for example purpose only, any combination that meets $ N_{DP} \times N_{PP} \times N_{TP} \times N_{CP} = N$ is valid as long as it meets other constraints, e.g., memory of individual GPUs. Hence, the most outer parallelization is DP with the DP group containing two replica of PP, TP, CP groups with GPU ranks $0,...,7$ and ranks $8,...,15$. Within each DP group, there are two identical PP groups, e.g., GPU ranks $0,...,3$ and $4,...,7$ within the first DP group.  Each PP group contains a TP,CP group of four GPUs. For example the TP groups within the first PP group of the first DP group is $0,1$ and $2,3$ where CP group is $0,2$ and $1,3$. This rank assignment considers the outer to inner ordering of the parallelization strategies used. Corresponding communication operation happens within the GPU ranks of each parallelization groups.

\paragraph{Communication Optimization in Multi-parallelism.}
In a distributed setting, communication can become a bottleneck when compute devices need to wait for certain dependent communication to be complete. To address this challenge, recent hybrid algorithms also focus one overlapping collective communications, such as reduce-scatter, all-gather and all-reduce , discussed in detail in Section~\ref{sec:comm_overlap}. When multiple parallelization strategies are combined in a hybrid parallelization for large scale models, it brings challenges of multiple collective communication patterns apearing at different part of forward and backward computation graph. Centauri~\cite{chen2024centauri} introduces a training system that enhances training efficiency and communication overlap in LLM training by partitioning collective communications into intra-node and inter-node group communications. It employs a hybrid parallelization scheduling approach that optimizes overlapping across operation, layer, and model levels. At the operation level, coordinating communication among partitioned groups facilitates fine-grained overlapping, while adaptive scheduling and model-level optimization maximize overlap across forward, backward, and model update phases in hybrid parallel configurations. For PP, traditional solution is to use a Virtual Pipeline (VP) which uses a virtual pipeline of generally 2 stages within each pipeline stages. This allows for the communication of pipeline stages of the half of the micro-batch to happen while the other half is doing computation. Prominent libraries, such as Megatron-LM uses compute-communication overlap strategies for reduce-scatter and all-gather by chunking the data and overlap compute and communication of independent chunks~\cite{megatronlm2025github}.

\subsection{Frameworks}
\label{sec:frameworks}
The foundational work on distributed training of large models emphasized parallelism strategies to partition model computation across multiple devices. Table~\ref{tab:parallelframeworks} presents a shortlist of most notable industry-grade frameworks.  

\paragraph{Megatron-LM and NeMo.} 
Megatron-LM pioneered tensor-model parallelism by splitting weight matrices across GPUs, later combining this with pipeline parallelism to achieve scalability for trillion-parameter models \cite{shoeybi2019megatron}. Megatron-LM and NeMo are two popular framework supporting multi-parallelism. Megatron-LM is widely recognized as the backbone of trillion-parameter LLMs such as GPT-3 and BLOOM, offering comprehensive support for data, tensor, pipeline, sequence, and expert parallelism. Its strength lies in enabling hybrid combinations of these strategies, allowing fine-grained profiling and tuning across GPU clusters to maximize throughput and memory efficiency. However, its complexity and reliance on NVIDIA hardware can limit accessibility for non-GPU ecosystems. NeMo complements Megatron-LM by focusing on multimodal and speech models, supporting tensor, pipeline, and sequence parallelism.

\paragraph{DeepSpeed.}
DeepSpeed focused on memory-footprint reduction with multi parallelism. It introduced the Zero Redundancy Optimizer (ZeRO) and gradient partitioning to minimize memory overhead, alongside mixed-precision training for efficiency at scale \cite{rasley2020deepspeed}. DeepSpeed supports data, tensor, and pipeline parallelism, with robust expert parallelism (MoE) capabilities, making it highly effective for hybrid parallelization strategies. Its profiling and tuning tools are tightly integrated with PyTorch, simplifying adoption across research and enterprise environments.

\paragraph{MindSpeed-LLM.}
MindSpeed-LLM presents a distributed training framework tailored for the Ascend AI platform~\cite{mindspeedllm,liao2021ascend}. It supports large-scale language model development through multi-parallelism strategies such as tensor, pipeline, sequence, and expert parallelism. By integrating these dimensions, MindSpeed-LLM enables efficient 3D and 4D parallel training optimized for Ascend NPUs. This platform also include MindSpeed-RL~\cite{feng2025mindspeed} for LLM post-training and Mind-VL~\cite{chen2025mindvl} for multi-modal LLMs.

\paragraph{SageMaker and Neuron.}
Amazon SageMaker Model Parallelism provided a cloud-native implementation of tensor, pipeline, and data parallelism, making large-scale model training accessible in industrial environments \cite{karakus2021amazon}. This framework presented an effective solution for distributed training and inference, even though it require to manually configure the combination of different parallelization groups. The Neuron SDK presents suitable system for custom hardware chips, supporting tensor, sequence, and data parallelism. Neuron excels in profiling and hardware-aware tuning, but expert parallelism remains constrained.

\paragraph{Miscellaneous.}
PaddlePaddle is a full-stack deep learning framework with support for data and model parallelism with distributed training and provides a rich ecosystem of domain-specific libraries for vision, language, and scientific applications \cite{ao2021end}. It extended memory optimization direction by incorporating hybrid parallelism that allow sharding memory across diverse hardware environments. It is widely used as more of a general-purpose platform than a specialized LLM training engine. In another direction of adopting multiple parallelization strategies and dynamic transition among them based on workload includes dynamic hot switching~\cite{ge2024enabling} and resharding~\cite{suseesaw}. HotSPa introduced a dynamic hot switching approach for training which apply different parallelism strategies within a single mini-batch by using their graph compiler and hot switch planner~\cite{ge2024enabling}. The graph compiler generates distributed computation graph while the hot switch planner heuristically determines a plan to accelerate between a pair of parallelism strategies. Seesaw, on the other hand, introduced an LLM inference engine that allows for reconfiguration of parallelization strategies from prefil to decode~\cite{suseesaw}. The method is termed as dynamic model resharding which adopts different parallelization to address unique workload challenges of prefil and decode.

Together, these frameworks illustrate the diversity of approaches to hybrid parallelism beyond the industry leaders. They serve as important complements and experimental testbeds, advancing ideas in profiling, tuning, and workload balancing that may eventually influence the mainstream frameworks dominating large-scale LLM training. Despite the richer feature sets, current distributed frameworks still require manual configuration of parallelization and memory-optimization strategies, leaving model developers with a complex search space to achieve peak efficiency.


\begin{table}[h]
\centering
\caption{Industry-grade frameworks supporting multi-parallelism (DP, TP, PP, SP, EP/MoE)}
\resizebox{\linewidth}{!}{
\begin{tabular}{|l|l|c|c|c|c|c|}
\hline
\textbf{Framework} & \textbf{Developer Company} & \textbf{DP} & \textbf{TP} & \textbf{PP} & \textbf{SP} & \textbf{EP/MoE} \\ \hline
Megatron-LM~\cite{shoeybi2019megatron} & NVIDIA & $\checkmark$ & $\checkmark$ & $\checkmark$ & $\checkmark$ & $\checkmark$ \\ \hline
MindSpeed-LLM~\cite{mindspeedllm} & Huawei & $\checkmark$ & $\checkmark$ & $\checkmark$ & $\checkmark$ & $\checkmark$ \\ \hline
DeepSpeed~\cite{rasley2020deepspeed} & Microsoft & $\checkmark$ & $\checkmark$ & $\checkmark$ & \textit{Limited} & $\checkmark$ \\ \hline
Amazon SageMaker~\cite{karakus2021amazon} & Amazon Web Services (AWS) & $\checkmark$ & $\checkmark$ & $\checkmark$ & \textit{Limited} & \textit{Limited} \\ \hline
AWS Neuron~\cite{awsneuronsdk} & Amazon Web Services (AWS) & $\checkmark$ & $\checkmark$ & $\checkmark$ & $\checkmark$ & \textit{Limited} \\ \hline
NVIDIA NeMo~\cite{nvidianemosdk} & NVIDIA & $\checkmark$ & $\checkmark$ & $\checkmark$ & $\checkmark$ & \textit{Limited} \\ \hline
\end{tabular}
}
\label{tab:parallelframeworks}
\end{table}

\section{Workload Parallelization Design}
\label{sec:system_design}

\subsection{System Design Implications} 
\label{sec:system_design_impl}

An optimized parallelization system design for large scale models, e.g., LLMs, training and inference has profound implications for scalability and efficiency. System level optimization makes the scaling to large scale models and training them on a large cluster and deployment on inference system efficient. Designing such a parallelization system for large scale models requires addressing the distinct  requirements for training and inference system optimization as discussed below. 


\paragraph{Commonly Used Metrics.} Before delving deeper into system design implications, it is due to introduce some commonly used metrics and terms in describing AI workload performance characteristics.
\bigskip

\textit{Time To First Token (TTFT).} This metric measures how quickly the model's output generation starts after providing a query~\cite{nvidiadocs4metrics}. Low waiting times for a response are essential in real-time interactions, such as chat applications. This metric is driven by the time required to process the prompt and then generate the first output token. Hence, TTFT is primarily determined by prefill time (i.e., prompt processing and attention state initialization), while system overhead (including tokenization, detokenization, networking, scheduling, and framework delays) constitutes secondary factors.

\[
TTFT \approx T_{\text{prefill}} + T_{\text{system\_overhead}}
\]

\textit{Time Per Output Token (TPOT).} This metric measures the time to generate each output token after the prefill time. This metric is also known as \textit{Inter-Token Latency (ITL)}~\cite{nvidiadocs4metrics}. This metric corresponds with how each user will perceive the "speed" of the model. TPOT (a.k.a. ITL) is define below:

\[
TPOT = \frac{ e2e\_latency - TTFT}{ Total\_output\_tokens - 1}
\]

\textit{Latency.}  The overall time it takes for the model to generate the full response for a query.  Overall response latency can be calculated as below:  

\[
\text{Latency} = TTFT + TPOT \cdot (Total\_output\_tokens - 1)
\]

\textit{Throughput.}  
Throughput in large language models (LLMs) is the rate of tokens or samples processed per unit time. This is also known as Token Per Second (TPS)~\cite{nvidiadocs4metrics}. During training it measures hardware efficiency as the number of training samples or tokens per second, while during inference it measures responsiveness as the number of output tokens per second across users. In both cases, it counts the number of tokens per second.
Formally,

\[
\text{Training Throughput} = \frac{\text{Total training tokens}}{\text{Training time}}
\]

\[
\text{Inference Throughput} = \frac{\text{Total output  tokens}}{\text{inference time}}
\]

 \textit{Arithmetic Intensity.} The \textit{Arithmetic Intensity} is defined as the ratio of total flops of an operation to total memory access. It is a characteristics of an operator. 
\begin{align}
\text{Arithmetic Intensity (A.I.)} = \frac{\text{Total FLOPs}}{\text{Total Bytes Transferred}}
\end{align}

\textit{Hardware Ridge Point.} The \textit{Hardware Ridge Point} is defined as as hardware characteristics with the ratio of peak flops to peak memory bandwidth. 

\begin{align}
\text{Ridge Point (I)} = \frac{\text{Peak FLOPs}}{\text{Peak Bandwidth}}
\end{align}

\[
\text{Operation is }
\begin{cases}
\textbf{Compute-bound}, & \text{if } \text{A.I.} > I, \\[6pt]
\textbf{Memory-bound},  & \text{if } \text{A.I.} < I.
\end{cases}
\]

\textit{Compute Bound.} These operations does not need to wait for memory access. 

\textit{Memory Bound.} These operations are bottle-necked by memory access and computing unit needs to wait for memory access to be completed. Hence computing capability of the hardware is idle during that time. 



\textit{Bandwidth Bound.} The \textit{Bandwidth Bound} scenario also can happen in distributed scenario when the distributed operator needs to wait for communication to be be completed for synchronization purposes. This also leads to idle time in computing unit. 

\textit{Model Flops Utilization.} The \textit{Model Flops Utilization} (MFU) have been used in the literature as a measure of Flops utilization~\cite{chowdhery2023palm} and its unit is percentage.

\begin{align}
\label{eq:mfu}
\text{Model FLOPs Utilization (MFU)} = 100.0 \cdot \frac{\text{Achieved FLOPs}}{\text{Peak FLOPs}}
\end{align}

Most of the operators or computations of transformer based models have higher arithmetic intensity during training and prefil phase. As a result, they are mostly compute bound and hence can utilize hardware peak flop capacity leading to high MFU. On the other hand, most of the operations of transformer models in the decode phase of LLMs have low arithmetic intensity and often memory bound resulting to low MFU. In general, high MFU results in high throughput. So, system optimization needs to consider individual stage workload characteristics to achieve optimal result in target metrics. 


\paragraph{Problem Formulation.}
The problem of hybrid parallelization system design can be formally defined as allocating computing accelerators of a given cluster across a set of parallelization strategies in order to optimize overall system performance while satisfying memory constraints and meeting service level objectives, such as maximizing throughput, minimizing latency and ensuring the highest accuracy. So, given a Model $M$, cluster $W$, and a set of parallelization strategies $S_l$ for layer $l$, the high level formulation of the problem is, 
\begin{align}
\text{optimize} \quad & \text{ServiceLevelObjective}(M,W,S_l) \nonumber\\
\text{subject to} \quad & \prod_{i=1}^{n} w_i = W & \quad \forall i \in S_l
\end{align}
where $S_l = \{DP,PP, TP,...\}$. We omit the detailed attributes of $M$ and $W$, e.g., model parameters and cluster configurations here in favor of simplicity. The model attributes to consider may include model parameter, depth, hidden size, layer type etc. The cluster configuration may include the memory capacity of each node, peak flops, memory access bandwidth, cluster topology, intra-node and inter-node bandwidth.  For example, we may consider a system using a cluster  $W$ with $N_w$ accelerators and employing $S_l = \{DP,PP,TP,CP \} $ as a $4D$ parallelization strategy for training a transformer model. Let $w_i$ denote the number of accelerators allocated to parallelization strategy $i$. Then, the total number of accelerators satisfies the relation $w_{DP}\times w_{PP} \times w_{TP} \times w_{CP} = W$ while meeting the memory constraints and maximizing the throughput and maintaining operators accuracy. 


The service level objective is generally different for training and inference with managing memory and communication overhead being a shared criteria. For training, the primary concern is how fast the training can be completed while inference requires rapid model execution with reduced computational overhead to deliver fast responses. In other words, training necessitates high throughput as a service level objective. This demands strategies that optimize computational load distribution and minimize synchronization delays. In contrast, time for individual queries matter during inference, as users expect quick replies. This necessitates prioritizing low latency and minimal memory footprint, often requiring a trade-off with throughput. Although achieving both high throughput and low latency is ideal, trade-offs are common. For instance, lower latency can imply higher throughput, but the reverse is not always true. An inference system may need to use a large batch size to increase throughput; however, this can result in higher latency. Therefore, latency is often prioritized when making an optimal trade-off with throughput, particularly in interactive chat based applications~\cite{thoppilan2022lamda}. The optimal partitioning strategy also changes when trading off between latency and throughput in addition to the model size, sequence length and cluster size considerations~\cite{pope2023efficiently}.

For example, the \textit{ServiceLevelObjective}$(M,W,S_l)$ may be to minimize latency for inference and maximize throughput for training. A \textit{Service Level Objective (SLO)} also defines the target performance level for a particular metric, setting the standard for what is considered acceptable service quality \cite{bentoml2025metrics}. For instance, an SLO for \textit{Time To First Token (TTFT)} might specify TTFT below $T_{thres}$ milliseconds. SLOs are typically part of a broader \textit{Service-Level Agreement (SLA)} between a provider and its users. Mathematically, the SLO for training can be, 

\[
P(\text{TPS}) \geq \tau 
\]
which implies that the throughput for training should be at least $\tau$. 
 
Similarly, the SLO for the inference is often expressed as percentile constraints on performance metrics \cite{ding2019characterizing,nastic2020sloc}:

\[
P_{q}(M) \leq T
\]

where \(P_{q}(M)\) is the \(q\)-th percentile of metric \(M\) (e.g., latency), and \(T\) is the threshold. A particular instance can be,

\[
P_{95}(\text{TTFT}) < 200 \, \text{ms}
\]

 meaning that 95\% of requests must have a Time To First Token under 200 milliseconds.

\paragraph{Parallelization Assignment.}
The assignment of accelerator to optimize service level objective conventionally considers grid search based approach. This type of approaches begin with considering search space of various combination of parallelization strategies spanning the accelerators on the given cluster. Then they employ the roofline model as a manual analytical framework to assess and compare parallelization strategies. By mapping arithmetic intensity against hardware ceilings such as memory bandwidth and peak floating-point throughput, the roofline model enables researchers to determine whether workloads are compute-bound or bandwidth-bound, guiding choices among data, model, pipeline, and hybrid parallelization approaches~\cite{williams2009roofline,yang2021hierarchical}. Extensions such as multi-dimensional and distributed roofline models have further supported performance analysis of deep learning workloads, offering insights into tensor operations, communication costs, and scaling inefficiencies across heterogeneous clusters~\cite{miao2022md}. More recently, roofline-based analysis has been applied to large language models (LLMs), highlighting its role in forecasting inference latency, training efficiency, and communication overheads in distributed GPU clusters ~\cite{kundu2024performance,imai2024predicting,yuan2024llm}.

Pruning of the search tree is a traditional way of optimizing the grid search. The criteria used to prune the search space includes memory constrains and cluster topology for communication bottleneck consideration for parallelization strategies. Despite the emergence of automated ML-driven predictors, the roofline model remains a foundational manual approach for workload characterization and parallelization design in LLM training and inference. While this conventional manual approach has been foundational, the future of workload design is increasingly optimistic, with emerging auto-parallelization strategies promising to automate bottleneck detection and parallelization decisions at scale. We include a comprehensive discussion on emerging auto-parallelism strategies in Section~\ref{sec:auto_parallel}.



Major challenges in hybrid parallelization for large-scale AI models requires careful optimization of communication overhead, computation overlap, memory efficiency, and bandwidth constraints to balance training throughput and inference latency. The inter-node bandwidth of the cluster topology often limits the TPSP 
group size. DP requires optimized gradient synchronization necessiating homogeneous bandwidth among the participating ranks. CP can be efficient when there is a good overlap of the ring communication with computation. 
The choice of TPSP, 
DP, and PP ranks directly impacts performance, requiring careful placement to align with intra-node and inter-node bandwidth constraints. Again the choice of number of accelerators in each parallelization is another important factor to performance. A well-designed hybrid system strategically integrates these parallelization methods to maximize scalability, throughput, and low-latency inference performance for AI workloads.

\subsection{System Design Considerations}
\label{sec:system_design_cons}



\paragraph{Parallelization Trade-Off Considerations.}
\label{sec:comparative Analysis}
The primary consideration of multi-parallelism strategies depends on the complementary benefits and trade-offs across different forms of parallelization. Different parallelization strategies offer unique advantages making their integration into a hybrid system to consider for complementary benefits while optimizing for specific service level demand, e.g., training or inference performance. The effectiveness of multi-parallelism lies in balancing these trade-offs.  Toward this end, we summarize them to establish the foundation for a systematic comparative analysis of parallelization strategies. A curated summary of complementary benefits and trade-offs of prominent parallelization strategies are presented in Table~\ref{tab:parallelism-pros-cons}.
 The analysis includes summary on scalability by splitting data across devices (DP), reducing memory and compute per device by distributing model parameters (TP), and pipeline micro-batches to improve throughput (PP). Corresponding to the current discussion of the benefits, DP and PP directly impacts throughput, TP/TPSP impacts latency, PP, TP, CP impacts memory. 
 
By strategically combining these parallelization methods with an optimal resource allocation to each group, a hybrid approach can achieve an optimal balance between compute distribution, memory efficiency and communication overhead reduction. As a result, systems can be optimized for a scalable and high-performance AI systems for specific service level demands. To complement the summary of DP, TP, and CP, we also provide a theoretical validation by deriving the FLOPs, memory consumption (including both weight and activation memory), and communication overhead for the GQA, MLP, and Mamba blocks in Section~\ref{sec:theory_parallel_strategies}.

\begin{table*}[t]
\centering
\caption{Pros and cons of different distributed parallelism strategies.}
\label{tab:parallelism-pros-cons}
\begin{tabular}{|l|p{6cm}|p{6cm}|}
\hline
\textbf{Strategy} & \textbf{Pros} & \textbf{Cons} \\
\hline
Data Parallel (DP) & 
\begin{itemize}[leftmargin=*]
  \item Simple to implement
  \item Boosts throughput by distributing training data across devices
  \item Minimal communication during forward/backward pass
\end{itemize} &
\begin{itemize}[leftmargin=*]
  \item Requires full model replica on each device
  \item High gradient synchronization overhead
  \item Inefficient memory usage for very large models
\end{itemize} \\
\hline
Tensor Parallel (TP) & 
\begin{itemize}[leftmargin=*]
  \item Splits weight and activation memory across devices
  \item Enables training of very large models
  \item High compute efficiency when scaling within a node
\end{itemize} &
\begin{itemize}[leftmargin=*]
  \item Heavy communication overhead across layers
  \item Best efficiency requires high bandwidth connectivity (e.g., NVLink, Infiniband)
  \item Harder to implement compared to DP
\end{itemize} \\
\hline
Context Parallelism (CP) &
\begin{itemize}[leftmargin=*]
  \item Reduces {activation} memory by partitioning sequence length dimension
  \item Can complement TP/DP
  \item Helps scale attention-heavy models {to support ultra-long sequences} especially in multinode settings
\end{itemize} &
\begin{itemize}[leftmargin=*]
  \item Load balancing is a critical challenge requiring sophisticated algorithms
  \item Requires sequence reassembly for some operations
  \item Requires replica of weights similar to DP
\end{itemize} \\
\hline
Expert Parallel (EP) &
\begin{itemize}[leftmargin=*]
  \item Efficient scaling for Mixture-of-Experts
  \item High parameter count without proportional compute
  \item Flexible load balancing with routing
\end{itemize} &
\begin{itemize}[leftmargin=*]
  \item Complex routing and load balancing
  \item Communication overhead from token routing
  \item Training instability if experts are imbalanced
\end{itemize} \\
\hline
Pipeline Parallel (PP) &
\begin{itemize}[leftmargin=*]
  \item Splits model layers across devices
  \item Reduces memory footprint per device
  \item Good for scaling long/deep networks
\end{itemize} &
\begin{itemize}[leftmargin=*]
  \item Pipeline bubbles reduce efficiency
  \item Imbalanced stage partitioning causes idle time
  \item More complex scheduling
\end{itemize} \\
\hline
\end{tabular}
\end{table*}



\paragraph{Memory Constraint.}
The parallelization group allocation search space generally begins with considering the memory constraints of the accelerators on the cluster. So, the first step filters the potential set of parallelization group combinations that allow to fit the model parameters, activation memory, and additional memory requirements on the given cluster. The additional memory requirements considered include gradient and optimizer state for training and transformer k/v-cache for inference. An important consideration here is the impact of different parallelization strategies on total memory footprint. The DP groups can reduce memory by using Zero optimizer which shards optimizer states, gradient and parameters among the accelerators in a DP group. However, there might still problem to fit a large model with DP only. PP groups can further reduce memory by vertically sharding the model layers among the PP groups in a pipeline. TP groups allow to reduce both weight and activation memory by distributing the parameter and compute of a layer. However conventionally it is used withing same node with high bandwidth connection due to the need for collective communication. In scenarios for very long sequence, where activation memory is the critical challenge compared to parameter memory, CP is a useful technique. In particular, the size of the model plays an important role in this choice as small model with long sequence length benefits from CP more than a large model with long context length.


\paragraph{Flops Utilization.} 
Flops Utilization is a critical metric of efficient resource allocation across distributed clusters. Model Flops Utilization (MFU) captures the alignment between the model’s computational graph and the hardware execution pattern. Both are essential for maximizing training throughput and minimizing cost per token. Arithmetic intensity based roof-line model has been widely used and is a dominant approach til date to analyze a parallelization group assignment under service level requirements, e.g., sequence length and maximum workload meeting memory constraints, e.g., batch size. The roof-line model works by categorizing operations under given workload as memory bound or compute bound where compute bound implies better flops utilization. 

To improve flops  utilization, both the model design and parallelization strategy design should consider for operations to be compute bound. Different parallelization strategies affect Flops Utilization in distinct ways. Data Parallelism (DP) scales batch processing across devices which typically increase Flops utilization by allowing larger batch size.
Pipeline bubbles in Pipeline Parallelism (PP) may reduce flops utilization requiring large batch sizes to saturate the pipeline and sophisticated algorithms with less bubbles. Tensor Parallelism (TP) splits tensor operations such as matrix multiplications across devices which can reduce latency. However, larger TP group can lead to too thin compute load resulting in lower arithmetic intensity and also demand high-bandwidth interconnects to avoid communication bottlenecks. Context Parallelism (CP) which is considered for accommodating ultra long sequences may increases orchestration complexity for synchronizing peer-to-peer communications for improved flops utilization. Careful considerations for these aspects are critically important for a good parallelization strategy allocation.

Furthermore, hyper-parameters such as batch size, sequence length, and micro-batch size directly influence Flops Utilization. Larger batch sizes improve flops utilization increasing arithmetic intensity. Longer sequences increase FLOPs per token which can enhance flops utilization. In pipeline parallelism, larger micro-batches reduce idle time. In grid search-based parallelization planning, Flops Utilization serves as a primary scoring criterion. Candidate configurations are first filtered for feasibility (e.g., memory fit), then ranked by estimated MFU for maximum possible workload. Yet there are some specific consideration regarding system design for training vs. inference.

\paragraph{Consideration in Training.} 
In the training phase of large language models (LLMs), the selection of optimal parallelization strategy groups is guided by success metrics and service-level objectives that balance performance, scalability, and resource efficiency. In addition to general considerations of Flops Utilization, key metrics specific to training include 
training throughput (tokens/sec) and time-to-target-loss. These success metrics collectively reflect the system's ability to meet convergence goals within hardware and budget constraints. These metrics are evaluated during parallelization plan search. Allocation of groups of accelerators in the cluster to parallelization strategies such as data, tensor, and pipeline parallelism are assessed for their impact on latency, memory footprint, and inter-device communication. DP enhances throughput by distributing training data across multiple accelerators while PP improves throughput by processing micro-batches in a staggered manner. TP efficiently distributes computation across GPUs while TPSP further optimizes LayerNorm with efficient sequence-level computations. A hybrid system leveraging DP for scale, TP for computational efficiency, and PP for pipelining can significantly enhance model training throughput while reducing synchronization delays. Grid search frameworks incorporate these objectives into scoring functions to ensure that that selected configurations satisfy feasibility constraints as well as optimize for training efficiency and reliability under real-world deployment conditions~\cite{narayanan2021efficient,lin2024nnscaler}.


\paragraph{Considerations in Inference.}
Considerations of success metrics and service-level objectives (SLOs) for the allocation of parallelization strategies to accelerators within a compute cluster during large language model (LLM) inference deployment generally include latency, throughput, and cost-efficiency. Unlike training, where the primary objective is to maximize throughput (tokens/second), inference emphasizes low latency to ensure real-time responsiveness to individual queries. 
Inference systems generally prioritize low latency, potentially trading off some throughput to meet interactive performance requirements. For instance, larger batch size leads to high throughput with high latency while small batch size leads to low throughput and low latency scenario, As a result an optimal balance with a trade-off is often considered. 


During inference, TP is particularly effective for compute-intensive layers, as it enables parallel execution of operations within a single layer across multiple devices. This approach enhances computational throughput by distributing workloads at a fine-grained level. In contrast, PP reduces memory footprint by staging execution across sequential model partitions. Data parallelism (DP) does not directly reduce the latency of individual queries. However, DP can still be beneficial in high-throughput environments serving highly frequent queries, where batching and distributing queries across replicas to improve overall throughput is also important. Competitive trade-offs arise when selecting between TP and PP: TP offers superior compute saturation but demands high-bandwidth interconnects, while PP alleviates memory pressure but may underutilize devices during sequential execution. State-of-the-art hybrid strategies balance memory constraints with latency requirements combining these strategies~\cite{aminabadi2022deepspeed,pope2023efficiently}. Optimal allocation strategies are typically determined through heuristic or grid search methods that evaluate candidate configurations against service-level objectives (SLOs), ensuring that the deployed plan satisfies latency constraints and maximizes accelerator utilization.


By carefully designing a hybrid parallelization system that combines different parallel strategies, AI models can achieve high efficiency, scalability, and performance for both training and inference. The right balance of parallelism ensures that training workloads maximize throughput, while inference remains low-latency and memory-efficient, making modern LLM systems robust and adaptable to growing computational demands.

\subsection{Toward Auto Parallelization}
\label{sec:auto_parallel}

Auto-parallelism is an automated method that selects and applies efficient parallel execution strategies for a neural network model on a specific hardware cluster~\cite{liang2023survey}. It can be formulated as an optimization problem: choosing the best combination of available techniques—including parallelism strategies and memory footprint reduction methods—while ensuring that the memory usage stays within the hardware capacity. The inputs to this problem are an LLM model, a target GPU hardware, and a global batch size. The search space of possible configurations is extremely large; for example, an 80-layer transformer can already yield up to the order of $10^{125}$ possible configurations due to the large number of parallel and memory optimizations(Figure 5 in ~\cite{zhu2025mist}). Auto-parallelism systems typically address this by building and optimizing a cost model tailored to the model–hardware setup, then deploying the strategy using a runtime system that maps tasks to devices. These techniques automatically divide computation and memory across devices, optimizing for training throughput, prefill latency, and low-latency decoding. This section discusses the motivation for auto-parallelization, reviews the successes and limitations of state-of-the-art approaches, and highlights remaining challenges and future directions. Our focus is on the most recent advances, while a more comprehensive overview of earlier works is available elsewhere~\cite{liang2023survey}. We primarily aim to provide a comparative discussion of modern auto parallelization approaches and to identify key opportunities for future research.


Neural scaling law contributed to significant increase in model size and training data which has outgrown the capability of single compute device~\cite{kaplan2020scaling,hoffmann2022training}. In response, parallelization strategies has been appeared which brings diverse advantages and solve some particular aspects of the whole problem~\cite{shoeybi2019megatron,narayanan2019pipedream}. Training and inference efficiency of the frontier models require to combine multiple parallelism method in a hybrid approach where scaling performance relies on properly tuning the combination of the parallelism strategies in a multi dimensional parallelism~\cite{xu2021gspmd,singh2023hybrid,singh20234d,grattafiori2024llama}. This type of hybrid combination also requires to use large clusters of heterogeneous hardware, such as GPUs, TPUs, NPUs~\cite{jiang2024megascale}. The hardware's are connected with different bandwidths and different topology requiring careful assignment of parallelization ranks to hardware. This scenario makes the manual engineering of finding optimal parallelization strategy tedious and time consuming or even impractical for industry grade models. 

Efficiency also requires to consider training and inference separately as well as the prefill and decode phase of the inference. The search space of possible combinations also grow exponentially with parallelization techniques and cluster size. Moreover, the complexity of the problem as becoming more sophisticated as recent proposal also include in-device parallelization in addition to inter device parallelization~\cite{dalal2025one}. Furthermore, this is not just a once a time work, change in hardware require repetition of the work tuning on different hardware types. This makes the manual analysis, estimate or cost model estimate for different configuration quite impractical. 

The need for automating parallelization is crucial for both model developers and infrastructure providers. From the model developer's perspective, large-scale models demand efficient utilization of computation and memory across distributed devices to reduce training time and cost. Meanwhile, hardware vendors and infrastructure providers must ensure that their systems support flexible parallelization strategies to meet varied workloads and client demands. Ignoring hardware-specific factors such as communication topology, memory bandwidth, and interconnect speed can lead to significant inefficiencies in multi-node environments~\cite{narayanan2021efficient}.

The growing scale of deep learning models and the increasing complexity of hardware clusters have motivated a large body of work on automatic parallelization. Early frameworks primarily combined existing forms of parallelism—such as data, pipeline, and tensor parallelism—into unified search spaces, and then developed cost models and optimization strategies to explore them (see Table~\ref{tab:auto-parallel}).

An early system is Piper~\cite{tarnawski2021piper}, which introduced dynamic programming for automatic partitioning, but did not fully consider device heterogeneity. Building on this foundation, several 2022 works refined the search process. Alpa~\cite{zheng2022alpa} organizes parallelism strategies into a hierarchical space aligned with the compute cluster topology, distinguishing between inter-operator partitioning and intra-operator partitioning. By solving these levels independently through integer linear programming and dynamic programming, Alpa achieves practical solutions, but remains sub-optimal and does not explicitly address load imbalance across microbatches. In the same year, Galvatron~\cite{miao2022galvatron} introduced a decision-tree decomposition of the search space, pruning suboptimal combinations and applying dynamic programming to optimize each pipeline stage under memory constraints, but without fully modeling optimizer states or device heterogeneity. Complementarily, AMP~\cite{li2022amp} targeted heterogeneous clusters using symbolic cost models and dynamic programming to select pipeline and tensor strategies efficiently, though its reliance on lookup tables limited accuracy across hardware types.

Subsequent works in 2023 extended automation and usability. Colossal-AI~\cite{li2023colossal} expanded automation to 4D parallelism (data, pipeline, tensor, and sequence), incorporating advanced tensor sharding schemes such as 2D, 2.5D, and 3D tensor parallelism~\cite{xu2023efficient,wang20212,bian2021maximizing}. These reduce communication costs but impose topological constraints. Colossal-AI further improves over Alpa with greedy sharding conversion and integrated activation checkpointing. Merak~\cite{lai2023merak} also embraces 3D parallelism but emphasizes usability with a PyTorch framework that automatically partitions and allocates models, requiring fewer code modifications than Alpa or Colossal-AI. In parallel, SuperScaler~\cite{lin2023superscaler} introduced a unified scheduling framework with three stages: model transformation, space-time scheduling, and data-dependency preserving execution, enabling flexible but not fully automatic parallelism planning.

In 2024, the focus shifted toward heterogeneity and overlap. nnScaler~\cite{lin2024nnscaler} extended scheduling approaches but demonstrated results only on transformer architectures. Metis~\cite{um2024metis} explicitly addressed heterogeneous GPU capabilities through profiling, cost estimation, and pruned search algorithms to balance loads across diverse devices. Aceso~\cite{liu2024aceso} tackled pipeline inefficiency by co-optimizing communication and computation at the microbatch level, reducing pipeline bubbles with overlap-aware scheduling.

Most recently, Mist~\cite{zhu2025mist} pushed the frontier by co-optimizing memory footprint reduction with parallelism strategy selection. Its imbalance-aware hierarchical tuning method decouples inter-stage pipeline partitioning from intra-stage microbatch balancing, linked via Pareto frontier sampling. This design mitigates search space explosion and pipeline imbalance while achieving fine-grained overlap between computation and communication.

Overall, these systems demonstrate a clear evolution: from early dynamic-programming partitioners, to hierarchical and greedy search frameworks, to heterogeneity-aware and overlap-optimized approaches. While significant progress has been made, challenges remain in controlling the combinatorial explosion of strategies, supporting diverse hardware environments, and ensuring usability in practice.

\begin{table*}[htbp]
\centering
\caption{Comparison of recent automatic parallelism frameworks (ordered by year).}
\label{tab:auto-parallel}
\begin{tabular}{lclll}
\toprule
\textbf{System} & \textbf{Year} & \textbf{Parallelism Types} & \textbf{Search / Optimization} \\
\midrule
Piper~\cite{tarnawski2021piper} & 2021 & PP, TP & Dynamic programming \\
Alpa~\cite{zheng2022alpa} & 2022 & DP, PP, TP & Integer linear programming, dynamic programming \\
Galvatron~\cite{miao2022galvatron} & 2022 & DP, PP, TP & Decision tree + Dynamic programming \\
AMP~\cite{li2022amp} & 2022 & PP, TP & Dynamic programming + symbolic cost model \\
Colossal-AI~\cite{li2023colossal} & 2023 & DP, PP, TP, SP & Greedy search \\
Merak~\cite{lai2023merak} & 2023 & DP, PP, TP & Graph-sharding heuristic \\
Metis~\cite{um2024metis} & 2024 & DP, PP, TP & Pruned search algorithm \\
Mist~\cite{zhu2025mist} & 2025 & DP, PP, TP & Mixed integer linear programming \\
\bottomrule
\end{tabular}
\end{table*}



\section{Parallel Strategies Theoretical Analysis}
\label{sec:theory_parallel_strategies}

In this section, we present a unified theoretical analysis of FLOPs, memory consumption, and communication overhead for GQA, MLP, and Mamba across different parallelization scenarios. Our work is inspired by prior analyses in \cite{korthikanti2023reducing, fujii2024accelerating}, which quantified per-GPU memory usage (weights and activations) for GQA and MLP layers. Building on these foundations, we extend the analysis by providing a detailed breakdown of FLOPs, memory, and communication for these blocks (Sections~\ref{sec:gqa_analysis} and \ref{sec:MLP_analysis}). Moreover, we incorporate the Mamba block \cite{dao2024transformers}, which has not been covered in previous studies (Section~\ref{sec:Mamba_analysis}). The parallel strategies in this section are TP, DP, and CP, while we experiment with PP in Section \ref{sec:cae_studies}.

For clarity, we ignore layer normalization and focus on the dominant multiply--add operations (vector FLOPs are omitted). Our analysis considers the forward pass over a full sequence, which corresponds to the \textit{forward} step during training and the \textit{prefill} stage of inference; we use these terms interchangeably. Extending the results to training is feasible: the backward pass roughly doubles the FLOPs due to gradient computations with respect to both inputs and weights. Memory analysis for the backward pass is left for future work. We also neglect the optimizer states and focus on key blocks' weights as well as activations.

We conclude with a summary of the main insights in Section~\ref{sec:summary_theory_analysis}. Parameter definitions and notation are provided in Table~\ref{tab:theor_params}.



\begin{table}[htbp]
\centering
\begin{tabular}{ll}
\hline
\textbf{Symbol} & \textbf{Meaning} \\
\hline
$b$ & Batch size \\
$s$ & Sequence length \\
$l$ & Chunk size \\
$n$ & State dimension (latent recurrent size) maintained by each SSM head \\
$d$ & Transformer hidden dimension \\
$d_{\mathrm{h}}$ & Transformer head dimension \\
$I$ & Intermediate dimension of the FFN (expansion size) \\
$a$ & Number of attention heads \\
$k$ & Number of KV heads \\
$v$ & Vocabulary size (unique tokens) \\
$\text{expand}_{\mathbf{mamba}}$ & Mamba expansion factor for transformer hidden dimension \\
$d_{\mathbf{inner}}$ & Total SSM channel width after expansion \\
$ngroups_{\mathbf{ssm}}$ & Number of groups used to partition SSM parameters \\
$h$ & Number of mamba heads \\
$p$ & Mamba head dimension \\
$a_{\text{byte}}$ & Bytes per activation element \\
$w_{\text{byte}}$ & Bytes per weight element \\
\hline
\end{tabular}
\caption{Parameter definitions used in the transformer and Mamba theoretical derivations.}
\label{tab:theor_params}
\end{table}

Let \text{TP} be the tensor-parallel degree, \text{DP} be the data-parallel degree, and \text{CP} be the context-parallel degree. Suppose that the total number of devices (world size) fixed such that
\[
\text{TP} \times \text{CP} \times \text{DP} = \text{worldsize}.
\]

\text{We decompose the transformer hidden dimension as } 
$d = a \cdot d_h$,
\text{ where } a \text{ is the number of attention heads.}

\text{The SSM hidden width is then given by }
$d_{\textbf{inner}} = \texttt{expand}_{\text{mamba}} \cdot d$.


\subsection{GQA Theoritical Analysis}
\label{sec:gqa_analysis}

\paragraph{Attention GQA FLOPs Breakdown}

\begin{itemize}[noitemsep,nolistsep]
    \item \textbf{Self-Attention (Q, K, V Linear Projection)}: \quad 2$bsad_h \cdot ad_h + 4bskd_h \cdot ad_h$ or 2$bsd \cdot d + 4bskd_hd$
    \item \textbf{Attention ($QK^\top$)}: \quad 2$b s^2 a d_h$ or 2$b s^2 d$
    \item \textbf{Softmax softmax($QK^\top$)}: \quad 5$b s^2 a$ or 5$b s^2 a$ [max, subtraction, exponent, sums, division]
    \item \textbf{Attention Output ($SV$)}: \quad 2$b s^2 a d_h$ or 2$b s^2 d$
    \item \textbf{Final Linear Projection (($SV) \cdot W_O$)}: \quad 2$bs a d_h \cdot a d_h$ or 2$bs d \cdot d$
\end{itemize}




\noindent \textbf{Total FLOPs for GQA:}

\[
 4bsd^2 + 4bskd_hd + 4bs^2d + 5bs^2a 
\]


\underline{FLOPs with TP Per Device:}

\[
 4bs\frac{d^2}{TP} + 4bs\frac{kd_h}{TP} + 4bs^2\frac{d}{TP} + 5bs^2\frac{a}{TP} 
\]

\underline{FLOPs with CP Per Device:}

\[
 4bd^2\frac{s}{CP} + 4bkd_h\frac{s}{CP} + 4bds\frac{s}{CP} + 5bsa\frac{s}{CP} 
\]


\underline{FLOPs with DP Per Device:}

\[
 4sd^2\frac{b}{DP} + 4skd_h\frac{b}{DP} + 4ds^2\frac{b}{DP} + 5s^2a\frac{b}{DP} 
\]

\paragraph{Conclusion:}
\[
    \text{FLOPs}(\text{TP},\text{CP}, \text{DP})
    \text{ is invariant w.r.t. the TP--CP--DP split.}
\]



\paragraph{GQA Activation Memory Breakdown}

\begin{itemize}[noitemsep,nolistsep]
    \item \textbf{Linear Projections ($W_Q, W_K, W_V$)}: \quad $bsd + 2bsd \cdot \frac{k}{a}$
    \item \textbf{Attention ($QK^\top$)}: \quad $0$ \quad \textit{(due to FlashAttention)}
    \item \textbf{Softmax}: \quad $0$ \quad \textit{(due to FlashAttention)}
    \item \textbf{Attention Output ($SV$)}: \quad $bsd$
    \item \textbf{Final Linear Projection ($SV \cdot W_O$)}: \quad $bsd$
\end{itemize}

\noindent \textbf{Total Activation Memory for GQA:}
\[
(3bsd + 2bsd \cdot \frac{k}{a}) \cdot a_{\text{byte}} 
\]

\noindent \textbf{Total Activation TP for GQA Per Device:}

\[
(bsd + 2bs\frac{d}{TP} + 2bs\frac{d}{TP} \cdot \frac{k}{a}) \cdot a_{\text{byte}} 
\]

\noindent \textbf{Total Activation CP for GQA Per Device:}
\[
(bd\frac{s}{CP} + 2bd\frac{s}{CP} + 2bd\frac{s}{CP} \cdot \frac{k}{a}) \cdot a_{\text{byte}} 
\]

\noindent \textbf{Total Activation DP for GQA Per Device:}
\[
(sd\frac{b}{DP} + 2sd\frac{b}{DP} + 2sd\frac{b}{DP} \cdot \frac{k}{a}) \cdot a_{\text{byte}} 
\]

\paragraph{Conclusion:}

$\text{memory}_{\text{act}}^{\text{(gqa)}}(\text{TP}) >\text{memory}_{\text{act}}^{\text{(gqa)}}(\text{CP})$
AND
$\text{memory}_{\text{act}}^{\text{(gqa)}}(\text{TP}) >\text{memory}_{\text{act}}^{\text{(gqa)}}(\text{DP})$
AND
$\text{memory}_{\text{act}}^{\text{(gqa)}}(\text{CP}) \approx\text{memory}_{\text{act}}^{\text{(gqa)}}(\text{DP})$



    
The activation memory per device is lower with a larger CP or DP value. At the same CP and DP values, the activation memory per device is the same for CP and DP.

\paragraph{GQA Weight Memory Breakdown}
\mbox{}\\ 

\textbf{Total Weight Memory for GQA:} (\(W_Q, W_K, W_V, W_O\)): \quad \(2d^2 \left(1 + \frac{k}{a}\right) \cdot w_{\text{byte}}\) 

\noindent \textbf{Total Weight TP for GQA per device:}

\[
2\frac{d^2}{TP} \left(1 + \frac{k}{a}\right) \cdot w_{\text{byte}}
\]

\noindent \textbf{Total Weight CP for GQA per device:}
\[
2d^2 \left(1 + \frac{k}{a}\right) \cdot w_{\text{byte}}
\]

\noindent \textbf{Total Weight DP for GQA per device:}
\[
2d^2 \left(1 + \frac{k}{a}\right) \cdot w_{\text{byte}}
\]


\paragraph{Conclusion:}

$\text{memory}_{\text{wt}}^{\text{(gqa)}}(\text{TP}) <\text{memory}_{\text{wt}}^{\text{(gqa)}}(\text{CP})$
AND
$\text{memory}_{\text{wt}}^{\text{(gqa)}}(\text{TP}) <\text{memory}_{\text{wt}}^{\text{(gqa)}}(\text{DP})$
AND
$\text{memory}_{\text{wt}}^{\text{(gqa)}}(\text{CP}) \approx\text{memory}_{\text{wt}}^{\text{(gqa)}}(\text{DP})$



The weight memory per device is lower with a larger TP value.

\subsection{MLP Theoretical Analysis}
\label{sec:MLP_analysis}

We assume that the SwiGLU is applied in this MLP as normally done in Llama architectures \cite{shazeer2020glu, fujii2024accelerating}. 

\paragraph{MLP FLOPs Breakdown}

\begin{itemize}[noitemsep,nolistsep]
    \item \textbf{MLP Up Projection}: \quad 2$bsd \cdot I$
    \item \textbf{MLP Gated Dot Product and SiLU operations}: \quad $5bsI$ (4 for the sigmoid and 1 for the dot product)
    \item \textbf{MLP Gated Projection}: \quad 2$bsd \cdot I$
    \item \textbf{MLP Down Projection}: \quad 2$bsd \cdot I$
\end{itemize}


\noindent \textbf{Total FLOPs for MLP:}
\[
6bsdI + 5bsI = bs(6dI + 5I)
\]

\underline{FLOPs with TP Per Device:}
\[
6bsdI + 5bsI = \frac{bs}{TP}(6dI + 5I)
\]

\underline{FLOPs with CP Per Device:}
\[
6bsdI + 5bsI = \frac{bs}{CP}(6dI + 5I)
\]

\underline{FLOPs with DP Per Device:}
\[
6bsdI + 5bsI = \frac{bs}{DP}(6dI + 5I)
\]


\paragraph{Conclusion:}
\[
    \text{FLOPs}(\text{TP},\text{CP}, \text{DP})
    \text{ is invariant w.r.t. the TP--CP--DP split.}
\]

\paragraph{MLP Activation Memory Breakdown}

\begin{itemize}[noitemsep,nolistsep]
    \item \textbf{Input}: \quad $bsd$
    \item \textbf{Output of Up Projection}: \quad $bsI$
    \item \textbf{Output of Gate Projection}: \quad $bsI$
    \item \textbf{GELU Activation}: \quad $bsI$
    \item \textbf{Input to Down Projection}: \quad $bsI$
\end{itemize}

\noindent \textbf{Total Activation Memory for FFN:}
\[
(2bsd + 4bsI) \cdot a_{\text{byte}}
\]

\noindent \textbf{Total Activation TP for MLP Per Device:}
\[
(2bsd + \frac{bs}{TP}4I) \cdot a_{\text{byte}}
\]


\noindent \textbf{Total Activation CP for MLP Per Device:}
\[
\frac{bs}{CP}(2d + 4I) \cdot a_{\text{byte}}
\]

\noindent \textbf{Total Activation DP for MLP Per Device:}
\[
\frac{bs}{DP}(2d + 4I) \cdot a_{\text{byte}}
\]


\paragraph{Conclusion:}

$\text{memory}_{\text{act}}^{\text{(mlp)}}(\text{TP}) >\text{memory}_{\text{act}}^{\text{(mlp)}}(\text{CP})$
AND
$\text{memory}_{\text{act}}^{\text{(mlp)}}(\text{TP}) >\text{memory}_{\text{act}}^{\text{(mlp)}}(\text{DP})$
AND
$\text{memory}_{\text{act}}^{\text{(mlp)}}(\text{CP}) \approx\text{memory}_{\text{act}}^{\text{(mlp)}}(\text{DP})$

The activation memory per device is lower with a larger CP or DP value. At the same CP and DP values, the activation memory per device is the same for CP and DP

    

\paragraph{MLP Weight Memory Breakdown}
\mbox{}\\ 

\textbf{Feed-Forward Network (FFN) Parameters}: \quad \(3dI\)

\textbf{Feed-Forward Network (FFN) Parameters with TP (TPSP)}: \quad \(\frac{3dI}{TP}\)

\textbf{Feed-Forward Network (FFN) Parameters with CP}: \quad \({3dI}\)

\textbf{Feed-Forward Network (FFN) Parameters with DP}: \quad \({3dI}\)

\paragraph{Conclusion:}

$\text{memory}_{\text{wt}}^{\text{(mlp)}}(\text{TP}) <\text{memory}_{\text{wt}}^{\text{(mlp)}}(\text{CP})$
AND
$\text{memory}_{\text{wt}}^{\text{(mlp)}}(\text{TP}) <\text{memory}_{\text{wt}}^{\text{(mlp)}}(\text{DP})$
AND
$\text{memory}_{\text{wt}}^{\text{(mlp)}}(\text{CP}) \approx\text{memory}_{\text{wt}}^{\text{(mlp)}}(\text{DP})$


The weight memory per device is lower with a larger TP value.

\paragraph{GQA and MLP Communication Overhead Analysis}


The total communication volume for both GQA and MLP for TP is:
\[
= 2 * s * a * d_h
\]

When CP with passKV is enabled, the total volume will be:
\[
= 2 * s * k * d_h
\]

Since the number of KV heads ($k$) is typically less than the number of attention heads $a$, the communication volume for CP is less than the communication volume for TP. 
For the DP and prefill scenario, there is no communication needed.

\subsection{Mamba Theoretical Analysis}
\label{sec:Mamba_analysis}

This section studies the Mamba-2 architecture \cite{dao2024transformers} and presents analysis for the corresponding FLOPs, memory, and communication. In this analysis, we do not consider the convolution operation and focus on major GEMM operations.


\paragraph{Mamba Theoretical Fundamentals}

Before deriving the characteristics (FLOPs, memory, and communication) of Mamba, we \textit{review} the theoretical fundamentals of Mamba-2 (originally developed in \cite{dao2024transformers}) in a simple way to align with our following derivations: 

\[
\textbf{SSM recurrence (scalar, time-varying):}
\qquad
h_{t+1} = a_{t+1}\, h_t + b_{t+1}\, x_{t+1}.
\]

\[
\text{Unroll one step:} \quad
\begin{aligned}
h_{t+1} 
&= a_{t+1} h_t + b_{t+1} x_{t+1} \\
&= a_{t+1} \big(a_t h_{t-1} + b_t x_t\big) + b_{t+1} x_{t+1} \\
&= a_{t+1} a_t h_{t-1} + a_{t+1} b_t x_t + b_{t+1} x_{t+1}.
\end{aligned}
\]


\[
\text{Unroll further:} \quad
\begin{aligned}
h_{t+1} 
&= a_{t+1} a_t a_{t-1} h_{t-2} 
   + a_{t+1} a_t b_{t-1} x_{t-1} 
   + a_{t+1} b_t x_{t+1} 
   + b_{t+1} x_{t+1}.
\end{aligned}
\]


\[
\text{General closed form (scalar, time-varying) for } t+1 \text{ steps:} \quad
h_{t+1} = \Big(\prod_{m=0}^{t} a_{m+1}\Big) h_0
+ \sum_{i=0}^{t} \Big(\prod_{m=i+1}^{t} a_{m+1}\Big) b_{i+1} x_{i+1}.
\]


\[
\textbf{Chunking notation:}
\quad
\begin{cases}
\text{chunk size } Q, \\
\text{chunk index } c\in\{0,\dots,K-1\},\\
\text{in-chunk position } j\in\{0,\dots,Q-1\},\\
\text{global time } t = cQ + j.
\end{cases}
\]

\[
\text{Write }h_{c,j}\text{ for the state at chunk }c\text{ position }j.
\]


\[
\textbf{Intra-chunk expansion (within chunk }c\text{):} \quad
h_{c,j} = \Big(\prod_{m=0}^{j} a_{c,m}\Big) h_{c,0} 
+ \sum_{i=0}^{j} \Big(\prod_{m=i+1}^{j} a_{c,m+1}\Big) b_{c,i+1}\, x_{c,i+1}.
\]

\begin{align*}
\text{In particular, the final state at the end of chunk }c\ (j=Q):\quad
h_{c,Q} &= \Big(\prod_{m=0}^{Q} a_{c,m}\Big) h_{c,0} \\
&\quad + \sum_{i=0}^{Q} \Big(\prod_{m=i+1}^{Q} a_{c,m+1}\Big) b_{c,i+1}\, x_{c,i+1}.
\end{align*}

\[
\textbf{Inter-chunk recurrence (carry between chunks):}
\qquad
h_{c+1,0} = A_{c}^{(Q)}\, h_{c,0} + U_{c+1},
\]
where we define the chunk-level decay and intra-chunk contribution
\[
A_{c}^{(Q)} \coloneqq \prod_{m=0}^{Q} a_{c,m},\qquad
U_{c+1} \coloneqq \sum_{i=0}^{Q} \Big(\prod_{m=i+1}^{Q} a_{c,m+1}\Big) b_{c,i+1}\, x_{c,i+1}.
\]


\[
\text{By recursion across chunks:} \quad
h_{c,0} = \Bigg(\prod_{i=0}^{c} A_i^{(Q)}\Bigg) h_{0,0} 
+ \sum_{i=0}^{c} \Bigg( \Bigg[\prod_{m=i+1}^{c} A_m^{(Q)} \Bigg] U_{i+1} \Bigg).
\]


\[
\textbf{Combine intra + inter: final expression for } h_{c,j}: \quad
\begin{aligned}
h_{c,j} 
&= \underbrace{\Big(\prod_{m=0}^{j} a_{c,m}\Big) h_{c,0}}_{\text{evolution of chunk-start state (inter)}} \\
&\quad + \underbrace{\sum_{i=0}^{j} \Big(\prod_{m=i+1}^{j} a_{c,m+1}\Big) b_{c,i+1}\, x_{c,i+1}}_{\text{intra-chunk inputs}}.
\end{aligned}
\]


\[
\text{And } h_{c,0} \text{ is given by the inter-chunk prefix above, so } h_{c,j} \text{ equals}
\]

\[
\boxed{
h_{c,j} 
= \Bigg(\prod_{m=0}^{j} a_{c,m}\Bigg)
\Bigg[
\Big(\prod_{i=0}^{c} A_{i}^{(Q)}\Big) h_{0,0}
+ \sum_{i=0}^{c} \Big(\prod_{m=i+1}^{c} A_m^{(Q)}\Big) U_{i+1}
\Bigg]
\;+\;
\sum_{i=0}^{j} \Big(\prod_{m=i+1}^{j} a_{c,m+1}\Big) b_{c,i+1}\, x_{c,i+1}
}.
\]

This equation clearly illustrates that in the Mamba2 architecture, the intra- and inter-chunk computations are structured such that each chunk depends only on its starting state.

\paragraph{Mamba FLOPs Breakdown}
Consider the following variable, 
\[
d_{inproj} = 2 * d_{inner} + 2 * ngroups * n + h
\]

The first term \(2 d_{\text{inner}}\) arises because the input \(X\) is split into two
branches: one feeding the state-space duality (SSD) path and the other producing the 
residual update \(Z\), each of size \(d_{\text{inner}}\). The second term
\(2(\text{ngroups} \cdot n)\) corresponds to the \(B\) and \(C\) parameters of the
selective {SSD}, where each {SSD} group contributes an \(n\)-dimensional vector for both
\(B\) and \(C\). The final term \(h\) comes from the projection generating the
\(A\)-parameter of the {SSD}, which has dimensionality \(h\).

\paragraph{In-Proj and Out-Proj FLOPs}

\begin{itemize}[noitemsep,nolistsep]
    \item \textbf{In Projection:} \quad $2 b s d \cdot d_{inproj}$
    \item \textbf{Out Projection:} \quad $2 b s \, d_{\text{inner}} \, d$
\end{itemize}



\paragraph{SSD FLOPs}

To derive SSD FLOPs, we consider the following variables:

\[
l = \text{chunk size}, \quad c = s/l.
\]

We analyze the operations outlined in the Mamba-2 algorithm in \cite{dao2024transformers}.




\paragraph{Step 1 \& Step 2: Computing $Y_{\text{diag}}$}


In Mamba-2 paper, they define the diagonal contribution as:
\[
Y_{\text{diag}}
    = \texttt{einsum}\!\left(
        \text{"bclhn, bcshn, bhcls, bcshp} \rightarrow \text{bclhp"},
        \; C,\, B,\, L,\, X
      \right).
\]

Because we defined $s$ to be the sequence length and the $s$ variable in the equation is different, we replace $s$ in the equation by $v$ in the diagonal contribution for clarity:
\[
Y_{\text{diag}}
    = \texttt{einsum}\!\left(
        \text{"bclhn, bcvhn, bhclv, bcvhp} \rightarrow \text{bclhp"},
        \; C,\, B,\, L,\, X
      \right).
\]

In \texttt{einsum}, any index that appears in \emph{both} input tensors but \emph{not in the output} is summed over (contracted). Based on this contraction rule, the computation of $Y_{\text{diag}}$ can be understood step by step as follows:



\begin{enumerate}[label=\Roman*.]
    \item 
    \[
        C_{bclhn} \cdot B_{bcvhn} \;\longrightarrow\; \text{intermediate}_{bcvhl}.
    \] 
    The index that appears in {both} input tensors but {not in the output} is $n$. All other indices ($b,c,s,h,l$) are kept in the output.

    For the diagonal kernels in Mamba, the effective sequence length inside each
    kernel is equal to the chunk size~$v=l$. This is because the diagonal SSM
    component operates only within each chunk, performing purely
    intra\mbox{-}chunk interactions. As a result, each chunk behaves as an
    independent sequence of length~$l$, and the diagonal computation forms an
    $l \times l$ interaction pattern. This is precisely why the FLOPs expressions
    contain an $l^{2}$ term.
    
    Over the full sequence of length~$s$, we process $s/l$ such chunks. Therefore,
    the FLOPs for the diagonal GEMMs are:
    \[
    \text{FLOPs}_{1}
        = 2 b \left(\frac{s}{l}\right) h\, l^{2} n
        \qquad\text{(first GEMM, $Y_{\text{diag}}$)}
    \]

    The key idea is that the diagonal structure forces an $l \times l$
    computation inside each chunk, while the factor $s/l$ accounts for the number
    of chunks required to cover the full sequence.



    \item 
    \[
        \text{intermediate}_{bcvhl} \cdot L_{bhclv}
        \;\longrightarrow\;
        \text{intermediate}_{bcvhl}.
    \]
    This is an elementwise operation with no contraction.  
    The shapes are \(bhclv \times bhclv \rightarrow bhcls\).  
    Therefore, the vector FLOPs are \(bhclv\). Since we focus on cube FLOPs, we ignore this one in the total sum.

    \item 
    \[
        \text{intermediate}_{bcvhl} \cdot X_{bcvhp} \;\longrightarrow\; Y_{\text{diag},bclhp}.
    \] 
    The contraction is over the $v$ dimension, and the final output has the desired shape $(b,c,l,h,p)$. The FLOPs for this is:
    \[
    \text{FLOPs}_{2}
        = 2 b \left(\frac{s}{l}\right) h\, l^{2} p
        \qquad\text{(second GEMM, $Y_{\text{diag}}$)}. 
    \]
\end{enumerate}







\paragraph{Step 3: Computing the intra-chunk states}

The updated states are computed using the einsum
\[
\text{states}
    = \texttt{einsum}\!\left(
        \text{"bclhn, bhcl, bclhp} \rightarrow \text{bchpn"},
        \; B,\, \text{decay\_states},\, X
      \right).
\]

Here, the contraction proceeds as follows:

\begin{itemize}

    \item $\text{decay\_states} \in \mathbb{R}^{b \times h \times c \times l}$ is applied as an element-wise multiplication with a broadcast across $n$, introducing the state decay. We ignore vector operations in total FLOPs.

    \item $B \in \mathbb{R}^{b \times c \times l \times h \times n}$ and $X \in \mathbb{R}^{b \times c \times l \times h \times p}$ are contracted over the chunk dimension $l$, producing the resulting output tensor has shape $b \times c \times h \times p \times n$, corresponding to all heads, channels, and features aggregated across the chunks.

    The corresponding cost is
    \[
    \text{FLOPs}_{3}
        = 2 b \left(\frac{s}{l}\right) h\, l\, p\, n .
    \]

\end{itemize}


\paragraph{Step 4: Computing the inter-chunk recurrent states}

\[
\text{new\_states}
    = \texttt{einsum}\!\left(
        \text{"bhzc, bchpn} \rightarrow \text{bzhpn"},
        \; \text{decay\_chunk},\, \text{states}
      \right).
\]
Note: 
\[
z = (\frac{s}{l} +1)
\]

This corresponds to
\[
\text{FLOPs}_{4,\text{naive}}
    = 2 b\, h \left(\frac{s}{l}\right)\!\left(\frac{s}{l}+1\right) p\, n .
\]

\paragraph{Note about inter-Chunk recurrent states calculation:} Based on the previous theoretical development, the inter-chunk recurrence is defined as:
\[
H_c = A_{c-1}^{(Q)} H_{c-1} + U_c,
\]
where $H_c$ is the state for chunk $c$, $A_{c-1}^{(Q)}$ is the decay factor between chunks, and $U_c$ is the intra-chunk contribution.

\[
U_{c} \coloneqq \sum_{i=0}^{Q-1} \Big(\prod_{m=i+1}^{Q-1} a_{c,m}\Big) b_{c,i+1}\, x_{c,i+1}.
\]

\textbf{Naive computation:}  
To compute all inter-chunk contributions explicitly, we would write:
\[
H_1 = U_1, \quad
H_2 = A_1 U_1 + U_2, \quad
H_3 = A_2(A_1 U_1 + U_2) + U_3, \dots
\]
For $Z = s/l$ chunks, computing $H_Z$ requires summing over all previous chunks, resulting in $\sim Z^2/2$ operations, i.e., quadratic in sequence length.

\textbf{Concrete example:}  
Suppose $s = 16$, $l = 4$, so $Z = s/l = 4$ chunks:

\[
\begin{aligned}
H_1 &= U_1 \\
H_2 &= A_1 H_1 + U_2 = A_1 U_1 + U_2 \\
H_3 &= A_2 H_2 + U_3 = A_2(A_1 U_1 + U_2) + U_3 \\
H_4 &= A_3 H_3 + U_4 = A_3(A_2(A_1 U_1 + U_2) + U_3) + U_4
\end{aligned}
\]

- Naively, we compute each $H_c$ by multiplying all previous $U$'s by the decay factors → quadratic in $Z$.

\textbf{Parallel scan trick in \cite{dao2024transformers}:}  

- The recurrence is \textbf{associative}:
\[
H_c = \sum_{i=1}^{c} \Big( \prod_{j=i}^{c-1} A_j \Big) U_i
\]
- This is exactly a \textbf{prefix-sum (scan)} over the chunks with multiplication by decay factors.\\
- Using a segmented parallel scan, we can compute all $H_c$ in \textbf{$\mathcal{O}(Z)$ total operations}:


\[
\begin{aligned}
\text{steps:} \\
1.\ & \text{Compute cumulative products of } A_j \text{ per chunk} \\
2.\ & \text{Multiply with } U_i \text{ using vectorized einsum} \\
3.\ & \text{Sum contributions across chunks}
\end{aligned}
\]

\textbf{Step-by-step for the example with $Z=4$:}  

\[
\text{Step 1 (cumulative products): } P = [1, A_1, A_1 A_2, A_1 A_2 A_3]
\]  
\[
\text{Step 2 (multiply with $U_i$): } H_i = P_i \cdot U_i
\]  
\[
\text{Step 3 (sum contributions for each chunk): } H_c = \sum_{i=1}^{c} H_i
\]

- Each chunk only \textbf{needs one multiplication per $U_i$ and one addition}, not a full nested loop over previous chunks. \\ 
- Total complexity reduces from $\mathcal{O}(Z^2)$ to $\mathcal{O}(Z)$, i.e., linear in the number of chunks.

\textbf{Resulting FLOPs:}  
\[
\text{FLOPs}_{4,\text{scan}} = 2 \cdot b \cdot h \cdot (s/l) \cdot p \cdot n \sim O(s)
\]

- This is why Mamba achieves \emph{linear complexity} in sequence length for inter-chunk recurrence.


\paragraph{Step 5: Computing $Y_{\text{off}}$ (state-to-output contribution)}

\[
Y_{\text{off}}
    = \texttt{einsum}\!\left(
        \text{"bclhn, bchpn, bhcl} \rightarrow \text{bclhp"},
        \; C,\, \text{states},\, \text{state\_decay\_out}
      \right).
\]

This operation has two steps: (i) a contraction operation between $C$ and $states$ over $n$; (ii) an element-wise operation with broadcast over $p$ between the intermediate tensor and $state\_decay\_out$.

Since we only focus on cube FLOPs, the cost of this final projection is the following:
\[
\text{FLOPs}_{5}
    = 2 b \left(\frac{s}{l}\right) h\, p\, n\, l .
\]


\textbf{Total naive FLOPs for SSD section}

\[
\text{FLOPs}_{\text{SSD,naive}}
= 2b \Big[ ((s/l)) h l^2 n + (s/l) h l^2 p + (s/l) h l p n + h (s/l)((s/l)+1) p n + (s/l) h p n l \Big]
\]

\textbf{Total Parallel Scan FLOPs for SSD section}

\[
\text{FLOPs}_{\text{SSD,parallel}}
= 2b \Big[ ((s/l)) h l^2 n + (s/l) h l^2 p + (s/l) h l p n + h (s/l)p n + (s/l) h p n l \Big]
\]

\paragraph{Total Mamba FLOPs with Parallel Scan}

We sum the contributions from SSD section, in-projection, and out-projection:

\[
\begin{aligned}
\text{FLOPs}_{\text{total}} &= 
\underbrace{2b \Big[ ((s/l)) h l^2 n + (s/l) h l^2 p + (s/l) h l p n + h (s/l) p n + (s/l) h p n l \Big]}_{\text{SSD (parallel scan)}} \\
&\quad + 
\underbrace{2 b s d \cdot d_{inproj}}_{\text{In Projection}} \\
&\quad + 
\underbrace{2 b s \, d_{\text{inner}} \, d}_{\text{Out Projection}}
\end{aligned}
\]

All terms are in \textbf{multiply-add FLOPs}.  

\textbf{Total Mamba FLOPs with Parallel Scan and TP Applied}


\[
\begin{aligned}
\text{FLOPs}_{\text{total}} &= 
\underbrace{\frac{2b}{TP} \Big[ ((s/l)) h l^2 n + (s/l) h l^2 p + (s/l) h l p n + h (s/l) p n + (s/l) h p n l \Big]}_{\text{SSD (parallel scan)}} \\
&\quad + 
\underbrace{2 b s d \cdot \frac{d_{inproj}}{TP}}_{\text{In Projection}} \\
&\quad + 
\underbrace{2 b s \, d_{\text{inner}} \, \frac{d}{TP}}_{\text{Out Projection}}
\end{aligned}
\]

\textbf{Total Mamba FLOPs with Parallel Scan and CP Applied}


\[
\begin{aligned}
\text{FLOPs}_{\text{total}} &= 
\underbrace{\frac{2b}{CP} \Big[ ((s/l)) h l^2 n + (s/l) h l^2 p + (s/l) h l p n + h (s/l) p n + (s/l) h p n l \Big]}_{\text{SSD (parallel scan)}} \\
&\quad + 
\underbrace{2 b d \frac{s}{CP} \cdot d_{inproj}}_{\text{In Projection}} \\
&\quad + 
\underbrace{2 b d \, d_{\text{inner}} \, \frac{s}{CP}}_{\text{Out Projection}}
\end{aligned}
\]


\textbf{Total Mamba FLOPs with Parallel Scan and DP Applied}

\[
\begin{aligned}
\text{FLOPs}_{\text{total}} &= 
\underbrace{\frac{2b}{DP} \Big[ ((s/l)) h l^2 n + (s/l) h l^2 p + (s/l) h l p n + h (s/l) p n + (s/l) h p n l \Big]}_{\text{SSD (parallel scan)}} \\
&\quad + 
\underbrace{2 s d \frac{b}{DP} \cdot d_{inproj}}_{\text{In Projection}} \\
&\quad + 
\underbrace{2 s d \, d_{\text{inner}} \, \frac{b}{DP}}_{\text{Out Projection}}
\end{aligned}
\]

\paragraph{Conclusion:}
\[
    \text{FLOPs}(\text{TP},\text{CP}, \text{DP})
    \text{ is invariant w.r.t. the TP--CP--DP split.}
\]


\paragraph{Mamba Memory Breakdown}



\paragraph{In-Projection Memory}
The symmetric in-projection dimension is
\[
d_{\text{in\_proj\_sym}} = 2\, d_{\text{ssm,local}} + 2\, n_{\text{groups,local}}\, s_{\text{ssm,state}} + n_{\text{heads,local}}.
\]

Then the memory cost for the in-projection is
\[
\begin{aligned}
\text{memory}_{\text{in\_proj}} 
&= b \cdot s \cdot d_\text{model} \cdot a_\text{byte} 
+ d_\text{model} \cdot d_\text{in\_proj\_sym} \cdot w_\text{byte} 
+ b \cdot s \cdot d_\text{in\_proj\_sym} \cdot a_\text{byte} 
\end{aligned}
\]

\paragraph{Out-Projection Memory}
\[
\begin{aligned}
\text{memory}_{\text{out\_proj}} &= 
b \cdot s \cdot d_\text{inner} \cdot a_\text{byte} + d_\text{model} \cdot d_\text{inner} \cdot w_\text{byte} 
+ b \cdot s \cdot d_\text{model} \cdot a_\text{byte} 
\end{aligned}
\]

\paragraph{SSD Section Memory per GEMM}

\begin{align*}
\text{1. First GEMM (}Y_{\text{diag}}\text{)} 
&: \text{bcshn,bclhn $\rightarrow$ bchls} \\
\text{memory}_1 
&= \Big(
    b \cdot l \cdot (s/l) \cdot n \cdot h 
    + b \cdot h \cdot (s/l) \cdot n 
    + h
\Big) \cdot a_{\text{byte}} \\
&\quad
+ \Big(
    b \cdot (s/l) \cdot h \cdot l^2
\Big) \cdot a_{\text{byte}}
\\[2mm]
\text{2. Second GEMM (}Y_{\text{diag}}\text{)} 
&: \text{bchls,bcshp $\rightarrow$ bclhp} \\
\text{memory}_2 
&= \Big(
    b \cdot (s/l) \cdot h \cdot l^2
    + b \cdot (s/l) \cdot l \cdot h \cdot p
\Big) \cdot a_{\text{byte}} \\
&\quad
+ \Big(
    b \cdot (s/l) \cdot l \cdot h \cdot p
\Big) \cdot a_{\text{byte}}
\\[2mm]
\text{3. Third GEMM (intra-chunk states)} 
&: \text{cbhpl,bclhn $\rightarrow$ bchpn} \\
\text{memory}_3 
&= \Big(
    b \cdot (s/l) \cdot h \cdot p \cdot l
    + b \cdot (s/l) \cdot l \cdot h \cdot n
\Big) \cdot a_{\text{byte}} \\
&\quad
+ \Big(
    b \cdot (s/l) \cdot h \cdot p \cdot n
\Big) \cdot a_{\text{byte}}
\\[2mm]
\text{4. Fourth GEMM (new\_states, inter-chunk)} 
&: \text{bhzc,bchpn $\rightarrow$ bzhpn} \\
\text{memory}_4 
&= \Big(
    b \cdot h \cdot ((s/l)+1) \cdot (s/l)
\Big) \cdot a_{\text{byte}} \\
&\quad
+ \Big(
    b \cdot (s/l) \cdot h \cdot p \cdot n
\Big) \cdot a_{\text{byte}} \\
&\quad
+ \Big(
    b \cdot ((s/l)+1) \cdot h \cdot p \cdot n
\Big) \cdot a_{\text{byte}}
\\[2mm]
\text{5. Fifth GEMM (}Y_{\text{off}}\text{)} 
&: \text{bchpn,bclhn $\rightarrow$ cbhpl} \\
\text{memory}_5
&= \Big(
    b \cdot (s/l) \cdot h \cdot p \cdot n
    + b \cdot (s/l) \cdot l \cdot h \cdot n
\Big) \cdot a_{\text{byte}} \\
&\quad
+ \Big(
    b \cdot (s/l) \cdot h \cdot p \cdot l
\Big) \cdot a_{\text{byte}}
\end{align*}

\paragraph{Total SSD Memory}
\[
\text{memory}_\text{SSD,total} = \sum_{i=1}^{5} \text{memory}_i
\]


\paragraph{Conclusion}

\begin{itemize}
    \item \textbf{In-/Out-Projection Weight Memory (per device).}

    $\text{memory}_{\text{weight}}^{\text{(in/out)}}(\text{TP}) <\text{memory}_{\text{weight}}^{\text{(in/out)}}(\text{CP})$
    AND
    $\text{memory}_{\text{weight}}^{\text{(in/out)}}(\text{TP}) <\text{memory}_{\text{weight}}^{\text{(in/out)}}(\text{DP})$
    AND
    $\text{memory}_{\text{weight}}^{\text{(in/out)}}(\text{CP}) \approx\text{memory}_{\text{weight}}^{\text{(in/out)}}(\text{DP})$
    
    Increasing tensor parallelism further partitions the projection weights,
    thus reducing the parameter memory per device.

    \item \textbf{In-/Out-Projection Activation Memory (per device).}

    
    \[
    \text{memory}_{\text{act}}^{\text{(in)}}(\text{TP},\text{CP}, \text{DP})
    \text{ is invariant w.r.t. the TP--CP--DP split.}
    \]

    \item \textbf{SSD / GEMM Memory (per device).}
    \[
    \text{memory}_{\text{SSD}}(\text{TP},\text{CP}, \text{DP})
    \text{ is approximately invariant w.r.t. the TP--CP--DP split.}
    \]
    The local chunk/state size per device does not roughly change as long as
    the total number of devices is fixed.  All the terms in the activation memory except one term $b \cdot (\frac{s}{l}) \cdot l^3$ has either $b$ or $s$ or $h$ where DP, CP, and TP divide, respectively. 
\end{itemize}

\paragraph{Mamba Communication Breakdown}

The communication cost of tensor-parallel state propagation (TPSP) scales as
\[
\text{Comm}_{\text{TPSP}}
= \mathcal{O}\big(d \cdot d_{\text{state}} \cdot \text{seqlen}\big).
\]

In contrast, the communication cost of context parallelism (CP) scales as
\[
\text{Comm}_{\text{CP}}
= \mathcal{O}\big(d \cdot d_{\text{state}} \cdot \text{CP}\big).
\]

Since typically
\[
\text{CP} \ll \text{seqlen},
\]
we conclude that
\[
\text{Comm}_{\text{CP}} < \text{Comm}_{\text{TPSP}},
\]
that is, \emph{CP with Mamba incurs less communication than TPSP. DP requires no communication in the prefill scenario.} We do not include the communication analysis for TP because Mamba typically uses TP for the Mamba layers and SP for the linear layers (TPSP).

\begin{table}[h]
\centering
\begin{tabular}{lcc}
\toprule
\textbf{Method} 
& \textbf{Communication Volume} 
& \textbf{Collective Operation} \\
\midrule
TPSP
& $\mathcal{O}\!\left(d \cdot d_{\text{state}} \cdot \text{seqlen}\right)$
& AllReduce \\
CP
& $\mathcal{O}\!\left(d \cdot d_{\text{state}} \cdot \text{CP}\right)$
& All-Gather \\
DP & 0 & - \\
\bottomrule
\end{tabular}
\caption{Comparison of TPSP, CP, DP communication in Mamba-2 Prefill.}
\end{table}

\subsection{Summary of the Theoretical Analysis}
\label{sec:summary_theory_analysis}

Bringing the analysis together, we emphasize the following key insights:
\begin{itemize}
    \item \textbf{FLOPs remain unchanged across parallelization strategies.}
    The total computational workload is invariant to whether we select TP, CP, or DP.

    \item \textbf{Communication cost is an important factor in selecting a parallel strategy.}
    Both communication volume and the available bandwidth largely determine which configuration is most efficient.

    \item \textbf{TP configurations generally require the least weight memory.}
    In contrast, CP and DP often incur higher weight storage overhead.

    \item \textbf{Activation memory is comparable across TP, CP, and DP.}
    None of the strategies significantly reduces or increases the activation memory footprint.
\end{itemize}


\section{Parallelization Strategy Selection for Transformer and Mamba: A Case Study}
\label{sec:cae_studies}


After establishing the background and theoretical foundations of parallelization strategies, we present a case study using Transformer-like and Mamba-like architectures to illustrate how to select an appropriate parallel strategy in a pretraining setting. We first describe the experimental setup, then report results for Transformer-like models, followed by results for Mamba models. We conclude the empirical analysis with a summary of key findings in Section~\ref{sec:summary-of-case-study}.


\subsection{Experimental Setup}

\paragraph{Models:} We used two Transformer and two Mamba models with parameter sizes of 1B and 7B in both categories for this case study. We employed LLaMA~\cite{zhang2024tinyllama,touvron2023llama} as the representative Transformer model and utilized Mamba-2~\cite{dao2024transformers} to conduct experiments with the Mamba architecture. The detailed configuration of the models in each category are described in detailed in their corresponding sections, ~\ref{sec:attention-model-experiments} and ~\ref{sec:mamba-model-experiments}. None of these models includes a Mixture-of-Experts, so EP is ignored in this empirical analysis.

\paragraph{Evaluation Metrics:} The evaluation metrics include throughput, step time, MFU, and average memory consumption. We consider best strategy with highest MFU which also corresponds to lowest step time and highest throughput. We report memory consumption as an additional metric.

\paragraph{Parallel Strategies:} We examine various combinations of four parallelization schemes: data parallelism (DP), pipeline parallelism (PP), tensor parallelism (TP), and context parallelism (CP). {For each model, we select the hybrid combination of parallelization strategies that delivers the highest throughput and analyze its operation in detail using our in-house analysis tool. Specifically, the tool feeds a random input tensor into the model and measures the elapsed time of each computation and communication operation. These timing measurements are then captured and used for detailed model performance analysis}. We describe the parallelization strategy using either the explicit tuple representation, (DP = m, PP = n, TP = o, CP = p), or the abbreviated format, (m, n, o, p) implying DP = m, PP = n, TP = o, and CP = p.  { As the primary goal of this section is to highlight the impact of parallelization configurations on MFU, we limit our analysis to reporting the average memory consumption, and defer the study of memory optimization techniques to future work.}

\paragraph{Hardware Configuration:}  All experiments are conducted on a single host equipped with eight Ascend 910B Neural Processing Unit (NPU) cards. This NPU is based on the Da Vinci architecture and each offers 60 GB of high bandwidth memory and delivers 378.88 teraflops (TF) of compute for matrix multiplication operations. Throughout our report, we use `Cube' to imply the operations performed at this high speed cores dedicated for batched matrix multiplication operations also known as general matrix multiply (GEMM) operations. The element-wise operations are reported as `Vector' operations.  

\paragraph{Implementation Details} Following many prior studies that use approximately 4 million (4M) tokens per training step for large-scale language models (e.g., \cite{touvron2023llama2}), we also adopt a step size of 4 million tokens in our experiments. To this end we set the micro-batch size to 1, a global batch size of 1024, and a sequence length of 4096 across all considered parallelization scenarios.

\subsection{Attention Experiments}
\label{sec:attention-model-experiments}
In this section, we present our observation and analysis of Transformer models using LLaMA~1B~\cite{meta-llama-2024-llama3.2-1b} and LLaMA~7B~\cite{touvron2023llama} as representative architecture. The configurations of the models under study are presented in Table \ref{tab:lamma_spec}.

\begin{table}[htbp]
\centering
\caption{LLaMA models configuration}
\label{tab:lamma_spec}
\begin{tabular}{lccccc}
\hline
\textbf{Model} & \textbf{Layers} & \textbf{Hidden size} & \textbf{Attention heads} & \textbf{Group query} & \textbf{FFN hidden size} \\
\hline
LLaMA~1B  & 16 & 2048 & 32 & 4& 8192\\
LLaMA~7B  & 32 & 4096 & 16 & -&11008 \\

\hline
\end{tabular}
\end{table}

\subsubsection{Performance Analysis of Parallelization Strategies for LLaMA 7B}

\paragraph{Observation.} We present the detailed result of various parallelization configuration for {LLaMA~7B} model in Table~\ref{tab:lamma7b}. 
We observe that the configuration \textbf{(DP = 4, PP = 2, TP = 1, CP = 1)} achieves \textbf{the highest MFU} of 63.7\%, with a step time of 101.8 s and throughput of 41.2 K tokens/s. This parallelization combination offers the best balance between meeting the memory constraints of each device and maximizing utilization of compute capability. The moderate pipeline depth of two stages allows addressing memory needs with effective parallel execution with minimal communication overhead as the memory footprint of $45.85$ GB remains well within the Ascend 910B’s available capacity. Conversely, maintaining four data-parallel replicas ensures sufficient global batch size for maximizing compute utilization.

In contrast, the configuration \textbf{(DP=1, PP=1, TP=4, CP=2) }exhibits the \textbf{lowest MFU} of 14.8\%, with a step time of 437.1 s and throughput of only 9.6 K tokens/s. The poor utilization stems from using high degree of tensor parallelism and additional context splitting at the cost of reduced data parallelism. Both of these strategy increase inter-NPU communication, and hence, may be beneficial when required to meet other constraints, such as memory or latency for extremely large models or ultra long sequences. Tensor parallelism introduces large all-reduce operations at every layer, while context parallelism fragments the attention computation across sequence segments leading to lower cube utilization for small sequences. The combined overhead of these two, where not required by other constraints, leads to under-utilization of compute capability of the device.

Other intermediate configurations, such as (DP=2, PP=4, TP=1, CP=1) and (DP=1, PP=8, TP=1, CP=1) yields second and third best results of MFU 59.7\% and 51.7\% indicating the benefits of PP as a strong candidate for preferred parallelization strategy to scale to satisfy memory constraints. Other combinations, such as (DP=2, PP=2, TP=1, CP=2) and (DP=2, PP=2, TP=2, CP=1)  achieve MFU value of 43.5\% and 40.0\% indicating additional context or tensor splits produce diminishing returns. These findings confirm that for mid-scale transformer models, coarse-grained model partitioning scaling DP and PP yields the most favorable trade-off between efficiency, latency, and memory scalability.

\begin{table}[ht]
\centering
\caption{Performance Analysis of Parallelization Strategies for LLaMA 7B}
\renewcommand{\arraystretch}{1.2}
\setlength{\tabcolsep}{6pt}
\begin{tabular}{lcccccccc}
\hline
\textbf{DP} & \textbf{PP} & \textbf{TP} & \textbf{CP} &
\textbf{Step Time (s)} & \textbf{Throughput (K tokens/s)} &
\textbf{Mem (GB)} & \textbf{MFU (\%)} \\
\hline
4 & 1 & 2 & 1 & 157.3 & 26.7 & 41.2 & 41.2 \\
4 & 2 & 1 & 1 & 101.8 & 41.2 & 45.9 & 63.7 \\
4 & 1 & 1 & 2 & 145.7 & 28.8 & 60.0 & 44.5 \\
2 & 1 & 4 & 1 & 172.4 & 24.3 & 26.6 & 37.6 \\
2 & 4 & 1 & 1 & 108.5 & 38.6 & 33.7 & 59.7 \\
2 & 1 & 1 & 4 & 180.6 & 23.2 & 55.3 & 35.9 \\
2 & 2 & 2 & 1 & 162.2 & 25.9 & 29.2 & 40.0 \\
2 & 2 & 1 & 2 & 149.2 & 28.1 & 38.9 & 43.5 \\
2 & 1 & 2 & 2 & 213.9 & 19.6 & 36.3 & 30.3 \\
1 & 1 & 8 & 1 & 353.6 & 11.9 & 18.6 & 18.3 \\
1 & 8 & 1 & 1 & 125.5 & 33.4 & 27.2 & 51.7 \\
1 & 1 & 1 & 8 & 363.8 & 11.5 & 53.3 & 17.8 \\
1 & 4 & 2 & 1 & 190.8 & 22.0 & 22.6 & 34.0 \\
1 & 1 & 2 & 4 & 434.6 & 9.7  & 33.8 & 14.9 \\
1 & 2 & 4 & 1 & 214.9 & 19.5 & 20.1 & 30.2 \\
1 & 1 & 4 & 2 & 437.1 & 9.6  & 24.0 & 14.8 \\
1 & 4 & 1 & 2 & 172.5 & 24.3 & 21.5 & 37.6 \\
1 & 2 & 1 & 4 & 214.9 & 19.5 & 35.3 & 30.2 \\
\hline
\end{tabular}
\label{tab:lamma7b}
\end{table}

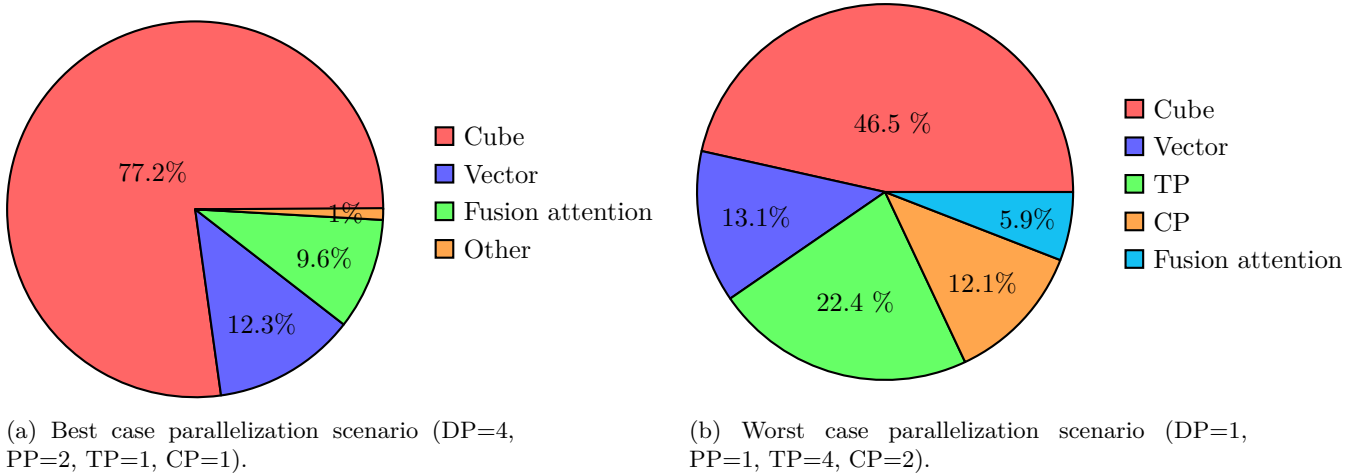
\begin{figure}[ht]
    \centering
    \begin{subfigure}[b]{0.41\textwidth}
            \centering
    \begin{tikzpicture}
        \pie[text=legend, radius=2.5, color={
            red!60,
            blue!60,
            green!60,
            orange!70,
            purple!60,
            cyan!60
        }]
        {77.2/Cube, 12.3/Vector,9.6/Fusion attention, 1/Other}
    \end{tikzpicture}
    \caption{Best case parallelization scenario (DP=4, PP=2, TP=1, CP=1).}
    \label{LLama7b:best}
    \end{subfigure}
    \hfill
    \begin{subfigure}[b]{0.45\textwidth}
        \centering
    \begin{tikzpicture}
        \pie[text=legend, radius=2.5, color={red!60, blue!60, green!60,orange!70,cyan!70}]
        {46.5 /Cube, 13.1/Vector, 22.4 /TP,12.1/CP,
        5.9/Fusion attention
}
    \end{tikzpicture}
    \caption{Worst case parallelization scenario (DP=1, PP=1, TP=4, CP=2).}
    \label{LLama7b:worst}
    \end{subfigure}
    \caption{Runtime composition of LLaMA 7B under best and worst parallelization strategies with (DP, PP, TP, CP) order. The best-case setup increases the cube operation from 46.5\% in the worst case to 77.2\% in the best case. Also, the worst case decreases the vector operation from 13.1\% to 12.3\%.}
\label{LLama7b_time}
\end{figure}

\paragraph{Analysis of Operations.}
To better understand the utilization patterns reported in Table~\ref{tab:lamma7b}, we examine the operation-level execution characteristics of the best and worst performing configurations.  
Figure~\ref{LLama7b_time} shows the kernel-level time distribution for the best and worst configurations, (DP=4, PP=2, TP=1, CP=1) and (DP=1, PP=1, TP=4, CP=2). 
The differences in the ratios of the time spent in different type of operations for these two configuration shed light on the mechanisms responsible for the observed MFU gap.

In the best-performing setting (DP=4, PP=2, TP=1, CP=1) , the execution is dominated by \textbf{Cube} kernels, which account for \textbf{77.2\%} of the total runtime.  These kernels correspond to the dense General Matrix Multiply (GEMM) operations arising in the attention projections and feed-forward pathways. TP = 1 presents large GEMM operations allowing the Ascend NPU cube units to operate close to their peak efficiency. The remaining operations of Vector kernels (12.3\%) and Fused Attention kernels (9.6\%) introduce minimal overhead, as the model is neither tensor- nor context-partitioned. Crucially, the choice of \textbf{PP=2, TP=1, CP=1} ensures that the attention kernels operate on the full sequence with full layer parameter in each pipeline stage requiring no communication overhead other than that between the two pipeline stages. This allows the attention computation to proceed with minimal synchronization and preserving a compute-heavy profile.

In contrast, the worst-performing parallelization combination (DP=1, PP=1, TP=4, CP=2) exhibit a higher ratio of time spent on communication compared to minimal communication overhead in best case. The share of time spent on GEMM operations drops sharply from 77.2\% to \textbf{46.5\%}, indicating lower utilization of Cube. The use of high degree of tensor parallelism (\textbf{TP=4}) splits GEMM operations into four shards reducing matrix dimensions and tile reuse as well as triggering frequent cross-device synchronization. This fragmentation manifests directly in the \textbf{22.4\%} of runtime consumed by TP-related communication, a stark contrast with the negligible overheads in the best case. Context parallelism (\textbf{CP=2}) further amplifies this effect by partitioning the sequence into two segments. This further contributes to an additional \textbf{12.1\%} CP communication overhead. 
Overall, the system spends a substantial portion of its runtime waiting for communication rather than performing numerical computation, which fully explains the sharp drop in MFU and throughput reported in Table~\ref{tab:lamma7b}.

\subsubsection{Performance Analysis of Parallelization Strategies for LLaMA 1B}

\paragraph{Observations.}
To study the impact of model size on different parallelization techniques, we also analyze the LLaMA~1B model. We present the detailed results of LLaMA~1B in table \ref{tab:lamm1b}. The results demonstrate a strong dependence between MFU and the chosen parallelization scheme. Similar to the 7B model, increasing the degree of data parallelism generally leads to higher flops utilization, while extensive tensor or context partitioning reduces efficiency due to communication and frequent synchronization overheads. However, because the 1B model is significantly smaller, the computation-to-communication ratio is lower, making the system more sensitive to communication latency. Thus, for smaller models, simpler parallelization structures—favoring data parallelism—tend to yield superior efficiency.

The \textbf{best configuration} for this model is \textbf{(DP=8, PP=1, TP=1, CP=1)} showcasing MFU of 43.3\%, with a step time of 28.3~s and a throughput of 148.4~K~tokens/s. This setup corresponds to a purely data-parallel configuration, where each NPU holds a full copy of the model and processes a distinct batch of data. Since the model and activations fits into each device’s memory, not using any model or activation parallelization scheme allows for best utilization of computational capability. The absence of tensor or pipeline communication minimizes synchronization delays resulting in minimal overhead across all configurations.

In contrast, the\textbf{ worst configuration} for LLaMA~1B is \textbf{(DP=1, PP=1, TP=4, CP=2)} which exhibits MFU of 6.2\%, with a step time of 197.0~s and a throughput of only 21.3~K~tokens/s. The substantial drop in efficiency arises from the combined use of high degree of tensor and context parallelism. Each transformer layer requires multiple all-reduce operations across four tensor partitions and synchronization of context parallelism communication. Because the model is relatively small, the communication-to-computation ratio becomes unfavorable, and NPUs spend a large portion of time waiting for synchronization rather than performing useful computation. As a result, hardware resources are severely underutilized.

Intermediate configurations, such as (DP=4, PP=2, TP=1, CP=1) and (DP=2, PP=4, TP=1, CP=1), yield MFU values in the 36–38\% range, with throughputs around 124–131~K~tokens/s. These settings demonstrate that shallow pipeline parallelism can complement data parallelism without imposing excessive communication overhead, but further increases in pipeline depth reduce efficiency due to synchronization bubbles. Configurations with $\textit{CP} \ge 4$ or $\textit{TP} \ge 4$ consistently report MFU below~20\%, confirming that fine-grained model partitioning introduces significant communication costs that outweigh computational benefits for small models.


 
  

\begin{figure}[t]
    \centering
    \begin{subfigure}[b]{0.41\textwidth}
            \centering
    \begin{tikzpicture}
        \pie[text=legend, radius=2.5, color={
            red!60,
            blue!60,
            green!60,
            orange!70,
            purple!60
        }]
        {64.0/Cube, 24.2/Vector,11.70/Fusion attention  ,0.1/Others}
    \end{tikzpicture}
    \caption{Best case parallelization scenario (DP=8, PP=1, TP=1, CP=1).}
    \label{fig:LLaMA 1Bbest}
    \end{subfigure}
    \hfill
    \begin{subfigure}[b]{0.45\textwidth}
        \centering
    \begin{tikzpicture}
        \pie[text=legend, radius=2.5, color={red!60, blue!60, green!60,purple!60,cyan!60,yellow!60}]
        {34.3/Cube, 18.4/Vector, 26.0/TP,14.7/CP,6.6/Fusion attention }
    \end{tikzpicture}
    \caption{ Worst case parallelization scenario (DP=1, PP=1, TP=4, CP=2).}
    \label{fig:LLaMA 1Bworse}
    \end{subfigure}
    \caption{Runtime composition of the LLaMA-1B model under the best and worst parallelization strategies, reported in \((\mathrm{DP}, \mathrm{PP}, \mathrm{TP}, \mathrm{CP})\) order. The best-case configuration \((\mathrm{DP}=8, \mathrm{PP}=1, \mathrm{TP}=1, \mathrm{CP}=1)\) is characterized by the highest proportion of cube operations, whereas the worst-case configuration \((\mathrm{DP}=1, \mathrm{PP}=1, \mathrm{TP}=4, \mathrm{CP}=2)\) exhibits a lower cube share with increased vector and parallelization overhead.}
\label{lamma_1b_time}
\end{figure}
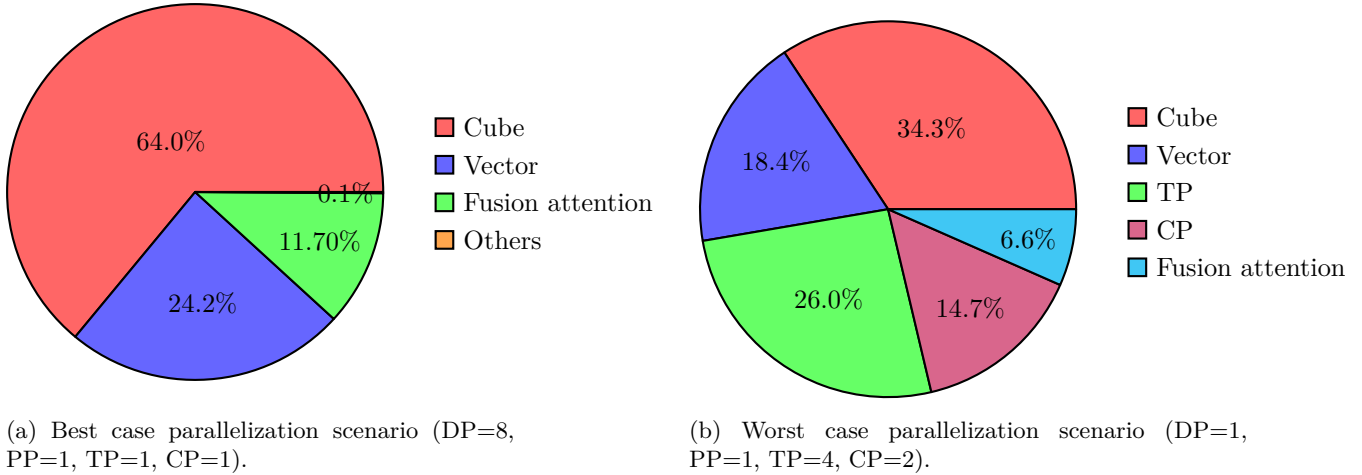


\begin{table}[ht]
\centering
\caption{Performance Analysis of Parallelization Strategies for LLaMA 1B}
\renewcommand{\arraystretch}{1.2}
\setlength{\tabcolsep}{6pt}
\begin{tabular}{lcccccccc}
\hline
\textbf{DP} & \textbf{PP} & \textbf{TP} & \textbf{CP} &
\textbf{Step Time (s)} & \textbf{Throughput (K tokens/s)} &
\textbf{Mem (GB)} & \textbf{MFU (\%)} \\
\hline
8 & 1 & 1 & 1 & 28.3 & 148.4 & 17.3 & 43.3 \\
4 & 1 & 2 & 1 & 61.7 & 68.0 & 11.0 & 19.8 \\
4 & 2 & 1 & 1 & 32.1 & 130.6 & 12.1 & 38.3 \\
4 & 1 & 1 & 2 & 43.7 & 95.9 & 14.1 & 28.0 \\
2 & 1 & 4 & 1 & 75.2 & 55.8 & 7.3 & 16.3 \\
2 & 4 & 1 & 1 & 33.9 & 123.9 & 9.4 & 36.1 \\
2 & 1 & 1 & 4 & 83.2 & 50.4 & 12.6 & 14.7 \\
2 & 2 & 2 & 1 & 46.8 & 89.5 & 8.3 & 26.1 \\
2 & 2 & 1 & 2 & 47.0 & 89.2 & 10.0 & 26.0 \\
2 & 1 & 2 & 2 & 96.7 & 43.4 & 9.4 & 12.6 \\
1 & 1 & 8 & 1 & 150.9 & 27.8 & 4.8 & 8.1 \\
1 & 8 & 1 & 1 & 43.6 & 96.2 & 7.6 & 28.1 \\
1 & 1 & 1 & 8 & 174.4 & 24.1 & 11.4 & 7.0 \\
1 & 4 & 2 & 1 & 59.2 & 70.9 & 6.6 & 20.5 \\
1 & 2 & 4 & 1 & 75.7 & 55.4 & 5.9 & 16.1 \\
1 & 1 & 4 & 2 & 197.0 & 21.3 & 6.5 & 6.2 \\
1 & 4 & 1 & 2 & 53.4 & 78.6 & 7.7 & 23.0 \\
1 & 2 & 1 & 4 & 87.1 & 48.1 & 8.9 & 14.1 \\
\hline
\end{tabular}
\label{tab:lamm1b}
\end{table}

 
\paragraph{Analysis of Operations.}
Figure~\ref{lamma_1b_time} presents the operation breakdown for the purely data-parallel configuration (DP=8, PP=1, TP=1, CP=1) which achieves the highest flops utilization, and for the most communication-heavy configuration of (DP=1, PP=1, TP=4, CP=2) resulting the lowest MFU.   These distributions highlight how the interplay between model size and partitioning strategy shapes the balance between computation and communication.

In the best-performing setting, shown in Figure~\ref{fig:LLaMA 1Bbest}, execution is dominated by \textbf{Cube} (GEMM) kernels, which constitute \textbf{64.0\%} of the total runtime. These large GEMMs form the bulk of the attention and feed-forward computations, and because \textbf{TP=1}, they remain fully intact and run efficiently on the Ascend NPU cube units. The remaining operations—\textbf{Vector} kernels (24.2\%) and \textbf{Fused Attention} kernels (11.70\%) execute with minimal overhead, as the model is neither tensor- nor context-partitioned. Notably, \textbf{CP=1} ensures that attention operates on the complete sequence without requiring any communication needs. Since the 1B model and activations for sequence length of 4K fits well into a single device’s memory, no context splitting is needed to meet memory constraints.  
These together enable each device to sustain a high ratio of computation relative to communication. This compute-dominated execution pattern explains the comparatively high MFU observed for this configuration and underscores why pure data parallelism is particularly effective at the 1B model scale with 4K input sequence.

In contrast, the poorest-performing configuration (DP=1, PP=1, TP=4, CP=2), shown in Figure~\ref{fig:LLaMA 1Bworse}, displays an execution profile that is far more communication-bound. Here, the proportion of time spent in GEMM computation drops to \textbf{34.3\%}, while communication-intensive categories expand significantly. Deep tensor parallelism (\textbf{TP=4}) forces every GEMM to be partitioned into four smaller fragments, reducing matrix sizes and dramatically lowering arithmetic intensity. This fragmentation is directly reflected in the \textbf{26.0\%} of total runtime attributed to TP-related synchronization. At the same time, \textbf{CP=2} divides the sequence dimension into two segments, requiring intermediate communication, which results in \textbf{14.7\%} of runtime spent on CP-related communication. The fused-attention pathway increases to \textbf{6.6\%}, not because attention becomes more computationally demanding, but because sequence partitioning disrupts its natural computation pattern and increases the number of communication steps required per attention head. 
In short, For a small model such as LLaMA~1B, whose computation-to-communication ratio is inherently lower, the communication overhead and synchronization costs of this setting dominate the iteration time, leading to the lowest MFU among all configurations.


\subsubsection{Impact of Parallelism on MFU and Memory Consumption for LLaMA models}


\begin{figure}[htbp]
    \centering
    \begin{subfigure}[b]{0.9\textwidth}
        \centering
        \includegraphics[width=\textwidth]{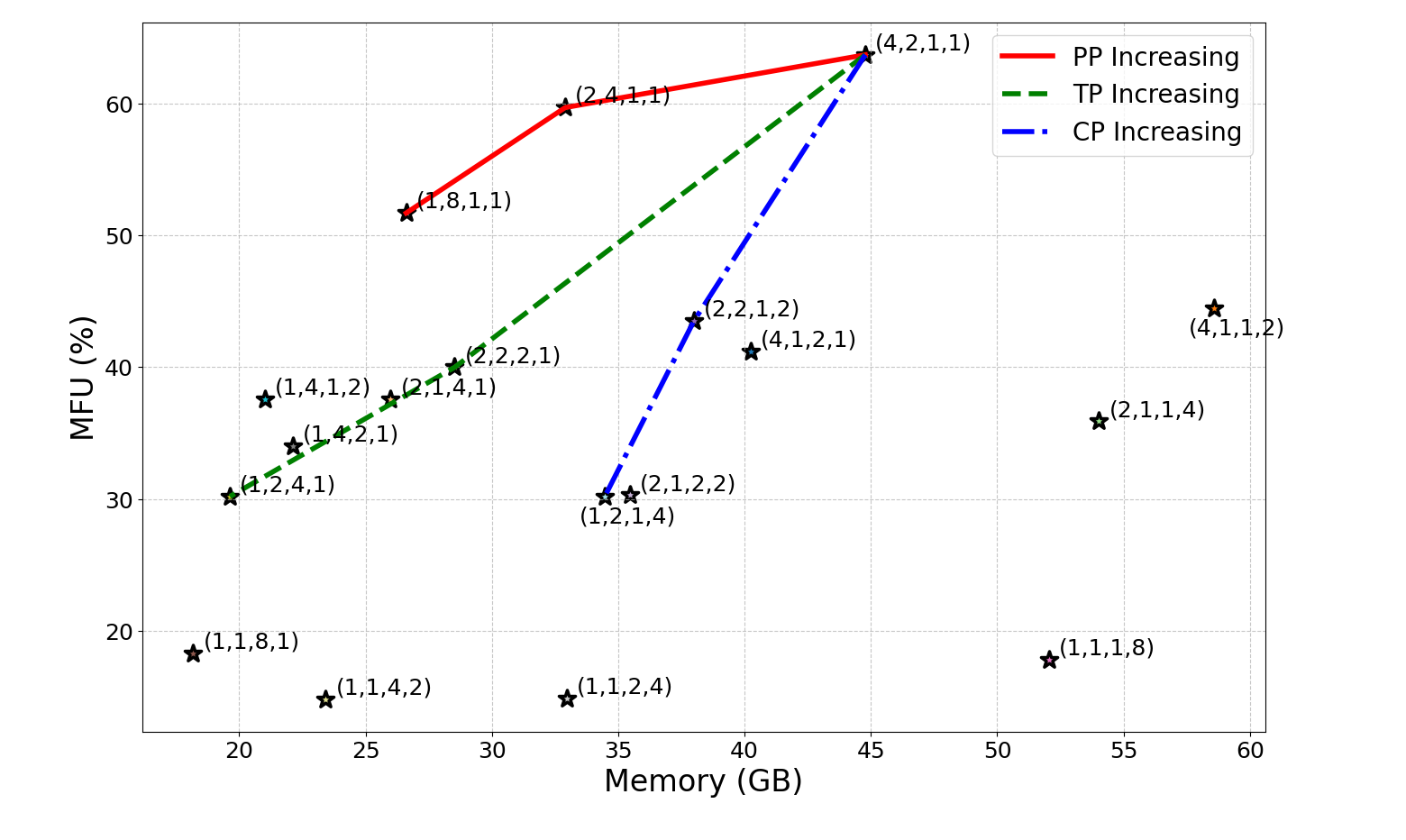}
        \caption{LLaMA 7B}
        \label{fig:Lamma7b_pic}
    \end{subfigure}
    
    \vspace{0.3cm}
    
    \begin{subfigure}[b]{0.9\textwidth}
        \centering
        \includegraphics[width=\textwidth]{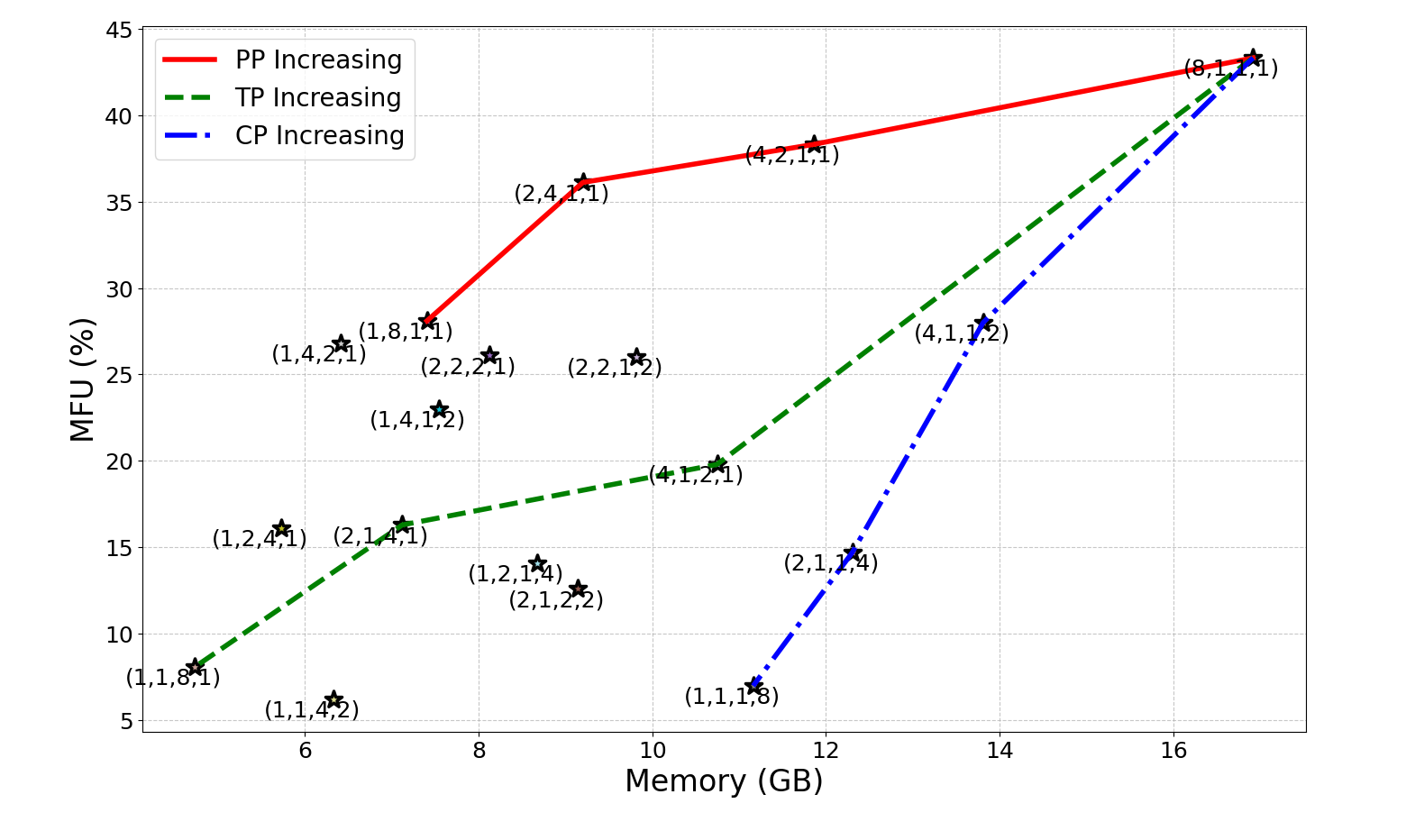}
        \caption{LLaMA 1B}
        \label{fig:Lamma1b_pic}
    \end{subfigure}
    
    \caption{MFU versus memory consumption for LLaMA models under different parallelism types. Starting from the highest MFU datapoint at the rightmost end of the curve, we vary PP, TP, or CP while keeping the total number of devices fixed.
    DP yields the highest efficiency, whereas TP minimizes memory usage; PP provides the strongest middle-ground trade-off. The figure’s tuple format (a,b,c,d) reflects the same (DP=a, PP=b, TP=c, CP=d) values used in the text.}
    \label{fig:Lamma_mem_mfu}
\end{figure}

\begin{figure}[t]
    \centering
    \begin{subfigure}[b]{0.41\textwidth}
            \centering
    \begin{tikzpicture}
        \pie[text=legend, radius=2.5, color={
            red!60,
            blue!60,
            green!60,
            orange!70,
            purple!60,
            cyan!60
        }]
        {49.0/Cube, 43.4/Vector,5.3/TP,
        2.3/DP}
    \end{tikzpicture}
    \caption{Best case parallelization scenario (DP=4, PP=1, TP=2, CP=1).}
    \label{fig:Mamba_7b_best}
    \end{subfigure}
    \hfill
    \begin{subfigure}[b]{0.45\textwidth}
        \centering
    \begin{tikzpicture}
         \pie[text=legend, radius=2.5, color={red!60, blue!60, green!60,purple!60,cyan!60,yellow!60}]
        {27.7/Cube, 71.7/Vector, 0.6/CP }
  
    \end{tikzpicture}
    \caption{Worst case parallelization scenario (DP=1, PP=1, TP=1, CP=8).}
    \label{fig:Mamba_7b_worse}
    \end{subfigure}
    \caption{Runtime operation breakdown of the Mamba-7B model for the best and worst parallelization configurations, reported in \((\mathrm{DP}, \mathrm{PP}, \mathrm{TP}, \mathrm{CP})\) order. In the best-case setup \((\mathrm{DP}=4, \mathrm{PP}=1, \mathrm{TP}=2, \mathrm{CP}=1)\), Cube and vector operations contribute comparably to the overall runtime. In contrast, the worst-case setup \((\mathrm{DP}=1, \mathrm{PP}=1, \mathrm{TP}=1, \mathrm{CP}=8)\) is overwhelmingly dominated by vector operations, with only a minor contribution from Cube operations.}
\label{fig:two_images_mamba7b}
\end{figure}
 
\noindent
To contextualize the behavior observed in Table~\ref{tab:lamma7b}, 
Figure~\ref{fig:Lamma_mem_mfu} illustrates the relationship between 
\textit{Model FLOPs Utilization (MFU)} and memory consumption across the 
all parallelization configurations evaluated for both LLaMA~7B and LLaMA~1B. \textcolor{black}{To better illustrate how MFU varies as data parallelism decreases, which corresponds to the highest MFU, and as other forms of parallelism increase, we connect the corresponding configurations with colored lines in our plots: the DP→PP trend is shown in red, DP→TP in green, and DP→CP in blue.} Results highlight that, for both considered models, the highest and lowest memory consumption are associated with the DP and TP partition groups, respectively. \textcolor{black}{For instance, the average memory consumption of (DP=4, PP=2, TP=1, CP=1) is 45 GB, while the average memory consumption of (DP=1, PP=1, TP=8, CP=1) is 27 GB for LLaMA-7B. A similar trend holds for LLaMA-1B, where the highest memory consumption is 17.3 GB for (DP=8, PP=1, TP=1, CP=1), and the lowest is 4.8 GB for (DP=1, PP=1, TP=8, CP=1).} The former has the  highest memory usage, since no partitioning occurs and both weights and activations are fully replicated across devices and the later has  low memory footprint since  most of the memory consumption related to weights. The PP memory footprint can be considered the second-best scenario in terms of memory consumption after TP, because both activations and weights are partitioned across different pipeline stages. CP ranks third, since it partitions only the activations, and due to the low sequence length, activation memory remains relatively small.   Considering MFU as the main evaluation metric, DP and PP has the first and second ranks respectively. The reason behind this order is that DP has no communication overhead, making it the most efficient. PP ranks second because it requires only limited communication, mainly for transferring activations between stages.

\noindent
 {For both LLaMA based models, graphs indicate that configurations with higher DP are preferable to achieve higher MFU. However, when memory becomes the limiting factor, PP should be considered first, as it offers a low memory footprint while maintaining high computational efficiency. The trade-off between memory and computational efficiency for PP is better than TP as shown in the curve.}

\subsection{Mamba Experiments}
\label{sec:mamba-model-experiments}
We further extend our analysis of parallelization strategy selection to the Mamba models and analyze the Mamba-7B and Mamba-1B models in detail. The configurations of the models under study are presented in Table \ref{tab:mamba}.

\begin{table}[ht]
\centering
\scriptsize
\caption{Mamba models configuration}
\label{tab:mamba}
\begin{tabular}{lccccccc}
\hline
\textbf{Model} & \textbf{Layers} & \textbf{Hidden size} & \textbf{State dim} & \textbf{Num groups} & \textbf{FFN hidden size}& \textbf{Expand}&\textbf{chunk siz}\\
\hline
Mamba-1B  & 16 & 2048 & 64& 8& 4096&2&64\\
Mamba-7B  & 40 & 4096 & 128 & 8&10880 &2&64\\

\hline
\end{tabular}
\end{table}

\subsubsection{Performance Analysis of Parallelization Strategies for Mamaba 7B}

\paragraph{Observation.}
Table~\ref{tab:mamba7b_mfu} summarizes the performance of the Mamba-7B model under the considered parallelization scenarios. Across all tested configurations, MFU ranges from approximately 8\% to 20\%, which is lower than that of LLaMA-based models and reflects the distinct computational pattern of the Mamba architecture—one that emphasizes sequential state updates rather than dense matrix multiplications—resulting in lower effective. The configuration with DP=8 could not satisfy the memory requirements on a single device. Even splitting the activations with CP=2 did not allow the model to fit on a single card as CP only distributed activation memory. Using PP=2 could be helpful to distribute layers across multiple devices; however, unlike the LLaMA model—where the best MFU is achieved with PP=2—the Mamba architecture cannot be placed across two cards due to its very large linear layers that generate state space duality (SSD) components.  As a result, the best MFU is obtained with TP=2, which is the only configuration with minimal model partitioning that fits the model by distributing its large linear and convolutional components efficiently across two tensor partitions while preserving strong data-parallel scaling.
 This configuration \textbf{(DP=4, PP=1, TP=2, CP=1)} achieves the \textbf{highest MFU} of 20.2\%, with a step time of 191.6~s and a throughput of 21.9~K~tokens/s.

The configuration \textbf{(DP=1, PP=1, TP=1, CP=8) }exhibits the \textbf{lowest MFU} of 8.4\%, with a step time of 461.1~s and a throughput of 9.1~K~tokens/s. The steep performance degradation is primarily attributed to the extensive context partitioning (\textit{CP=8}), which significantly reduces the amount of cube operations proportion and consequently lowers overall compute utilization. Moreover, in Mamba, the vector operations contain large fixed costs that do not shrink much when the sequence becomes shorter. These include chunk-boundary prefix work, memory-bound state updates, and fixed NPU overheads such as launching many small kernels, repeated synchronizations, and shared-memory setup. Because these costs remain nearly constant while the matmul-based cube operations shrink substantially, vector time decreases much less than cube time, increasing its share in the overall runtime. 

Intermediate setups such as (DP=2, PP=2, TP=2, CP=1) and (DP=1, PP=2, TP=4, CP=1) yield MFU values around 16–18\% with throughput between 18–20~K~tokens/s. Overall, moderate tensor parallelism provides slight gains by enabling limited intra-layer concurrency, whereas excessive pipeline or context partitioning consistently degrades performance. Heavy context partitioning shifts computation away from cube-intensive operations toward lightweight vector work, leaving the cube units underutilized and the hardware effectively “compute-starved.” Since the model already fits under TP, additional pipelining is unnecessary and would only introduce bubbles, further reducing arithmetic intensity, throughput, and utilization.

\begin{table}[h!]
\centering
\caption{Performance Analysis of Parallelization Strategies for Mamba 7B}
\renewcommand{\arraystretch}{1.2}
\setlength{\tabcolsep}{6pt}
\begin{tabular}{lcccccccc}
\hline
\textbf{DP} & \textbf{PP} & \textbf{TP} & \textbf{CP} &
\textbf{Step Time (s)} & \textbf{Throughput (K tokens/s)} &
\textbf{Mem (GB)} & \textbf{MFU (\%)} \\
\hline
4 & 1 & 2 & 1 & 191.6 & 21.9 & 56.0 & 20.2 \\
2 & 1 & 4 & 1 & 210.9 & 19.9 & 33.2 & 18.4 \\
2 & 2 & 2 & 1 & 215.0 & 19.5 & 40.7 & 18.0 \\
2 & 2 & 1 & 2 & 217.3 & 19.3 & 53.8 & 17.8 \\
2 & 1 & 2 & 2 & 247.5 & 16.9 & 45.0 & 15.6 \\
1 & 1 & 8 & 1 & 299.9 & 14.0 & 20.5 & 12.9 \\
1 & 1 & 1 & 8 & 461.1 & 9.1  & 59.2 & 8.4  \\
1 & 4 & 2 & 1 & 254.2 & 16.5 & 31.1 & 15.2 \\
1 & 1 & 2 & 4 & 435.1 & 9.6  & 38.2 & 8.9  \\
1 & 2 & 4 & 1 & 232.3 & 18.1 & 25.0 & 16.7 \\
1 & 1 & 4 & 2 & 359.3 & 11.7 & 27.3 & 10.8 \\
1 & 4 & 1 & 2 & 264.0 & 15.9 & 38.8 & 14.7 \\
1 & 2 & 1 & 4 & 295.2 & 14.2 & 42.7 & 13.1 \\
\hline
\end{tabular}
\label{tab:mamba7b_mfu}
\end{table}


\paragraph{Analysis of Operations}

Figure~\ref{fig:Mamba_7b_best} presents the operation breakdown for the configuration (DP=4, PP=1, TP=2, CP=1), which achieves the highest overall performance, and, in contrast, the configuration (DP=1, PP=1, TP=1, CP=8), which yields the lowest MFU.   A comparison between these two settings shows that the primary reason for MFU degradation is the substantial reduction in Cube FLOPs from 49.0 to 27.7 percent. This drop results from overly partitioning the sequence length, which reduces the amount of GEMM computation per device and leaves the Ascend hardware under-utilized. Since the computational throughput of Ascend devices is significantly higher for GEMM operations than for vector operators, the reduction in GEMM workload is especially detrimental, further reducing the MFU. 

Compared to attention-based models, the Mamba architecture exhibits a lower ratio of Cube FLOPs in both its best and worst configurations. This indicates that Mamba is inherently less efficient in terms of hardware utilization, as it provides fewer high-intensity GEMM operations relative to attention. This confirms that Mamba has lower A.I. than LLaMA, showing that Mamba operates in a memory-bound regime, while attention-based models are compute-bound. Another important observation is that the communication overhead of CP in Mamba is substantially lower than in attention. This is because Mamba only exchanges its recurrent state, whose dimension does not depend on the sequence length, whereas attention-based models must exchange the full key–value tensors, leading to significantly higher communication costs.


\subsubsection{Performance Analysis of Parallelization Strategies for Mamaba 1B}

\paragraph{Observation.}
The MFU values for Mamba~1B  range from  3.1\% to 20.4\% across all parallelization configurations under consideration. This behavior reflects Mamba’s state-space architecture: sequential state propagation limits fully parallelizable work, reducing arithmetic intensity relative to attention-based models. Results show that the configuration \textbf{(DP=8, PP=1, TP=1, CP=1)} achieves the  \textbf{highest MFU} of 20.4\% , with a step time of 29.1~s and a throughput of 144.3~K~tokens/s. This setting benefits from distributing the batch across eight devices, thereby maximizing locally available computation per card while keeping both pipeline and context partitioning disabled. The absence of deep model partitioning minimizes synchronization, and each device operates on sufficiently large micro-batches allowing the hardware to operate at its highest possible effective capacity. 


In contrast, the configuration \textbf{(DP=1, PP=1, TP=1, CP=8) }yields the \textbf{lowest MFU} of 3.1\%, with a step time of 191.1~s and a throughput of only 22.0~K~tokens/s. The same trend holds for TP=8 and PP=8, with slightly better computational use. This degradation arises from excessive decrease of the workload on each device. In addition to that, communication overhead and bubble time are two reasons for the performance degradation of TP and PP.

Intermediate configurations such as (DP=4, PP=1, TP=2, CP=1), (DP=2, PP=1, TP=4, CP=1) and (DP=1, PP=2, TP=4, CP=1) yield MFU values in the 6--12\% range. These results indicate that tensor parallelism provides moderate improvements by enabling limited intra-layer concurrency, yet its effectiveness diminishes rapidly once additional parallel dimensions are introduced. Configurations with \(\textit{PP} \ge 4\) or \(\textit{CP} \ge 2\) consistently exhibit reduced MFU. Increasing pipeline depth introduces unnecessary bubble time, diminishing compute overlap. In additio

\begin{table}[ht]
\centering
\caption{Performance Analysis of Parallelization Strategies for Mamba 1B (MFU recalculated)}
\renewcommand{\arraystretch}{1.2}
\setlength{\tabcolsep}{6pt}
\begin{tabular}{cccccccc}
\hline
\textbf{DP} & \textbf{PP} & \textbf{TP} & \textbf{CP} &
\textbf{Step Time (s)} & \textbf{Throughput (K tokens/s)} &
\textbf{Mem (GB)} & \textbf{MFU (\%)} \\
\hline
8 & 1 & 1 & 1 & 29.1 & 144.3 & 24.8 & 20.4 \\

4 & 1 & 2 & 1 & 47.7 & 88.0 & 14.7 & 12.4 \\
4 & 2 & 1 & 1 & 48.2 & 87.0 & 16.8 & 12.3 \\
4 & 1 & 1 & 2 & 48.2 & 86.9 & 18.1 & 12.3 \\

2 & 1 & 4 & 1 & 68.2 & 61.5 & 9.3  & 8.7  \\
2 & 4 & 1 & 1 & 81.3 & 51.6 & 11.6 & 7.3  \\
2 & 1 & 1 & 4 & 97.7 & 42.9 & 15.4 & 6.1  \\
2 & 2 & 2 & 1 & 64.5 & 65.0 & 11.6 & 9.2  \\
2 & 1 & 2 & 2 & 84.6 & 49.6 & 11.6 & 7.0  \\
2 & 2 & 1 & 2 & 66.9 & 62.7 & 13.5 & 8.9  \\

1 & 1 & 8 & 1 & 128.9 & 32.5 & 5.8  & 4.6  \\
1 & 8 & 1 & 1 & 148.1 & 28.3 & 8.7  & 4.0  \\
1 & 1 & 1 & 8 & 191.1 & 22.0 & 14.2 & 3.1  \\

1 & 4 & 2 & 1 & 98.8 & 42.4 & 7.6  & 6.0  \\
1 & 1 & 2 & 4 & 187.4 & 22.4 & 9.8  & 3.2  \\
1 & 2 & 4 & 1 & 84.1 & 49.9 & 6.9  & 7.1  \\
1 & 1 & 4 & 2 & 166.6 & 25.2 & 7.6  & 3.6  \\
1 & 4 & 1 & 2 & 104.5 & 40.1 & 9.3  & 5.7  \\
1 & 2 & 1 & 4 & 118.0 & 35.6 & 11.4 & 5.0  \\
\hline
\end{tabular}
\label{tab:mamba1b_mfu}
\end{table}

\paragraph{Analysis of Operations} 
Here, we examine the best and worst configurations for Mamba-1B, namely (DP=8, PP=1, TP=1, CP=1) and (DP=1, PP=1, TP=1, CP=8). As discussed extensively earlier, the highest MFU is achieved with (DP=8, PP=1, TP=1, CP=1), since this configuration maximizes data distribution while avoiding the additional communication overhead introduced by TP and CP as well as the pipeline bubble time caused by PP. In this setting, the highest possible amount of GEMM computation is executed. \textcolor{black}{Similarly, context parallelism partitions the recurrent computation across
devices, this shift increases the relative contribution of vector communication overhead compared to compute time, causing communication to dominate and ultimately degrading MFU.}

Regarding the Mamba model, comparing the Cube FLOP ratio of Mamba-1B with LLaMA~1B shows a significant drop from 64.0 to 42.8 percent, alongside an increase in the vector-operation ratio from 24.2 to 57.1 percent. This reduction in cube-intensive computation, and the corresponding rise in vector-dominated workloads, is the main factor behind the MFU degradation when moving from LLaMA to Mamba (from 43.3\% to 20.4\%), reflecting the inherent structure of the Mamba architecture. 

The operation-time breakdown of the lowest MFU configuration of (DP=1, PP=1, TP=1, CP=8) shows similar pattern to the Mamba-7B worst case scenario results. Excessively partitioning the sequence length reduces the proportion of Cube operations and forces the model to operate primarily on lightweight vector computations. This behavior indicates that, for relatively short sequence lengths such as 4K, the Mamba architecture does not benefit from high CP values.

\begin{figure}[ht]
    \centering
    \begin{subfigure}[b]{0.41\textwidth}
            \centering
    \begin{tikzpicture}
         \pie[text=legend, radius=2.5, color={
            red!60,
            blue!60,
            green!60,
        }]
        {42.8/Cube, 57.1/Vector, 0.1/Other}
    \end{tikzpicture}
    \caption{Best case parallelization scenario (DP=8, PP=1, TP=1, CP=1).}
    \label{fig:maba1bbest}
    \end{subfigure}
    \hfill
    \begin{subfigure}[b]{0.45\textwidth}
        \centering
    \begin{tikzpicture}
        \pie[text=legend, radius=2.5, color={red!60, blue!60, green!60,purple!60,cyan!60,yellow!60}]
        {22.5/Cube, 77.3/Vector, 0.2/CP }
    \end{tikzpicture}
    \caption{Worst case parallelization scenario (DP=1, PP=1, TP=1, CP=8).}
    \label{fig:maba1b_worse}
    \end{subfigure}
    \caption{Runtime operation breakdown of the Mamba-1B model under the best and worst parallelization configurations, reported in \((\mathrm{DP}, \mathrm{PP}, \mathrm{TP}, \mathrm{CP})\) order. In the best-case configuration \((\mathrm{DP}=8, \mathrm{PP}=1, \mathrm{TP}=1, \mathrm{CP}=1)\), vector operations slightly outweigh Cube operations, with both contributing substantially to overall runtime. In the worst-case configuration \((\mathrm{DP}=1, \mathrm{PP}=1, \mathrm{TP}=1, \mathrm{CP}=8)\), execution time is dominated by vector operations, while Cube operations account for a much smaller fraction.}
\label{fig:two_images_mamba1b}
\end{figure}
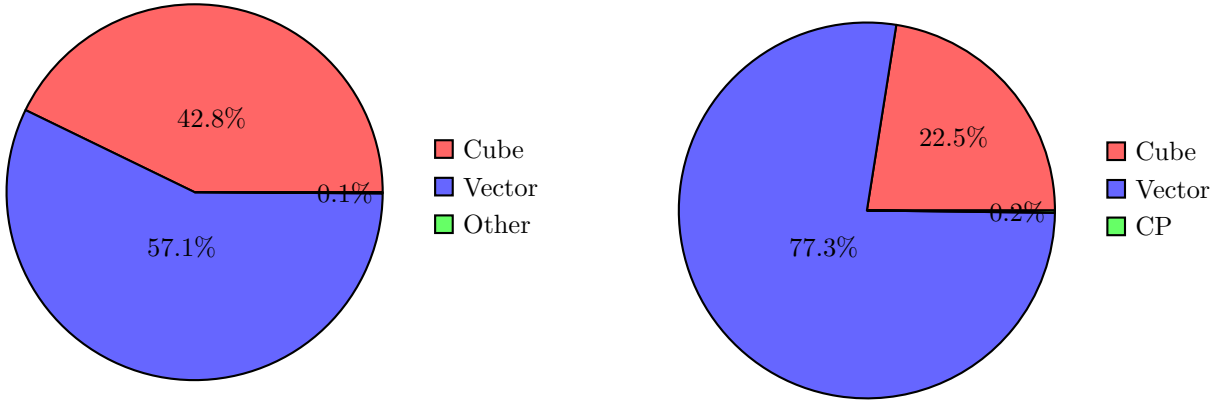

\subsubsection{Impact of Parallelism on MFU and Memory Consumption for Mamba models}
\begin{figure}[htbp]
    \centering
    \begin{subfigure}[b]{0.9\textwidth}
        \centering
        \includegraphics[width=\textwidth]{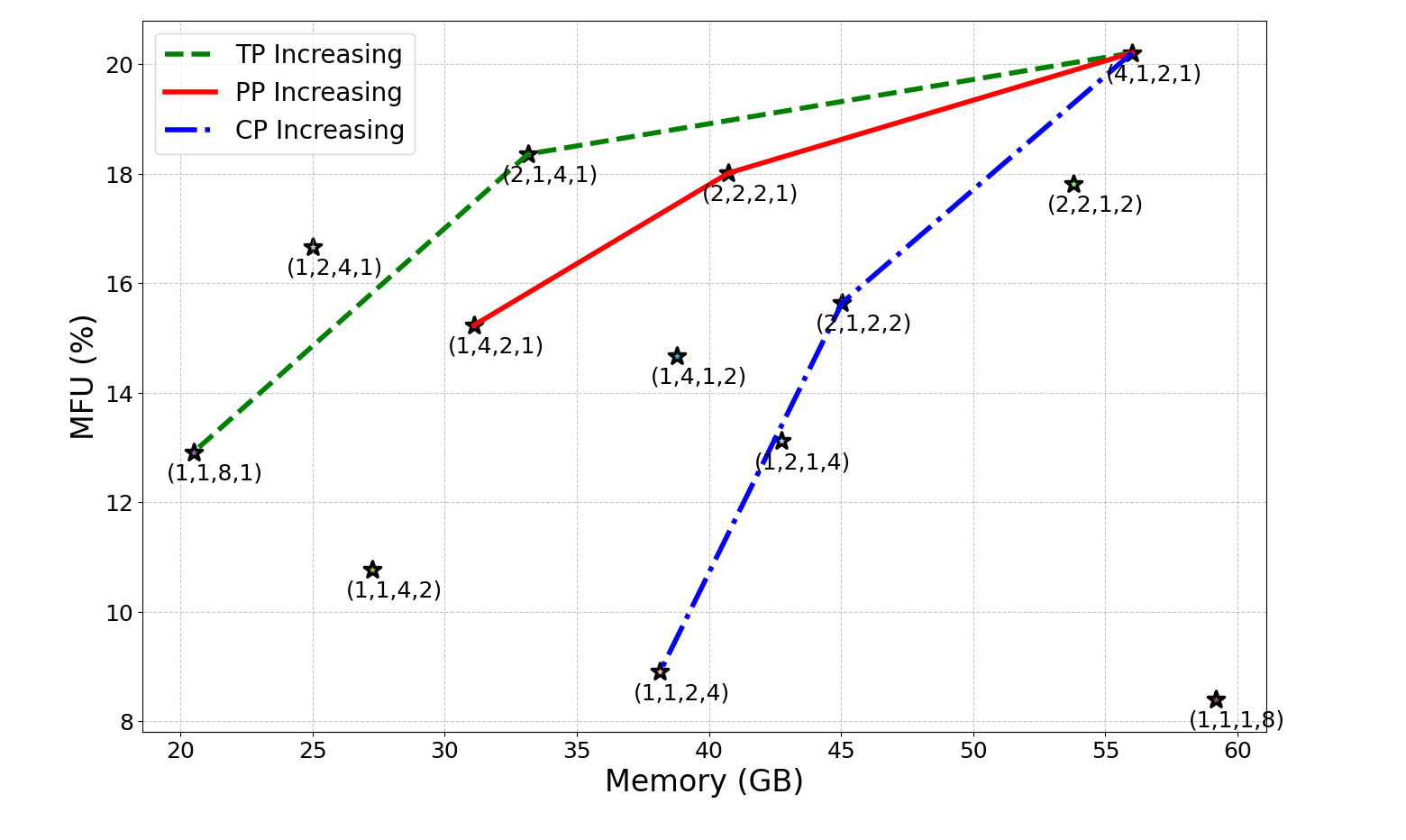}
        \caption{Mamba 7B}
        \label{fig:Mamba7B_mem}
    \end{subfigure}

    \vspace{0.4cm}

    \begin{subfigure}[b]{0.9\textwidth}
        \centering
        \includegraphics[width=\textwidth]{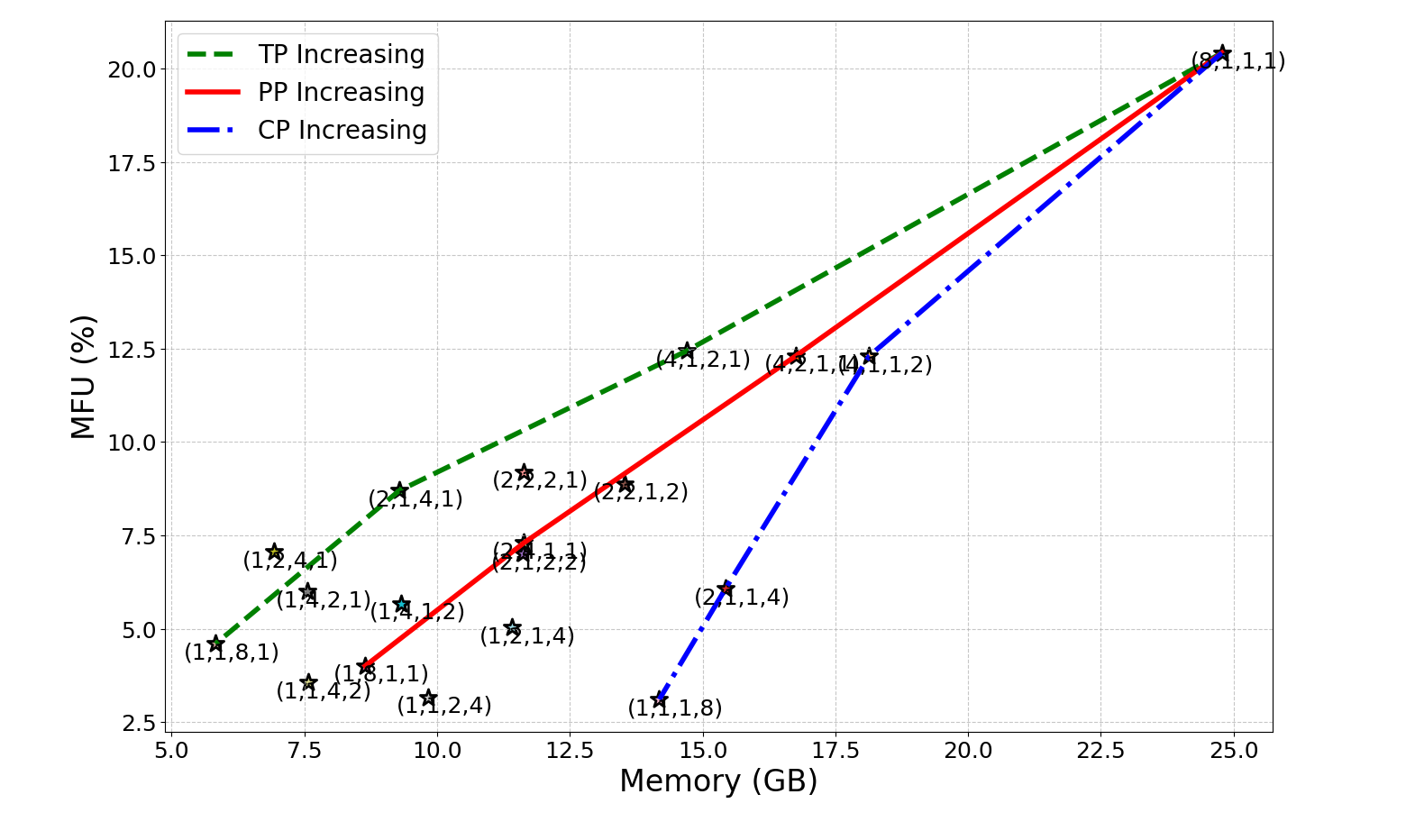}
        \caption{Mamba 1B}
        \label{fig:Mamba1B_mem}
    \end{subfigure}

    \caption{MFU as a function of memory consumption for Mamba under different parallelism settings.  Starting from the highest MFU datapoint at the rightmost end of the curve, we vary PP, TP, or CP while keeping the total number of devices fixed.
    TP provides the most memory-efficient scaling, whereas CP results in the lowest MFU and highest memory usage.
    The tuple format $(a,b,c,d)$ denotes $(\mathrm{DP}=a,\mathrm{PP}=b,\mathrm{TP}=c,\mathrm{CP}=d)$.}
    \label{fig:Mamba_mem}
\end{figure}

In Figure \ref{fig:Mamba_mem}, we examine MFU as a function of consumed memory across different parallelization configurations. In both figures, the dashed green curve corresponding to TP consistently appears above the others, indicating that TP achieves the highest MFU for the Mamba model. In addition, TP exhibits a smaller memory footprint compared to the other parallelization strategies. \textcolor{black}{For example, Mamba-7B exhibits an average memory usage of 56 GB under the configuration (DP=4, PP=1, TP=2, CP=1), compared to 20.5 GB for (DP=1, PP=1, TP=8, CP=1). A similar pattern is observed for Mamba-1B, where memory usage peaks at 24.8 GB with (DP=8, PP=1, TP=1, CP=1) and reaches a minimum of 5.8 GB under (DP=1, PP=1, TP=8, CP=1).} In contrast, the dashed blue CP curve is positioned at the bottom, reflecting the lowest performance in terms of both memory usage and MFU. Therefore, when the model does not fit on a single device, TP is the most effective choice to apply first. For both LLaMA and Mamba, TP and CP deliver the highest and lowest memory efficiency, respectively. The underlying reason is the short sequence length, where model weights dominate the memory footprint compared to activations, making TP more beneficial than CP under these conditions. PP ranks second in memory efficiency because it partitions layers across devices but still requires each stage to store full activations for its portion of the sequence, leading to   higher memory consumption than TP.


\subsection{Summary of Empirical Insights}
\label{sec:summary-of-case-study}
To obtain a broader perspective on the impact of parallelization strategy combinations across different model types and sizes, we summarize the best and worst strategies for all four models in our case study (Table~\ref{tab:best-vs-worse-summary}). Notably, for small models of both Attention and Mamba, the optimal parallelization strategy involves using data parallelism (DP) only. As model size grows beyond the memory capacity of a single device, incorporating additional forms of model parallelism becomes beneficial, though only up to a moderate scale. For instance, scaling the Attention model from 1B to 7B parameters benefits from pipeline parallelism (PP = 2) while reducing DP to 4. Similarly, scaling Mamba from 1B to 7B parameters benefits from tensor parallelism (TP = 2) while also reducing DP to 4.


\begin{table}[h!]
\centering
\caption{Best and worst parallelization for Attention and Mamba models in case study}
\renewcommand{\arraystretch}{1.2}
\setlength{\tabcolsep}{6pt}
\begin{tabular}{|l|ll|ll|}
\hline
\multirow{2}{*}{Model} & \multicolumn{2}{l|}{{Best Case}}    & \multicolumn{2}{l|}{{Worst Case}}    \\ \cline{2-5}
                  & \multicolumn{1}{l|}{{(DP,PP,TP,CP)}} & {MFU(\%)} & \multicolumn{1}{l|}{{(DP,PP,TP,CP)}} & {MFU(\%)} \\ \hline
LLaMA~1B    & \multicolumn{1}{l|}{{(8,1,1,1)}} & {43.3}  & \multicolumn{1}{l|}{{(1,1,4,2)}} & {6.2} \\ \hline
LLaMA~7B    & \multicolumn{1}{l|}{(4,2,1,1)} & 63.7 & \multicolumn{1}{l|}{(1,1,4,2)}& 14.8 \\ \hline
Mamba-1B    & \multicolumn{1}{l|}{(8,1,1,1)} & 20.4 & \multicolumn{1}{l|}{(1,1,1,8)} & 3.1 \\ \hline
Mamba-7B    & \multicolumn{1}{l|}{(4,1,2,1)} & 20.2 & \multicolumn{1}{l|}{(1,1,1,8)} & 8.4 \\ \hline
\end{tabular}
\label{tab:best-vs-worse-summary}
\end{table}



Insights on parallelization design practices evident from our experiments are summarized below:

\begin{itemize}
    \item For small models (e.g., 1B), data parallelism (DP) is generally preferred, as both the model and activations fit within memory constraints.
    
    \item As model size increases to moderate scales (e.g., 7B), introducing limited model parallelism (such as PP or TP) while maximizing DP within memory limits provides a balanced trade-off and yields the highest MFU. {More specifically, PP is the most favorable choice for LLaMA models, whereas TP is more effective for Mamba models.}
    \item Further increases in the degree of model parallelism (PP/TP) are expected to benefit extremely large models. Increasing context parallelism (CP) is expected to be advantageous for very long sequences; however, due to resource limitations, experimental validation of these scenarios is left for future work.

   
\end{itemize}

\section{Open Challenges and Future Directions}
\label{sec:future_directions}

This survey has examined acceleration strategies for efficient training and inference. Although the design of distributed AI infrastructure has advanced to support large-scale models and ultra-long contexts, significant challenges remain. In this section, we identify key unresolved issues and highlight research gaps that point to promising directions for future exploration. Some major open challenges include:

\paragraph{Resource Utilization} Despite recent advancements in distributed training and inference systems, optimal hardware utilization remains a significant challenge. Overhead exists from various sources, such as memory access delays in memory-bound operations and communication wait times in bandwidth-bound operations. Additional auxiliary overhead also exists; some examples include kernel launches for small operations and cases where tensor shape dimensions do not favor optimal tiling to the parallel cores of the accelerators. Recent progress in Model FLOPS Utilization (MFU) has shown improved efficiency in attention-based architectures, yet a substantial gap remains compared to the hardware’s peak performance. Moreover, emerging models such as Mamba exhibit significantly lower MFU compared to conventional architectures like Transformers. This gap highlights the need for further research into both model architecture design and system-level optimization strategies that can better exploit hardware capabilities.

\paragraph{Energy Considerations}
As frontier models approach the trillion-parameter scale, energy constraints are becoming as critical as hardware availability. OpenAI has recently committed to gigawatt-scale compute, and Meta is investing billions to acquire roughly 350,000 GPUs—but such expansions are only feasible when sufficient power capacity exists to operate them. Energy availability increasingly shapes both research capability and service deployment \cite{chung2025ml, elsworth2025measuring}. A fundamental yet underexplored trade-off exists between inference latency or training throughput and energy consumption, forming a Pareto frontier that is not fully addressed in this work or in the broader literature. Future research should systematically examine the theoretical and empirical relationships between energy usage, power efficiency, and distributed parallelization strategies. Advancing this line of inquiry will be essential for enabling both sustainable model development and truly \emph{Green AI}.

\paragraph{Communication Overhead} Recent research on reducing communication overhead in distributed training and inference has increasingly focused on communication overlap strategies. Emerging approaches in this direction have explored fine-grained slicing techniques, wherein the communication of one slice is overlapped with the computation of another. While these methods offer promising improvements in efficiency, they continue to suffer from tail-end overhead. Additionally, they also impose strict synchronization requirements at intermediate stages of the algorithm. These constraints present open challenges and underscore the need for further investigation into more flexible and scalable collective communication strategies.

\paragraph{Auto-parallelism} Trending approaches for automatically finding optimal parallelization strategies typically rely on hierarchical search-based approaches. These methods often depend on profiling or executing the model under various configurations, which is costly, time-consuming, and often infeasible for extremely large models. Cost model based approaches are on the other hand are known to produce suboptimal solutions and often exhibit hardware dependencies. These limitations highlight the need for further research into efficient auto-parallelism methods.

In light of the current limitations, further advancements in model acceleration may be achievable through the following avenues:
\paragraph{Model-System Codesign} Recent research in \emph{Efficient AI} has observed a shift toward models and infrastructure evolving through a co-design paradigm rather than advancing independently. Despite this shift, joint optimization of hardware-aware model architectures and efficient distributed systems remains underexplored. Key challenges in this direction include designing efficient architectures that generalize across multiple device types and accounting for compute-bound, memory-bound, bandwidth-bound, and energy consumption considerations in model layer design. Further research can focus on developing concrete mathematical models of design objectives and solid theoretical foundations, in addition to relying solely on empirical or cost-model simulation–based optimization.

\paragraph{Parallel Strategies for Model Blocks}  
Future work should place greater emphasis on developing principled guidelines for selecting parallelization strategies tailored to specific model blocks, such as attention mechanisms, Mamba blocks, or other emerging architectures across different modalities, such as text, vision, or speech. These architectures are certainly different and should be carefully studied. Current research often provides insights with limited theoretical examples into how different parallel strategies should be chosen under varying architectural or workload settings. Constructing analytical models with formal mathematical underpinnings could help bridge this gap, enabling systematic evaluation of parallelization strategies. Such models would create a closed loop between workload analysis and architecture design, guiding the development of next-generation architectures that are both computation and communication-efficient.  

\paragraph{Neural Autoparallelism}  
Another promising direction is the advancement of autoparallelism---automatically selecting optimal parallelization strategies within the vast design space. In parallel, the rise of AI-driven techniques for solving low-level system challenges, such as automated GPU kernel generation (AI for Kernels), provides strong precedent for leveraging neural approaches in distributed training \cite{popcorn2025}. Similarly, methods inspired by neural architecture search, equipped with tailored cost functions, have shown success in discovering efficient mixtures-of-experts (MoE) and transformer architectures~\cite{jawahar2022automoe, gu2025jet, bakos2025searching}. Extending these ideas to the problem of parallel strategy selection could yield significant breakthroughs. We therefore recommend further exploration of neural and learning-based methods for auto-parallelism as a key avenue for future research.

\section{Conclusion}


In this survey, we conduct a comprehensive review of distributed strategies for efficient LLM training and inference. Based on our findings, we present system design guidelines with a concrete formulation of design objectives. Most existing works show improvements in distributed systems to support trillion-parameter models with context lengths of up to millions of tokens, enabling efficient execution through a hybrid combination of multiple parallelism strategies and other acceleration techniques. Auto-parallelism is also gaining traction, where the optimal strategy can be searched in a simulation environment using a cost model for deploying large models. We presented theoretical and empirical analyses to guide researchers and practitioners to select parallel strategies for model development. In light of current trends, we also discussed prevailing concerns and potential directions for further improvements in accelerating large models. Finally, we hope this study serves as a hands-on reference for better understanding the current research progress on distributed strategies for efficient AI infrastructure and helps readers with system design recommendations and guidance for future research. Grounded in prior work, this survey reflects the current state of the field and is intended to guide and inspire subsequent research efforts.






\subsubsection*{Broader Impact Statement}

This work focuses on the analysis and design of distributed parallelization strategies for training and deploying large language models. While our contribution is primarily methodological and systems-oriented, the techniques discussed here may indirectly influence the scale, accessibility, and efficiency of future AI systems.

On the positive side, improved parallel strategies can reduce computational overhead, lower energy consumption, and make large-scale training more cost-effective. These advances may enable a broader community of researchers and practitioners to experiment with state-of-the-art models, fostering openness, reproducibility, and innovation in the field.

However, the increased efficiency and scalability that arise from better parallelism may also accelerate the development and deployment of increasingly large models, which can raise ethical considerations. These include heightened energy usage, environmental impact, concentration of computational resources among a few institutions, and the downstream risks associated with more capable AI systems, such as misuse, bias propagation, or harmful applications. Although our work does not introduce new model capabilities, improved distributed training frameworks could indirectly contribute to enabling such systems.

We encourage users of this research to consider these broader implications when applying parallelization strategies in practice, particularly with respect to responsible resource allocation, transparency in system design, and the societal implications of large-scale AI development. Our analysis is intended to inform and support principled decision-making rather than promote unconstrained scaling, and we hope it motivates further research into environmentally conscious and ethically aligned AI infrastructure.

\subsubsection*{Acknowledgments}
The authors gratefully acknowledge the support of the Toronto Ascend team. The authors also extend their special thanks to Austin Wen for generously sharing his knowledge and insights throughout this work.


\bibliography{tmlr}
\bibliographystyle{tmlr}

\appendix


\end{document}